\pdfoutput=1
\documentclass[11pt]{article}
\usepackage[table]{xcolor}

\usepackage{acl}
\usepackage{tikz-dependency}
\usepackage{times}
\usepackage{latexsym}
\usepackage[T1]{fontenc}
\usepackage[utf8]{inputenc}
\usepackage{enumitem}

\usepackage{amsmath}
\usepackage{amssymb}

\usepackage{graphicx}
\usepackage{booktabs}
\usepackage{longtable}
\usepackage{multirow}
\usepackage{makecell} 
\usepackage{adjustbox}
\usepackage{rotating}
\usepackage{pdflscape}
\usepackage{pdfpages}
\usepackage{tikz}
\usepackage{caption}
\usepackage{siunitx}
\usepackage{gb4e}
\noautomath

\usepackage{tabularx}
\usepackage{booktabs}

\usepackage[T2A,T1]{fontenc}
\usepackage{cmap}
\newcommand{\cyr}[1]{{\fontencoding{T2A}\selectfont #1}}
\usepackage{array}
\usetikzlibrary{patterns}
\usepackage{wrapfig}
\noautomath  
\definecolor{softgreen}{RGB}{216, 235, 204} %
\definecolor{walk}{RGB}{0,128,60}

\usepackage[table]{xcolor}

\definecolor{liccol}{RGB}{0,114,178}   
\definecolor{npicol}{RGB}{213,94,0}    
\newcommand{\mpnpi}[1]{\textcolor{npicol}{#1}}              
\newcommand{\mplic}[1]{\textcolor{liccol}{\underline{#1}}}  
\newcommand{\mpbad}[1]{\underline{#1}}                      
\usepackage{todonotes}
\makeatletter
\newcommand*\iftodonotes{\if@todonotes@disabled\expandafter\@secondoftwo\else\expandafter\@firstoftwo\fi} 
\makeatother

\newcommand{\blimppi}{BLiMPPI}

\title{BLiMPPI: Benchmark of Linguistic Minimal Pairs for Polarity Items}

\author{Arzu Burcu Güven$^{1}$ \quad Anna Rogers$^{1}$ \quad Leonie Weissweiler$^{2}$ \quad Joakim Nivre$^{3}$ \\ \textbf{Rob van der Goot$^{1}$} \\[.1cm]
$^1$IT University of Copenhagen 
$^2$Leipzig University \& ScaDS.AI Dresden/Leipzig \\
$^3$Uppsala University \\
\texttt{argy@itu.dk, arog@itu.dk, leonie.weissweiler@uni-leipzig.de} \\
\texttt{joakim.nivre@lingfil.uu.se, robv@itu.dk} 
}

\begin{document}
\maketitle
\begin{abstract}

Negative polarity items (NPIs), such as \textit{anymore} and \textit{ever}, are grammatical only in select environments, predominantly in the context of negation. Because their grammaticality depends on both syntax and semantics, they have a rich, multilingual literature in theoretical linguistics, yet work on NPIs in language models is almost entirely confined to English. We introduce \blimppi, a multilingual evaluation suite of over 12,000 NPI minimal pairs in 33 languages, built by a pipeline that discovers NPIs in Universal Dependencies (UD, \citealp{ud}) treebanks, labels their licensing environments, and perturbs these into minimal pairs. Evaluating 12 multilingual and 2 monolingual models alongside n-gram baselines, we find that multilingual models struggle most with lower-resource languages and with pairs that are not n-gram solvable. We further examine the effects of NPI typology, language family, licensing variability, and NPI--licenser distance.\footnote{Data: \url{https://github.com/arzuburcuguven/blimppi-data}; each language subset is released under the license of its source treebank, as listed in the repository.}

\end{abstract}
\section{Introduction}
\label{sec:Intro}

Negative polarity items (NPIs) are a well-known phenomenon that has received a lot of attention from semanticists \citep{zwartsThreeTypesPolarity1998, woudenNegativeContextsCollocation1997, giannakidouLongDistanceLicensingNegative1997}. NPIs are barred from simple affirmative sentences and generally occur under negation \citep{hoeksemaNegativePolarityItems2000}. Prototypical examples are pronouns and adverbs, e.g.\ English \textit{anything},
\textit{at all}, \textit{anymore}: 

{\small
\begin{exe}
  \ex \label{ex:npi-good} She does \textcolor{liccol}{not} drink milk \textcolor{npicol}{anymore}.
  \ex \label{ex:npi-bad} *She drinks milk \textcolor{npicol}{anymore}.
\end{exe}
}
\noindent
The examples above illustrate how an NPI requires a specific environment, here negation, to be grammatical, and what the minimal pairs produced by our pipeline test for.

For LM evaluation, NPIs provide a unique test bed because they sit at the intersection of meaning and structure where meaning controls \textit{grammaticality}. It is essentially a semantic problem whether an adverb or pronoun fits the context, but when the conditions of NPI use are violated, the effect aligns with that of structural violations rather than semantic incongruity \citep{10.3389/fpsyg.2019.00376,
Yanilmaz02012018}. Moreover, NPIs are challenging empirically: they have consistently been among the hardest phenomena in different syntactic benchmarks \citep{marvinTargetedSyntacticEvaluation2018a, warstadtBLiMPBenchmarkLinguistic2020}. Even for humans, proficiency in NPIs comes later; it has been shown that as late as the age of 11--12 years, children lack adult-like comprehension of NPIs \citep{schwab2021acquisition}. 


Psycholinguistic \citep{xiangDependencydependentInterferenceNPI2013, Yanilmaz02012018, 10.3389/fpsyg.2019.00376} and linguistic studies (\autoref{tab:resources}, Appendix~\ref{sec:Consulted}) have examined NPIs in various languages. However, in model evaluation, multilingual benchmarks have so far focused mainly on agreement \citep{jumeletMultiBLiMP10Massively2026}, with in-depth NPI studies only in English \citep{jumeletLanguageModelsUse2021, decarloNPIsArentExactly2023}.


In light of the properties outlined above, we propose a multilingual evaluation suite targeting NPIs which extends the understanding of LMs' linguistic skills in two directions: (i) how multilingual LMs perform on rare, challenging phenomena, and (ii) whether, and how evenly, such a phenomenon is learned across typologically diverse languages. Our contributions are as follows:
 \begin{itemize}[noitemsep,topsep=2pt,leftmargin=*]
    \item We present a minimal pair evaluation suite on NPIs, \blimppi, constructed by perturbing natural sentences.  
    \item We perform an in-depth analysis of 14 different LMs across 8 families; 
    provide an overview of what properties of a naturalistic benchmark cause the most adversity to model performance.

    \item We look at 7 monolingual models' checkpoints to compare how NPIs are ``learned'' across typologies.
    
    
\end{itemize}

\section{Background: NPIs}
\label{sec:background}

In this section, we first give an overview of the environments in which NPIs occur. Then we introduce two concepts: \textit{strength hierarchy} and \textit{negative concord}. These concepts are used to identify cases where NPIs are more challenging for models (Section~\ref{tab:acc-subsets}).


\paragraph{What is a licensing environment?} 

NPIs are made grammatical through licensers, which can manifest as function words, adverbs or sentence types.\footnote{A better term would be \textit{felicitous}, since NPIs are not strictly syntactic; however, to keep the term inventory brief, we opt for ``grammatical''.}  The canonical example of a licenser is negation.
\noindent As we have seen in example (\ref{ex:npi-good}), a licenser is necessary to make the NPI grammatical. In addition to negation, near-negative adverbs such as \textit{hardly}, \textit{barely}, or
\textit{rarely} can also license NPIs; we mark \textcolor{liccol}{licensers} in blue and \textcolor{npicol}{NPIs} in orange throughout:

{\small
\begin{exe}
\ex She \textcolor{liccol}{hardly} drinks milk \textcolor{npicol}{anymore}.
\end{exe}
}
\noindent Questions (\ref{ex:Que}), comparatives (\ref{ex:Comp}) and antecedents of
conditionals (\ref{ex:Cond}) are also common licensing environments:

{\small
\begin{exe}
\ex \label{ex:Que} Does she \textcolor{npicol}{ever} drink milk?
\ex \label{ex:Comp} Now, she drinks milk \textcolor{liccol}{more than} \textcolor{npicol}{ever}.
\ex \label{ex:Cond} \textcolor{liccol}{If} she \textcolor{npicol}{ever} drinks milk again, I will be surprised.
\end{exe}
}
\noindent
In addition to the above, we consider the following as licensers: \textit{n}-series items
(e.g., English \textit{nobody} as in \textit{`\textcolor{liccol}{Nobody} \textcolor{npicol}{ever} drinks milk \textcolor{npicol}{anymore}'}), \textit{without}-type adpositions, negative predicates, and other near-negative expressions, as well as non-negative environments that license NPIs, such as superlatives, the quantifier \textit{only}, and modals.


\paragraph{Strength hierarchy.}  
Idiosyncratic distribution of NPIs, commonly found in
Germanic languages, refers to the fact that each item has its own
preferred licensing environments. For example, some NPIs, such as \textit{either}, prefer only negative contexts:

{\small
\begin{exe}
\ex She does \textcolor{liccol}{not} drink milk \textcolor{npicol}{either}.
\end{exe}
}

\noindent This is in contrast to NPIs that are more flexible; for example, \textit{ever} can be licensed by questions (\ref{ex:Que}), but \textit{either} cannot: 

{\small
\begin{exe}
\ex *Does she drink milk \textcolor{npicol}{either}?
\end{exe}
}
\noindent
Moreover, NPIs are challenging because they constitute a rare phenomenon with a theoretically unbounded dependency; consider:

{
\small
\begin{exe}
\ex \label{ex:LD} \textcolor{liccol}{No} woman who struggles with the condition of lactose intolerance \textcolor{npicol}{ever} drinks milk.
\end{exe}
}
\noindent
The intervening clauses do not interfere with the NPIs' grammaticality.

\citet{zwartsThreeTypesPolarity1998} and  \citet{woudenNegativeContextsCollocation1997}
therefore sort NPIs into classes according to their licenser preferences:
\textit{superstrong} NPIs, such as \textit{either}, which are restricted to
negative environments; \textit{strong} NPIs, such as \textit{anymore}, which
also occur in near-negative ones; and \textit{weak} NPIs, such as
\textit{ever}, which are grammatical across a much wider range of licensing
contexts.

\paragraph{Negative Concord Items.} A higher-level classification regarding NPIs is concerned with typology. In Germanic languages, Latin and Mandarin Chinese, NPIs have freer licensing paradigms \cite{haspelmathIndefinitePronouns2017}, but a much larger group of languages spanning the rest of the Indo-European family, as well as Turkic, Uralic, and Semitic families, has rigid paradigms called negative concord. In this paradigm, the counterparts of NPIs are called Negative Concord Items (NCIs) (e.g., Polish \textit{nikt} `nobody', Spanish \textit{nada} `nothing', Turkish \textit{hiç kimse} `nobody'), as they require a strict collocation with negation. 

{\small
\begin{exe}
\ex \gll \mpnpi{Hiç kimse} gel\mplic{me}di.\\
     Nobody come-\textsc{NEG}-\textsc{pst}\\
\glt `Nobody came'
\ex \gll \mpnpi{Nikt} \mplic{nie} przyszedł.\\
     nobody \textsc{neg} come.\textsc{pst.3sg.m}\\
\glt `Nobody came'
\end{exe}
}
\noindent
Furthermore, other NPI environments such as conditionals, questions or comparatives rarely function as licensers in NC languages. 

We combine NPIs and NCIs in a single polarity item benchmark, following \citet{haspelmathNegindefinitesNegativeConcord2024}, who argues that NCIs are a subgroup of NPIs. 





\section{Related Work}

\paragraph{NPIs.}NPIs were among the target phenomena in the now seminal work by \citet{marvinTargetedSyntacticEvaluation2018a}, who introduced forced choice by minimal pairs over LM probabilities as the standard paradigm for syntactic evaluation. They create the minimal pairs via templatic generation, focusing on quantifier licensers (\textit{no}, \textit{few}). Their results show that n-gram and LSTM LMs (one multi-task trained with CCG supertagging) perform poorly on NPIs. \citet{warstadtBLiMPBenchmarkLinguistic2020} include NPIs among the 12 phenomena in BLiMP, constructing 7 test sets resulting in 7000 minimal pairs. However, all of these examples focus on a single NPI: \textit{ever}. Non-English benchmarks that include NPIs are ZhoBLiMP \citep{liuZhoBLiMPSystematicAssessment2024ac} in Chinese and RuBLiMP \citep{taktashevaRuBLiMPRussianBenchmark2024a} in Russian. All of these studies report that NPIs are either the most challenging probe in their benchmark or among the most challenging.\footnote{Two other studies in Korean and Spanish also include NPIs as part of their benchmarks, but they only report results aggregated over all phenomena~\citep{perez-mayosAssessingSyntacticCapabilities2021b, shinKoBALTKoreanBenchmark2025}}

\citet{jumeletLanguageModelsUse2021} and \citet{decarloNPIsArentExactly2023} study NPIs more closely; both focus on variability of licensing environments. 
\citet{jumeletLanguageModelsUse2021} train a probe to distinguish licensing from non-licensing contexts, and show that it generalizes to held-out environment types and NPIs. \citet{decarloNPIsArentExactly2023} examine each NPI with respect to its own licensing environments and find that variability in licensing hampers model performance. In other words, a strong NPI such as \textit{exactly}, which only partakes in negative contexts, is learned better than a weak NPI such as \textit{any}.

\paragraph{Multilingual Benchmarks.}
MultiBLiMP \citep{jumeletMultiBLiMP10Massively2026} is the first large-scale multilingual benchmark of minimal pairs, covering 101 languages with over 128,000 pairs, targeting subject-verb agreement.
It derives typological information automatically and language-agnostically from UD treebanks;  NPIs are too rare for this, so we construct our benchmark language by language.

We see our work as complementary to prior NPI evaluations and multilingual benchmarks. Unlike existing NPI benchmarks, it is multilingual, and the data is naturalistic rather than templatic, leading to greater syntactic and lexical diversity. We also target a much rarer phenomenon compared to MultiBLiMP.


\section{Benchmark Creation}

\subsection{Data sources}
\label{sec:data_sources}

\begin{table}[t]
\centering
\footnotesize
\setlength{\tabcolsep}{4pt}
\begin{tabular}{@{}llrcl@{}}
\toprule
\textbf{Typology} & \textbf{Language} & \textbf{N} & \textbf{Src.} & \textbf{Ver.} \\
\midrule
\multirow{16}{*}{\shortstack[l]{NC (strict)\\\textit{N=8,171}}}
  & Belarusian & 171 & D & MT \\
  & Bulgarian & 131 & D & MT \\
  & Croatian & 71 & D & Hum \\
  & Czech & 1,213 & D & Hum \\
  & Greek & 87 & R & Hum \\
  & Hungarian & 13 & R & Hum \\
  & Lithuanian & 46 & R & Hum \\
  & Persian & 449 & D & MT \\
  & Polish & 764 & R+D & Hum \\
  & Romanian & 246 & D & Hum \\
  & Russian & 4,199 & D & Hum \\
  & Serbian & 24 & D & Hum \\
  & Slovak & 294 & D & MT \\
  & Slovenian & 168 & D & MT \\
  & Ukrainian & 295 & D & Hum \\
\midrule
\multirow{2}{*}{NC (non-str.)}
  & Catalan & 220 & R & MT \\
  & Spanish & 272 & R & Hum \\
\midrule
\multirow{4}{*}{\shortstack[l]{NC\\\textit{N=1,351}}}
  & Armenian & 42 & R & Hum \\
  & Finnish & 400 & R & Hum \\
  & Hebrew & 96 & R & MT \\
  & Turkish & 813 & R+D & Hum \\
\midrule
\multirow{12}{*}{\shortstack[l]{NPI\\\textit{N=2,852}}}
  & Chinese & 34 & R & Hum \\
  & Danish & 55 & R & Hum \\
  & Dutch* & 1,771 & R & Hum \\
  & English* & 212 & R & Hum \\
  & German & 428 & R & Hum \\
  & Hindi & 107 & R & Hum \\
  & Icelandic & 47 & R & MT \\
  & Indonesian & 6 & R & MT \\
  & Nor.\ Bokm\aa{}l & 43 & R & Hum \\
  & Nor.\ Nynorsk & 30 & R & Hum \\
  & Swedish & 59 & R & Hum \\
  & Urdu & 60 & R & Hum \\
\midrule
\multicolumn{5}{l}{\textbf{Total} \quad 33 languages \quad N = 12,866} \\
\bottomrule
\end{tabular}
\caption{Polarity-item typology, item count, NPI inventory source, and verification status for the 33 languages in \blimppi. NPI inventory source (Src.): R = linguistic resources, D = automatic discovery, R+D = both. All data from UD v2.18, except Dutch* \citep{hoeksemaNegatiefpolaireUitdrukkingenVan2024} and English* (UD v2.18 + NPI bank).}
\label{tab:languages}
\end{table}


Our starting point is UD v2.18 treebanks that satisfy the following criteria: (i) at least 100k words, since NPIs are relatively rare and would not occur with sufficient frequency;\footnote{With the exception of Hungarian, for which we had the opportunity to work with a Hungarian linguist.} 
(ii) languages where the negation paradigm is stable;\footnote{Following the Jespersen cycle \citep{jespersen1917negation}, some languages have fluctuating negation paradigms e.g.\ French \textit{ne} $\rightarrow$ \textit{ne \dots\ pas} $\rightarrow$ colloquial \textit{pas}, where the former minimizer \textit{pas} blurs the line between negator and NPI.} (iii) languages with symmetrical negation;\footnote{Typologically, negation is either symmetrical, where affirmative and negative clauses differ only in the negation marker, or asymmetrical, where negation causes further structural changes.} (iv) negation recoverable from UD annotations or lemmas \citep[see][for the challenges of identifying negation in UD]{findlay-haug-2025-negation}; (v) a living language, so that validation with speakers is possible. Appendix~\ref{sec:Excl} lists the excluded treebanks with reasons.

We complement UD with English data from the NPI bank \citep{npibank} and Dutch data from the Dutch NPI Lexicon~\citep{hoeksemaNegatiefpolaireUitdrukkingenVan2024}, bringing the total to 33 languages. These complementary sources were already annotated for NPIs, licensers and licenser types. We refer to the original studies for the annotation details.

\subsection{Minimal Pair Creation Pipeline}


\paragraph{NPI selection.} We use linguistic resources to construct NPI inventories per language where possible. \citet{haspelmathIndefinitePronouns2017} provides a detailed illustration of negative pronouns and their usage patterns in 40 languages, half of which are used in our study. As such, we use it as our backbone. Other theoretical resources, albeit centered on fewer languages, cover the licensing conditions
in more detail \citep{hoeksemaNegativePolarityItems2000,
woudenNegativeContextsCollocation1997,giannakidouPolaritySensitivityNonVeridical1998}. These are supplemented by language-specific resources, listed in Appendix \ref{sec:Consulted}.

When linguistic resources are not available, we discover the NPIs from UD treebanks. For automatic NPI discovery, we only target lemmas of the determiner, pronoun, and adverb types. This means that any item of these lemma types that occurs at least 5 times in the corpus is a candidate. We iterate over all the sentences in the treebanks, checking whether the candidate occurs more often in licensing environments than non-licensing ones. 

We consider a given candidate an NPI based on a threshold set empirically per language, since the reliability of the ratio varies with the type of item involved. To set the ratio, we translate the resulting candidate list, and choose the cut-off that retains the negative indefinites. We find that higher thresholds around 70\% work well in NC languages: NCIs are largely unambiguous and occur almost exclusively under negation, which gives both the UD annotations and our filters a reliable signal. In languages where NPIs have a more diverse distribution and exhibit polysemy, the co-occurrence statistics alone can be misleading.\footnote{Consider \textit{any} having both free-choice and NPI functions \citep{kadmonAny1993}.} Indeed, in NPI languages only 56\% of NPI occurrences in UD fall in a detected licensing environment, whereas in NC languages this holds for 78\% of NCI occurrences.

When an NPI list is available from linguistic resources, we remove UPOS- and lemma-based discovery constraints; canonical NPIs could be in multi-word form (English \textit{at all}; Hindi \textit{ko{\=i} bh{\=i}} `any') or of an unexpected part of speech.\footnote{For example, the Dutch verb \textit{hoeven} is an NPI, whereas NPIs commonly are pronouns, determiners, and adverbs.}

\paragraph{Licenser Detection and Resolution.} 

We use syntactic filters to detect licensing environments; an overview is given in \autoref{tab:envs}.

\begin{table}[t]
\centering\small
\renewcommand{\arraystretch}{1.15}
\begin{tabularx}{\linewidth}{@{}>{\raggedright\arraybackslash}p{1.6cm}>{\raggedright\arraybackslash}X@{}}
\toprule
\textbf{Environm.} & \textbf{Detection} \\
\midrule
Standard negation & Bound negation on the predicate head; unbound negation attached to the clause head via \texttt{advmod}, \texttt{advmod:neg}, \texttt{compound}, \texttt{compound:lvc}, \texttt{cop}, \texttt{aux}, or \texttt{aux:pass}. Checked for \texttt{Polarity=Neg} and the negation marker's lemma. \\
\addlinespace
N-series & For languages without duplex negation, n-series items are added to the language configuration. Pronouns: must share the clause head with the NPI. Determiners: \texttt{det} of the subject or object of the NPI clause, preceding the NPI. \\
\addlinespace
Matrix clause & For non-island complements (\texttt{ccomp}, \texttt{csubj}, \texttt{csubj:cop}, \texttt{xcomp}), walk to the matrix clause and check for scoping licensers (\autoref{fig:resolution}). \\
\addlinespace
Lexical licensers & \textit{Without}-type adpositions, near-negatives, negative predicates, comparatives and conditional antecedents, tracked by lemma and syntactic position. \\
\addlinespace
Questions \& superlatives & Feature annotations and punctuation. \\
\bottomrule
\end{tabularx}
\caption{Licensing environments and how they are detected in UD trees.}
\label{tab:envs}
\end{table}

\begin{figure}[t]
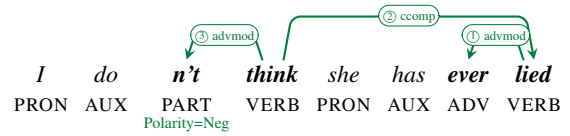

\centering
\footnotesize
\begin{dependency}[edge style={->, thick, walk}, label style={draw=walk, text=walk, rounded corners, fill=white, font=\scriptsize}, edge unit distance=2.2ex]
\begin{deptext}[column sep=0.2em, row sep=0ex]
  \textit{I} \& \textit{do} \& \textbf{\textit{n't}} \& \textbf{\textit{think}} \& \textit{she} \& \textit{has} \& \textbf{\textit{ever}} \& \textbf{\textit{lied}} \\
  \textsc{pron} \& \textsc{aux} \& \textsc{part} \& \textsc{verb} \& \textsc{pron} \& \textsc{aux} \& \textsc{adv} \& \textsc{verb} \\[-1.2ex]
  \& \& \textcolor{walk}{\tiny Polarity=Neg} \& \& \& \& \& \\
\end{deptext}
\depedge{4}{3}{\textcircled{\tiny 3} advmod}
\depedge[edge unit distance=1ex]{4}{8}{\textcircled{\tiny 2} ccomp}
\depedge{8}{7}{\textcircled{\tiny 1} advmod}
\end{dependency}
\caption{Licenser resolution for the NPI \textit{ever}.
\textcircled{\scriptsize 1} Climb from the NPI to its clause head \textit{lied}; no licenser is detected there.
\textcircled{\scriptsize 2} Climb to the matrix clause, allowed because \texttt{ccomp} is scope-permeable.
\textcircled{\scriptsize 3} On \textit{think}, the detectors find \textit{n't}, an \texttt{advmod} with \texttt{Polarity=Neg}: a \emph{matrix} licenser.}
\label{fig:resolution}
\end{figure}

As shown in \autoref{fig:resolution}, we look at more than one clause for the licenser. We discard examples with more than one licenser with overlapping scopes from our pipeline. In the case of examples containing multiple NPIs (7.5\%), we include them in our benchmark with the flag that they have multiple NPIs. The effect of the multiple NPIs on model accuracy is negligible; on average, they account for a difference of +0.9 points. Appendix~\ref{app:scope} provides more details on scope resolution.


\paragraph{Minimal pair construction} 

We create the pairs via modification functions for UD trees. While most of our functions are language-agnostic, we adjusted them to language-specific needs when necessary (e.g., restoring contracted forms in Romanian: nu-l $\rightarrow$ îl); a full list of language-specific interventions can be found in Appendix~\ref{appendix:spe-cases}. In the case of standard negation, we either drop the particle/auxiliary or re-inflect the verb to its positive form. For unremovable licensers, such as negative predicates, adverbs, or n-series items, we swap them with their positive counterparts. Appendix~\ref{appendix:swapdrop} show the effect of removal and swap-based perturbations on model performance. The complete set of strategies, namely reinflection, re-inversion of yes/no questions, licenser deletion, and licenser swapping, are exemplified with \blimppi{} pairs below:

{\small
\begin{exe}
\ex[]{Reinflection (Turkish)}
\begin{xlist}
\ex \gll Ahmet \mpnpi{hiçbir} şey yap\mplic{ma}dı.\\
     Ahmet \textsc{neg.det} thing do-\textsc{neg}-\textsc{pst}\\
\glt `Ahmet didn't do anything.'
\ex \gll \llap{*}Ahmet \mpnpi{hiçbir} şey yaptı.\\
     Ahmet \textsc{neg.det} thing do-\textsc{pst}\\
\end{xlist}

\ex[]{Negation removal (Ukrainian)}
\begin{xlist}
\ex \glll \mpnpi{\cyr{Ніколи}} \mpnpi{\cyr{нікого}} \mplic{\cyr{не}} \cyr{вдарив.}\\
     Nikoly nikoho ne vdaryv\\
     never nobody.\textsc{gen} \textsc{neg} hit.\textsc{pst.m.sg}\\
\glt `He never hit anyone.'
\ex \glll \llap{*}\mpnpi{\cyr{Ніколи}} \mpnpi{\cyr{нікого}} \cyr{вдарив.}\\
     Nikoly nikoho vdaryv\\
     never nobody.\textsc{gen} hit.\textsc{pst.m.sg}\\
\end{xlist}

\ex[]{$n$-series swap  (Dutch)}
\begin{xlist}
\ex \gll Mij \mpnpi{hoef} je \mplic{geen} succes te wensen.\\
     me.\textsc{dat} need.\textsc{prs} you no success to wish.\textsc{inf}\\
\glt `You needn't wish me any success.'
\ex \gll \llap{*}Mij \mpnpi{hoef} je \mpbad{een} succes te wensen.\\
     me.\textsc{dat} need.\textsc{prs} you a success to wish.\textsc{inf}\\
\end{xlist}

\ex[]{Quantifier swap (Swedish)}
\begin{xlist}
\ex \gll \mplic{Få} av dem återsåg henne \mpnpi{någonsin}.\\
     few of them again.saw her ever\\
\glt `Few of them ever saw her again.'
\ex \gll \llap{*}\mpbad{Några} av dem återsåg henne \mpnpi{någonsin}.\\
     some of them again.saw her ever\\
\end{xlist}

\ex[]{Aux inversion (English)}
\begin{xlist}
\ex[]{\mplic{Have you} sent this to congress \mpnpi{yet}?}
\ex[*]{\mpbad{You have} sent this to congress \mpnpi{yet}.}
\end{xlist}
\end{exe}
}

\paragraph{Quality control}

We subject the benchmarks to iterative quality checks. Seven languages -- Dutch, Finnish, Hungarian, Lithuanian, Norwegian-Bokmål, Norwegian-Nynorsk and Swedish -- were developed in consultation with a native speaker, a linguist or both. As such, their examples were vetted during construction. For the rest, we use
parallel UD benchmarks where available, and otherwise machine translation, to
check that the intended contrast was achieved. Finally, we conduct an auditing step for 17 further languages, asking participants to review a sample of 30 pairs per language (23 or 15 where fewer were available). Each sample consists of 10 pairs where models preferred the correct sentence most strongly, 10 where they preferred the wrong one most strongly, and 10 selected at random.

This yielded 638 judgements over 428 distinct pairs, of which 61 (9.6\%) were
flagged as not carrying the intended contrast. Flag rates range from 0\% to
30\% and separate by typology: 0--7\% in the NCI languages against 10\% for Danish, 20\% for German and 30\% for English. We attribute the Germanic figures to the greater likelihood of a perturbed sentence landing on a grammatical reading due to polysemy. Per-language results are given in Appendix~\ref{app:audit}.

We then apply post-fixes: items flagged as not carrying the intended contrast are corrected or discarded, and where the problem was systematic, we revise the configuration and regenerate. Most flags were
source-sentence problems: an already odd treebank sentence, or an idiom
whose fixed parts carry no \texttt{fixed} relation, so the perturbation breaks
the idiom rather than the licensing. The two systematic English cases are described in Appendix~\ref{app:audit}. 

\section{Results}

\begin{table*}[t]
\centering
\scriptsize
\setlength{\tabcolsep}{3.2pt}
\resizebox{\textwidth}{!}{%
\begin{tabular}{l l | cccccccccccc | c | ccc}
\toprule
 &  & \multicolumn{2}{c}{Llama-3.1} & \multicolumn{1}{c}{Aya} & \multicolumn{4}{c}{Gemma-3} & \multicolumn{3}{c}{Qwen3} & \multicolumn{2}{c}{Apertus} & \multicolumn{1}{c}{\textit{Mean}} & \multicolumn{2}{c}{Goldfish} & \multicolumn{1}{c}{Other} \\
\cmidrule(lr){3-4} \cmidrule(lr){5-5} \cmidrule(lr){6-9} \cmidrule(lr){10-12} \cmidrule(lr){13-14} \cmidrule(lr){16-17} \cmidrule(lr){18-18}
\texttt{ISO} & Language & \texttt{8B} & \texttt{70B} & \texttt{32B} & \texttt{1B} & \texttt{4B} & \texttt{12B} & \texttt{27B} & \texttt{8B} & \texttt{14B} & \texttt{32B} & \texttt{8B} & \texttt{70B} &  & \texttt{100MB} & \texttt{1GB} & \texttt{5-gram} \\
\midrule
\texttt{hye} & Armenian & \cellcolor[HTML]{59B380}95.2 & \cellcolor[HTML]{71C087}92.9 & \cellcolor[HTML]{59B380}95.2 & \cellcolor[HTML]{B9E193}85.7 & \cellcolor[HTML]{50A377}\textbf{97.6} & \cellcolor[HTML]{59B380}95.2 & \cellcolor[HTML]{50A377}\textbf{97.6} & \cellcolor[HTML]{8ACC8D}90.5 & \cellcolor[HTML]{8ACC8D}90.5 & \cellcolor[HTML]{B9E193}85.7 & \cellcolor[HTML]{A1D790}88.1 & \cellcolor[HTML]{50A377}\textbf{97.6} & \cellcolor[HTML]{73C187}\textbf{92.7} & \cellcolor[HTML]{A1D790}88.1 & \cellcolor[HTML]{59B380}95.2 & \cellcolor[HTML]{FECC94}64.3 \\
\texttt{bel} & Belarusian & \cellcolor[HTML]{C4E697}84.2 & \cellcolor[HTML]{62B883}94.2 & \cellcolor[HTML]{8FCE8E}90.1 & \cellcolor[HTML]{8FCE8E}90.1 & \cellcolor[HTML]{7BC589}91.8 & \cellcolor[HTML]{56AD7D}95.9 & \cellcolor[HTML]{4D9D75}\textbf{98.2} & \cellcolor[HTML]{CDEA9D}83.0 & \cellcolor[HTML]{94D18F}89.5 & \cellcolor[HTML]{7BC589}91.8 & \cellcolor[HTML]{58B17F}95.3 & \cellcolor[HTML]{54A97B}96.5 & \cellcolor[HTML]{7EC68A}\textbf{91.7} & \cellcolor[HTML]{94D18F}89.5 & \cellcolor[HTML]{52A579}97.1 & \cellcolor[HTML]{FCFECD}75.4 \\
\texttt{bul} & Bulgarian & \cellcolor[HTML]{CBE99C}83.2 & \cellcolor[HTML]{66BB84}93.9 & \cellcolor[HTML]{FEEAAE}69.5 & \cellcolor[HTML]{A5D891}87.8 & \cellcolor[HTML]{52A77A}96.9 & \cellcolor[HTML]{55AC7C}96.2 & \cellcolor[HTML]{4A9671}\textbf{99.2} & \cellcolor[HTML]{A5D891}87.8 & \cellcolor[HTML]{86CA8C}90.8 & \cellcolor[HTML]{9DD590}88.5 & \cellcolor[HTML]{4D9C74}98.5 & \cellcolor[HTML]{4A9671}\textbf{99.2} & \cellcolor[HTML]{86CA8C}\textbf{91.0} & \cellcolor[HTML]{77C388}92.4 & \cellcolor[HTML]{52A77A}96.9 & \cellcolor[HTML]{FFF8C6}73.3 \\
\texttt{cat} & Catalan & \cellcolor[HTML]{6CBE86}93.2 & \cellcolor[HTML]{51A478}97.3 & \cellcolor[HTML]{82C88B}91.4 & \cellcolor[HTML]{98D38F}89.1 & \cellcolor[HTML]{73C187}92.7 & \cellcolor[HTML]{55AB7C}96.4 & \cellcolor[HTML]{4C9A73}\textbf{98.6} & \cellcolor[HTML]{98D38F}89.1 & \cellcolor[HTML]{6CBE86}93.2 & \cellcolor[HTML]{64BA83}94.1 & \cellcolor[HTML]{56AD7D}95.9 & \cellcolor[HTML]{56AD7D}95.9 & \cellcolor[HTML]{66BB84}\textbf{93.9} & \cellcolor[HTML]{51A478}97.3 & \cellcolor[HTML]{4FA177}97.7 & \cellcolor[HTML]{F38D74}55.9 \\
\texttt{zho} & Chinese & \cellcolor[HTML]{64BA83}94.1 & \cellcolor[HTML]{52A77A}\textbf{97.1} & \cellcolor[HTML]{A1D790}88.2 & \cellcolor[HTML]{A1D790}88.2 & \cellcolor[HTML]{84C98B}91.2 & \cellcolor[HTML]{84C98B}91.2 & \cellcolor[HTML]{52A77A}\textbf{97.1} & \cellcolor[HTML]{A1D790}88.2 & \cellcolor[HTML]{64BA83}94.1 & \cellcolor[HTML]{52A77A}\textbf{97.1} & \cellcolor[HTML]{64BA83}94.1 & \cellcolor[HTML]{A1D790}88.2 & \cellcolor[HTML]{75C288}\textbf{92.4} & \cellcolor[HTML]{FFF9C7}73.5 & \cellcolor[HTML]{A1D790}88.2 & \cellcolor[HTML]{FFF9C7}73.5 \\
\texttt{hrv} & Croatian & \cellcolor[HTML]{5FB782}94.4 & \cellcolor[HTML]{57AF7E}95.8 & \cellcolor[HTML]{C3E596}84.5 & \cellcolor[HTML]{71C087}93.0 & \cellcolor[HTML]{57AF7E}95.8 & \cellcolor[HTML]{4D9C74}98.6 & \cellcolor[HTML]{47926F}\textbf{100.0} & \cellcolor[HTML]{A8DA91}87.3 & \cellcolor[HTML]{B7E093}85.9 & \cellcolor[HTML]{9CD490}88.7 & \cellcolor[HTML]{52A579}97.2 & \cellcolor[HTML]{47926F}\textbf{100.0} & \cellcolor[HTML]{6ABD85}\textbf{93.4} & \cellcolor[HTML]{9CD490}88.7 & \cellcolor[HTML]{57AF7E}95.8 & \cellcolor[HTML]{FCBC89}62.0 \\
\texttt{ces} & Czech & \cellcolor[HTML]{C0E494}84.8 & \cellcolor[HTML]{64BA83}94.1 & \cellcolor[HTML]{5FB782}94.5 & \cellcolor[HTML]{ECF7B7}78.4 & \cellcolor[HTML]{6FBF86}93.0 & \cellcolor[HTML]{5AB481}95.1 & \cellcolor[HTML]{52A579}97.2 & \cellcolor[HTML]{E9F6B3}79.1 & \cellcolor[HTML]{C5E698}84.2 & \cellcolor[HTML]{93D08F}89.7 & \cellcolor[HTML]{59B380}95.3 & \cellcolor[HTML]{4FA177}\textbf{97.8} & \cellcolor[HTML]{8DCD8E}\textbf{90.3} & \cellcolor[HTML]{FFFDCD}74.5 & \cellcolor[HTML]{75C288}92.4 & \cellcolor[HTML]{E87769}52.8 \\
\texttt{dan} & Danish & \cellcolor[HTML]{86CA8C}90.9 & \cellcolor[HTML]{73C187}92.7 & \cellcolor[HTML]{AADB91}87.3 & \cellcolor[HTML]{73C187}92.7 & \cellcolor[HTML]{73C187}92.7 & \cellcolor[HTML]{47926F}\textbf{100.0} & \cellcolor[HTML]{47926F}\textbf{100.0} & \cellcolor[HTML]{AADB91}87.3 & \cellcolor[HTML]{5DB682}94.5 & \cellcolor[HTML]{86CA8C}90.9 & \cellcolor[HTML]{5DB682}94.5 & \cellcolor[HTML]{55AB7C}96.4 & \cellcolor[HTML]{6CBE86}\textbf{93.3} & \cellcolor[HTML]{73C187}92.7 & \cellcolor[HTML]{73C187}92.7 & \cellcolor[HTML]{E4F3AB}80.0 \\
\texttt{nld} & Dutch & \cellcolor[HTML]{A5D891}87.8 & \cellcolor[HTML]{5BB581}94.9 & \cellcolor[HTML]{80C78A}91.6 & \cellcolor[HTML]{BAE293}85.5 & \cellcolor[HTML]{8ACC8D}90.5 & \cellcolor[HTML]{64BA83}94.1 & \cellcolor[HTML]{57AF7E}\textbf{95.8} & \cellcolor[HTML]{FBFDCC}75.7 & \cellcolor[HTML]{D8EFA4}81.5 & \cellcolor[HTML]{C1E595}84.7 & \cellcolor[HTML]{93D08F}89.7 & \cellcolor[HTML]{64BA83}94.1 & \cellcolor[HTML]{9CD490}\textbf{88.8} & \cellcolor[HTML]{FEDBA0}66.6 & \cellcolor[HTML]{B0DD92}86.6 & \cellcolor[HTML]{FAAE83}60.2 \\
\texttt{eng} & English & \cellcolor[HTML]{4A9872}\textbf{99.1} & \cellcolor[HTML]{4A9872}\textbf{99.1} & \cellcolor[HTML]{4D9C74}98.6 & \cellcolor[HTML]{4E9E75}98.1 & \cellcolor[HTML]{4D9C74}98.6 & \cellcolor[HTML]{4E9E75}98.1 & \cellcolor[HTML]{4A9872}\textbf{99.1} & \cellcolor[HTML]{4E9E75}98.1 & \cellcolor[HTML]{4D9C74}98.6 & \cellcolor[HTML]{4E9E75}98.1 & \cellcolor[HTML]{4D9C74}98.6 & \cellcolor[HTML]{4E9E75}98.1 & \cellcolor[HTML]{4D9C74}\textbf{98.5} & \cellcolor[HTML]{B3DF92}86.3 & \cellcolor[HTML]{71C087}92.9 & \cellcolor[HTML]{FDBE8A}62.3 \\
\texttt{fin} & Finnish & \cellcolor[HTML]{77C388}92.2 & \cellcolor[HTML]{5AB481}95.0 & \cellcolor[HTML]{D1EB9F}82.5 & \cellcolor[HTML]{82C88B}91.2 & \cellcolor[HTML]{6FBF86}93.0 & \cellcolor[HTML]{53A87A}96.8 & \cellcolor[HTML]{5AB481}95.0 & \cellcolor[HTML]{E6F4AE}79.5 & \cellcolor[HTML]{C4E697}84.2 & \cellcolor[HTML]{A1D790}88.2 & \cellcolor[HTML]{56AD7D}96.0 & \cellcolor[HTML]{52A77A}\textbf{97.0} & \cellcolor[HTML]{86CA8C}\textbf{90.9} & \cellcolor[HTML]{59B380}95.2 & \cellcolor[HTML]{4C9A73}98.8 & \cellcolor[HTML]{94D18F}89.5 \\
\texttt{deu} & German & \cellcolor[HTML]{4FA076}97.9 & \cellcolor[HTML]{4B9973}98.8 & \cellcolor[HTML]{4D9C74}98.6 & \cellcolor[HTML]{58B07F}95.6 & \cellcolor[HTML]{4FA177}97.7 & \cellcolor[HTML]{4D9C74}98.6 & \cellcolor[HTML]{4A9872}99.1 & \cellcolor[HTML]{51A478}97.4 & \cellcolor[HTML]{4A9872}99.1 & \cellcolor[HTML]{4FA177}97.7 & \cellcolor[HTML]{4D9C74}98.6 & \cellcolor[HTML]{499570}\textbf{99.5} & \cellcolor[HTML]{4E9E75}\textbf{98.2} & \cellcolor[HTML]{9DD590}88.6 & \cellcolor[HTML]{5BB581}94.9 & \cellcolor[HTML]{CAE89B}83.4 \\
\texttt{ell} & Greek & \cellcolor[HTML]{E7F5B0}79.3 & \cellcolor[HTML]{7BC589}92.0 & \cellcolor[HTML]{54A97B}\textbf{96.6} & \cellcolor[HTML]{9DD590}88.5 & \cellcolor[HTML]{B3DF92}86.2 & \cellcolor[HTML]{6FBF86}93.1 & \cellcolor[HTML]{54A97B}\textbf{96.6} & \cellcolor[HTML]{FFFECF}74.7 & \cellcolor[HTML]{D8EFA4}81.6 & \cellcolor[HTML]{B3DF92}86.2 & \cellcolor[HTML]{6FBF86}93.1 & \cellcolor[HTML]{54A97B}\textbf{96.6} & \cellcolor[HTML]{9CD490}\textbf{88.7} & \cellcolor[HTML]{B3DF92}86.2 & \cellcolor[HTML]{6FBF86}93.1 & \cellcolor[HTML]{F89A79}57.5 \\
\texttt{heb} & Hebrew & \cellcolor[HTML]{88CB8D}90.6 & \cellcolor[HTML]{7EC68A}91.7 & \cellcolor[HTML]{5BB581}94.8 & \cellcolor[HTML]{D2ECA0}82.3 & \cellcolor[HTML]{88CB8D}90.6 & \cellcolor[HTML]{7EC68A}91.7 & \cellcolor[HTML]{5BB581}94.8 & \cellcolor[HTML]{B1DE92}86.5 & \cellcolor[HTML]{BAE293}85.4 & \cellcolor[HTML]{E8F5B1}79.2 & \cellcolor[HTML]{57AF7E}\textbf{95.8} & \cellcolor[HTML]{7EC68A}91.7 & \cellcolor[HTML]{94D18F}\textbf{89.6} & \cellcolor[HTML]{CBE99C}83.3 & \cellcolor[HTML]{73C187}92.7 & \cellcolor[HTML]{F59175}56.2 \\
\texttt{hin} & Hindi & \cellcolor[HTML]{5FB782}94.4 & \cellcolor[HTML]{4A9872}\textbf{99.1} & \cellcolor[HTML]{4E9E75}98.1 & \cellcolor[HTML]{75C288}92.5 & \cellcolor[HTML]{55AC7C}96.3 & \cellcolor[HTML]{4A9872}\textbf{99.1} & \cellcolor[HTML]{4E9E75}98.1 & \cellcolor[HTML]{5FB782}94.4 & \cellcolor[HTML]{75C288}92.5 & \cellcolor[HTML]{58B17F}95.3 & \cellcolor[HTML]{6ABD85}93.5 & \cellcolor[HTML]{55AC7C}96.3 & \cellcolor[HTML]{57AF7E}\textbf{95.8} & \cellcolor[HTML]{E1F2AA}80.4 & \cellcolor[HTML]{80C78A}91.6 & \cellcolor[HTML]{D35C63}48.6 \\
\texttt{hun} & Hungarian & \cellcolor[HTML]{77C388}92.3 & \cellcolor[HTML]{47926F}\textbf{100.0} & \cellcolor[HTML]{C1E595}84.6 & \cellcolor[HTML]{C1E595}84.6 & \cellcolor[HTML]{C1E595}84.6 & \cellcolor[HTML]{47926F}\textbf{100.0} & \cellcolor[HTML]{47926F}\textbf{100.0} & \cellcolor[HTML]{77C388}92.3 & \cellcolor[HTML]{77C388}92.3 & \cellcolor[HTML]{C1E595}84.6 & \cellcolor[HTML]{47926F}\textbf{100.0} & \cellcolor[HTML]{47926F}\textbf{100.0} & \cellcolor[HTML]{71C087}\textbf{92.9} & \cellcolor[HTML]{47926F}100.0 & \cellcolor[HTML]{47926F}100.0 & \cellcolor[HTML]{77C388}92.3 \\
\texttt{isl} & Icelandic & \cellcolor[HTML]{FEE3A7}68.1 & \cellcolor[HTML]{BEE394}85.1 & \cellcolor[HTML]{E9796A}53.2 & \cellcolor[HTML]{FEE3A7}68.1 & \cellcolor[HTML]{80C78A}91.5 & \cellcolor[HTML]{80C78A}91.5 & \cellcolor[HTML]{68BC84}\textbf{93.6} & \cellcolor[HTML]{FED69C}66.0 & \cellcolor[HTML]{F89A79}57.4 & \cellcolor[HTML]{F6FBC4}76.6 & \cellcolor[HTML]{AADB91}87.2 & \cellcolor[HTML]{68BC84}\textbf{93.6} & \cellcolor[HTML]{F1F9BD}\textbf{77.7} & \cellcolor[HTML]{F6FBC4}76.6 & \cellcolor[HTML]{80C78A}91.5 & \cellcolor[HTML]{E9796A}53.2 \\
\texttt{ind} & Indonesian & \cellcolor[HTML]{CBE99C}83.3 & \cellcolor[HTML]{47926F}\textbf{100.0} & \cellcolor[HTML]{47926F}\textbf{100.0} & \cellcolor[HTML]{47926F}\textbf{100.0} & \cellcolor[HTML]{47926F}\textbf{100.0} & \cellcolor[HTML]{47926F}\textbf{100.0} & \cellcolor[HTML]{47926F}\textbf{100.0} & \cellcolor[HTML]{47926F}\textbf{100.0} & \cellcolor[HTML]{47926F}\textbf{100.0} & \cellcolor[HTML]{47926F}\textbf{100.0} & \cellcolor[HTML]{CBE99C}83.3 & \cellcolor[HTML]{47926F}\textbf{100.0} & \cellcolor[HTML]{52A579}\textbf{97.2} & \cellcolor[HTML]{47926F}100.0 & \cellcolor[HTML]{47926F}100.0 & \cellcolor[HTML]{DC6463}50.0 \\
\texttt{lit} & Lithuanian & \cellcolor[HTML]{ACDB92}87.0 & \cellcolor[HTML]{6ABD85}93.5 & \cellcolor[HTML]{6ABD85}93.5 & \cellcolor[HTML]{EDF8B9}78.3 & \cellcolor[HTML]{82C88B}91.3 & \cellcolor[HTML]{82C88B}91.3 & \cellcolor[HTML]{4FA177}\textbf{97.8} & \cellcolor[HTML]{ACDB92}87.0 & \cellcolor[HTML]{ACDB92}87.0 & \cellcolor[HTML]{4FA177}\textbf{97.8} & \cellcolor[HTML]{6ABD85}93.5 & \cellcolor[HTML]{4FA177}\textbf{97.8} & \cellcolor[HTML]{82C88B}\textbf{91.3} & \cellcolor[HTML]{6ABD85}93.5 & \cellcolor[HTML]{58B07F}95.7 & \cellcolor[HTML]{98D38F}89.1 \\
\texttt{nob} & Norwegian-Bokmaal & \cellcolor[HTML]{C8E89A}83.7 & \cellcolor[HTML]{88CB8D}90.7 & \cellcolor[HTML]{B5DF93}86.0 & \cellcolor[HTML]{C8E89A}83.7 & \cellcolor[HTML]{B5DF93}86.0 & \cellcolor[HTML]{4FA177}\textbf{97.7} & \cellcolor[HTML]{4FA177}\textbf{97.7} & \cellcolor[HTML]{FEEBB0}69.8 & \cellcolor[HTML]{DAEFA5}81.4 & \cellcolor[HTML]{9FD690}88.4 & \cellcolor[HTML]{DAEFA5}81.4 & \cellcolor[HTML]{4FA177}\textbf{97.7} & \cellcolor[HTML]{ACDB92}\textbf{87.0} & \cellcolor[HTML]{C8E89A}83.7 & \cellcolor[HTML]{6FBF86}93.0 & \cellcolor[HTML]{BE4763}32.6 \\
\texttt{nno} & Norwegian-Nynorsk & \cellcolor[HTML]{FFF8C6}73.3 & \cellcolor[HTML]{CBE99C}83.3 & \cellcolor[HTML]{FEC78F}63.3 & \cellcolor[HTML]{F6FBC4}76.7 & \cellcolor[HTML]{91CF8F}90.0 & \cellcolor[HTML]{6CBE86}93.3 & \cellcolor[HTML]{54A97B}\textbf{96.7} & \cellcolor[HTML]{FEECB1}70.0 & \cellcolor[HTML]{F6FBC4}76.7 & \cellcolor[HTML]{B0DD92}86.7 & \cellcolor[HTML]{6CBE86}93.3 & \cellcolor[HTML]{6CBE86}93.3 & \cellcolor[HTML]{CDEA9D}\textbf{83.1} & \cellcolor[HTML]{E4F3AB}80.0 & --- & \cellcolor[HTML]{FAAC82}60.0 \\
\texttt{pes} & Persian & \cellcolor[HTML]{54A97B}96.7 & \cellcolor[HTML]{4C9A73}98.7 & \cellcolor[HTML]{51A478}97.3 & \cellcolor[HTML]{88CB8D}90.6 & \cellcolor[HTML]{54A97B}96.7 & \cellcolor[HTML]{4E9E75}98.2 & \cellcolor[HTML]{4A9872}\textbf{99.1} & \cellcolor[HTML]{D2ECA0}82.4 & \cellcolor[HTML]{9FD690}88.4 & \cellcolor[HTML]{75C288}92.4 & \cellcolor[HTML]{4FA076}98.0 & \cellcolor[HTML]{4B9973}98.9 & \cellcolor[HTML]{5BB581}\textbf{94.8} & \cellcolor[HTML]{96D28F}89.3 & \cellcolor[HTML]{50A377}97.6 & \cellcolor[HTML]{BE4763}34.1 \\
\texttt{pol} & Polish & \cellcolor[HTML]{5BB581}94.8 & \cellcolor[HTML]{5BB581}94.9 & \cellcolor[HTML]{56AD7D}95.9 & \cellcolor[HTML]{5BB581}94.8 & \cellcolor[HTML]{4C9A73}98.7 & \cellcolor[HTML]{4C9A73}98.7 & \cellcolor[HTML]{499570}\textbf{99.6} & \cellcolor[HTML]{77C388}92.3 & \cellcolor[HTML]{5AB481}95.0 & \cellcolor[HTML]{58B17F}95.4 & \cellcolor[HTML]{53A87A}96.9 & \cellcolor[HTML]{4FA076}97.9 & \cellcolor[HTML]{55AC7C}\textbf{96.2} & \cellcolor[HTML]{68BC84}93.6 & \cellcolor[HTML]{4FA076}98.0 & \cellcolor[HTML]{FFFCCC}74.1 \\
\texttt{ron} & Romanian & \cellcolor[HTML]{B0DD92}86.6 & \cellcolor[HTML]{6ABD85}93.5 & \cellcolor[HTML]{62B883}94.3 & \cellcolor[HTML]{E2F3AB}80.1 & \cellcolor[HTML]{62B883}94.3 & \cellcolor[HTML]{56AD7D}\textbf{95.9} & \cellcolor[HTML]{58B07F}95.5 & \cellcolor[HTML]{E2F3AB}80.1 & \cellcolor[HTML]{88CB8D}90.7 & \cellcolor[HTML]{8DCD8E}90.2 & \cellcolor[HTML]{56AD7D}\textbf{95.9} & \cellcolor[HTML]{59B380}95.1 & \cellcolor[HTML]{84C98B}\textbf{91.0} & \cellcolor[HTML]{A1D790}88.2 & \cellcolor[HTML]{50A377}97.6 & \cellcolor[HTML]{BE4763}45.1 \\
\texttt{rus} & Russian & \cellcolor[HTML]{4D9D75}98.4 & \cellcolor[HTML]{4A9671}99.3 & \cellcolor[HTML]{4C9A73}98.7 & \cellcolor[HTML]{4D9D75}98.3 & \cellcolor[HTML]{4B9973}98.9 & \cellcolor[HTML]{4A9671}99.3 & \cellcolor[HTML]{499570}\textbf{99.6} & \cellcolor[HTML]{50A377}97.6 & \cellcolor[HTML]{50A377}97.5 & \cellcolor[HTML]{4FA076}98.0 & \cellcolor[HTML]{4A9872}99.1 & \cellcolor[HTML]{499570}99.6 & \cellcolor[HTML]{4C9A73}\textbf{98.7} & \cellcolor[HTML]{6CBE86}93.2 & \cellcolor[HTML]{51A478}97.3 & \cellcolor[HTML]{EDF8B9}78.2 \\
\texttt{srp} & Serbian & \cellcolor[HTML]{E8F5B1}79.2 & \cellcolor[HTML]{7EC68A}91.7 & \cellcolor[HTML]{FEFFD0}75.0 & \cellcolor[HTML]{CBE99C}83.3 & \cellcolor[HTML]{7EC68A}91.7 & \cellcolor[HTML]{47926F}\textbf{100.0} & \cellcolor[HTML]{47926F}\textbf{100.0} & \cellcolor[HTML]{FEFFD0}75.0 & \cellcolor[HTML]{CBE99C}83.3 & \cellcolor[HTML]{57AF7E}95.8 & \cellcolor[HTML]{57AF7E}95.8 & \cellcolor[HTML]{47926F}\textbf{100.0} & \cellcolor[HTML]{98D38F}\textbf{89.2} & \cellcolor[HTML]{CBE99C}83.3 & \cellcolor[HTML]{A7D991}87.5 & \cellcolor[HTML]{FEFFD0}75.0 \\
\texttt{slk} & Slovak & \cellcolor[HTML]{FEE3A7}68.0 & \cellcolor[HTML]{C4E697}84.4 & \cellcolor[HTML]{CFEB9E}82.7 & \cellcolor[HTML]{FFFFD0}74.8 & \cellcolor[HTML]{A8DA91}87.4 & \cellcolor[HTML]{80C78A}91.5 & \cellcolor[HTML]{58B07F}\textbf{95.6} & \cellcolor[HTML]{F7FCC6}76.5 & \cellcolor[HTML]{E9F6B3}78.9 & \cellcolor[HTML]{CEEA9E}83.0 & \cellcolor[HTML]{6ABD85}93.5 & \cellcolor[HTML]{5BB581}94.9 & \cellcolor[HTML]{C4E697}\textbf{84.3} & \cellcolor[HTML]{F3FAC0}77.2 & \cellcolor[HTML]{55AC7C}96.3 & \cellcolor[HTML]{DC6463}50.0 \\
\texttt{slv} & Slovenian & \cellcolor[HTML]{E5F4AD}79.8 & \cellcolor[HTML]{AEDC92}86.9 & \cellcolor[HTML]{EEF8BA}78.0 & \cellcolor[HTML]{DCF0A7}81.0 & \cellcolor[HTML]{77C388}92.3 & \cellcolor[HTML]{59B380}95.2 & \cellcolor[HTML]{4E9E75}\textbf{98.2} & \cellcolor[HTML]{E1F2AA}80.4 & \cellcolor[HTML]{B9E193}85.7 & \cellcolor[HTML]{A7D991}87.5 & \cellcolor[HTML]{7EC68A}91.7 & \cellcolor[HTML]{55AB7C}96.4 & \cellcolor[HTML]{A5D891}\textbf{87.7} & \cellcolor[HTML]{AEDC92}86.9 & \cellcolor[HTML]{50A377}97.6 & \cellcolor[HTML]{FFF0B7}70.8 \\
\texttt{spa} & Spanish & \cellcolor[HTML]{6FBF86}93.0 & \cellcolor[HTML]{4FA177}97.8 & \cellcolor[HTML]{4E9E75}\textbf{98.2} & \cellcolor[HTML]{58B07F}95.6 & \cellcolor[HTML]{55AB7C}96.3 & \cellcolor[HTML]{4E9E75}\textbf{98.2} & \cellcolor[HTML]{4E9E75}\textbf{98.2} & \cellcolor[HTML]{59B380}95.2 & \cellcolor[HTML]{5BB581}94.9 & \cellcolor[HTML]{66BB84}93.8 & \cellcolor[HTML]{4FA177}97.8 & \cellcolor[HTML]{59B380}95.2 & \cellcolor[HTML]{55AC7C}\textbf{96.2} & \cellcolor[HTML]{AADB91}87.1 & \cellcolor[HTML]{58B07F}95.6 & \cellcolor[HTML]{E57066}51.8 \\
\texttt{swe} & Swedish & \cellcolor[HTML]{54A97B}96.6 & \cellcolor[HTML]{47926F}\textbf{100.0} & \cellcolor[HTML]{A1D790}88.1 & \cellcolor[HTML]{5BB581}94.9 & \cellcolor[HTML]{4D9D75}98.3 & \cellcolor[HTML]{4D9D75}98.3 & \cellcolor[HTML]{4D9D75}98.3 & \cellcolor[HTML]{B1DE92}86.4 & \cellcolor[HTML]{5BB581}94.9 & \cellcolor[HTML]{6CBE86}93.2 & \cellcolor[HTML]{54A97B}96.6 & \cellcolor[HTML]{54A97B}96.6 & \cellcolor[HTML]{59B380}\textbf{95.2} & \cellcolor[HTML]{5BB581}94.9 & \cellcolor[HTML]{6CBE86}93.2 & \cellcolor[HTML]{FFF1B9}71.2 \\
\texttt{tur} & Turkish & \cellcolor[HTML]{79C489}92.1 & \cellcolor[HTML]{55AC7C}96.2 & \cellcolor[HTML]{55AC7C}96.2 & \cellcolor[HTML]{71C087}92.9 & \cellcolor[HTML]{58B17F}95.3 & \cellcolor[HTML]{4FA177}97.7 & \cellcolor[HTML]{4FA177}\textbf{97.8} & \cellcolor[HTML]{A3D791}87.9 & \cellcolor[HTML]{71C087}92.9 & \cellcolor[HTML]{73C187}92.6 & \cellcolor[HTML]{55AB7C}96.3 & \cellcolor[HTML]{52A579}97.2 & \cellcolor[HTML]{5DB682}\textbf{94.6} & \cellcolor[HTML]{FDC28C}62.7 & \cellcolor[HTML]{6CBE86}93.2 & \cellcolor[HTML]{BE4763}40.7 \\
\texttt{ukr} & Ukrainian & \cellcolor[HTML]{56AD7D}95.9 & \cellcolor[HTML]{4FA076}98.0 & \cellcolor[HTML]{54A97B}96.6 & \cellcolor[HTML]{68BC84}93.6 & \cellcolor[HTML]{52A77A}96.9 & \cellcolor[HTML]{4FA076}98.0 & \cellcolor[HTML]{4A9671}\textbf{99.3} & \cellcolor[HTML]{7BC589}91.9 & \cellcolor[HTML]{75C288}92.5 & \cellcolor[HTML]{6CBE86}93.2 & \cellcolor[HTML]{54A97B}96.6 & \cellcolor[HTML]{50A377}97.6 & \cellcolor[HTML]{57AF7E}\textbf{95.8} & \cellcolor[HTML]{A1D790}88.1 & \cellcolor[HTML]{54A97B}96.6 & \cellcolor[HTML]{FEE2A6}67.8 \\
\texttt{urd} & Urdu & \cellcolor[HTML]{91CF8F}90.0 & \cellcolor[HTML]{4D9D75}98.3 & \cellcolor[HTML]{6CBE86}93.3 & \cellcolor[HTML]{91CF8F}90.0 & \cellcolor[HTML]{7EC68A}91.7 & \cellcolor[HTML]{47926F}\textbf{100.0} & \cellcolor[HTML]{54A97B}96.7 & \cellcolor[HTML]{9FD690}88.3 & \cellcolor[HTML]{6CBE86}93.3 & \cellcolor[HTML]{5AB481}95.0 & \cellcolor[HTML]{5AB481}95.0 & \cellcolor[HTML]{4D9D75}98.3 & \cellcolor[HTML]{62B883}\textbf{94.2} & \cellcolor[HTML]{6CBE86}93.3 & \cellcolor[HTML]{5AB481}95.0 & \cellcolor[HTML]{EA7A6B}53.3 \\
\midrule
 & \textbf{Mean} & \cellcolor[HTML]{9DD590}88.5 & \cellcolor[HTML]{5DB682}94.6 & \cellcolor[HTML]{9AD390}89.0 & \cellcolor[HTML]{A5D891}87.9 & \cellcolor[HTML]{6ABD85}93.5 & \cellcolor[HTML]{54A97B}96.5 & \cellcolor[HTML]{4FA076}\textbf{97.9} & \cellcolor[HTML]{BAE293}85.4 & \cellcolor[HTML]{9CD490}88.7 & \cellcolor[HTML]{86CA8C}90.9 & \cellcolor[HTML]{5FB782}94.4 & \cellcolor[HTML]{53A87A}96.8 & \cellcolor[HTML]{79C489}\textbf{92.0} & \cellcolor[HTML]{B1DE92}86.5 & \cellcolor[HTML]{5BB581}94.8 & \cellcolor[HTML]{FEC58E}63.2 \\
\bottomrule
\end{tabular}
}
\caption{Accuracy (\%) by language and model over 12,866 minimal pairs in 33 languages. \emph{Mean} averages the multilingual models to its left; the columns to its right are reference points rather than systems under comparison and are excluded from it, as is the bolding. Best multilingual model per language in bold; \textemdash{} marks a language the model does not cover. The bottom row is the unweighted average across languages.}
\label{tab:acc-full}
\end{table*}

Following previous work~\citep[e.g.][]{warstadtBLiMPBenchmarkLinguistic2020,jumeletMultiBLiMP10Massively2026}, we use perplexity on the sentence level to evaluate the models' preference within each minimal pair. We report accuracy scores, which reflect the number of sentences for which the model assigns a higher probability (i.e., lower perplexity) to the correct sentence.


As shown in \autoref{tab:acc-full}, we evaluate n-grams and monolingual models as baseline. We train the n-gram models ourselves: training details are given in \autoref{appendix:ngrams}. For multilingual evaluation we choose 5 recent open-weight model families in 1-70B range (details in \autoref{app:transformers}). We find that Gemma and Apertus language families outperform the other models, with Gemma-3-27B being the best model. Interestingly, the 8B and 14B variants of Llama and Qwen perform on par with 100MB Goldfish models, the smallest neural models in our sample.

Scores generally cluster by the model family rather than model size. The underlying reason may be the data distribution, but the reporting practices for training data distribution are subpar \citep{xu2025survey}. The best performing model families, Gemma and Apertus, prioritize multilingual coverage in pretraining, and Apertus further reports that roughly 40\% of its pretraining data is non-English \cite{apertusApertusDemocratizingOpen2025}.

All multilingual models exceed 98\% on English, although the English subset is not easy: the $n$-gram scores 62.3 and Goldfish 86/93. As such, English is the one language where multilingual models hold a clear advantage, likely reflecting English-dominant pretraining \citep{xu2025survey}. Models perform worst on Icelandic (77), Norwegian Nynorsk (83), Slovak (84) and Norwegian Bokmål (87), all mid to low resource \citep{joshi-etal-2020-state}.






\section{Analysis}

\begin{figure*}
    \centering
    \includegraphics[width=\textwidth]{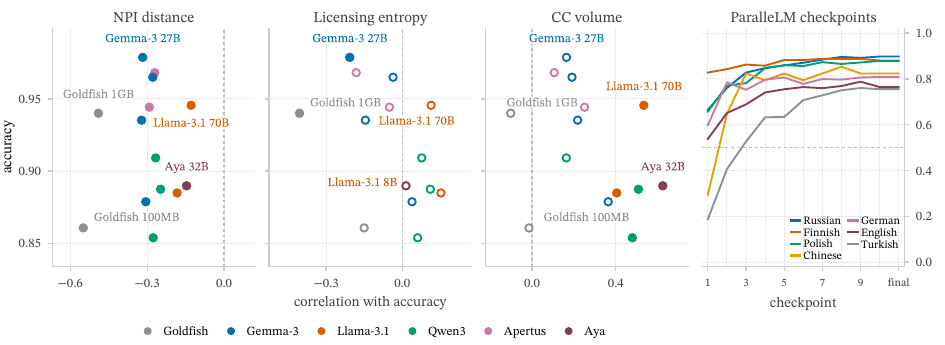}
    \caption{Model sensitivity to three benchmark properties, and ParalleLM learning curves. In the first three panels, each point is one model: $x$ is the correlation between the property and accuracy, $y$ is overall accuracy. Filled points are significant. Correlations: distance, Pearson over pairs; licensing entropy, Pearson over lemmas; CC volume, Spearman over languages.}
    \label{fig:corr_strips}
\end{figure*}

We examine how different data and language-intrinsic factors affect model performance. We first look at the categorical separations in \autoref{tab:acc-subsets}, then continuous factors (\autoref{fig:corr_strips}). Lastly, we examine the NPI's learning trajectory using ParaleLM \citep{yangAutoregressiveTransformerModels2026} checkpoints. 


\begin{table*}[t]
\centering
\small
\setlength{\tabcolsep}{3.2pt}
\resizebox{\textwidth}{!}{%
\begin{tabular}{ll r r | cccccccccccc | c | ccc}
\toprule
& & & & \multicolumn{2}{c}{Llama-3.1} & \multicolumn{1}{c}{Aya} & \multicolumn{4}{c}{Gemma-3} & \multicolumn{3}{c}{Qwen3} & \multicolumn{2}{c}{Apertus} &  & \multicolumn{2}{c}{Goldfish} & \multicolumn{1}{c}{Other} \\
\cmidrule(lr){5-6} \cmidrule(lr){7-7} \cmidrule(lr){8-11} \cmidrule(lr){12-14} \cmidrule(lr){15-16} \cmidrule(lr){18-19} \cmidrule(lr){20-20}
& Subset & Langs & Pairs & \texttt{8B} & \texttt{70B} & \texttt{32B} & \texttt{1B} & \texttt{4B} & \texttt{12B} & \texttt{27B} & \texttt{8B} & \texttt{14B} & \texttt{32B} & \texttt{8B} & \texttt{70B} & \textbf{Mean} & \texttt{100MB} & \texttt{1GB} & \texttt{5-gram} \\
\midrule
\multicolumn{2}{l}{\textbf{All}} & 33 & 12,866 & \cellcolor[HTML]{9DD590}88.5 & \cellcolor[HTML]{5DB682}94.6 & \cellcolor[HTML]{9AD390}89.0 & \cellcolor[HTML]{A5D891}87.9 & \cellcolor[HTML]{6ABD85}93.5 & \cellcolor[HTML]{54A97B}96.5 & \cellcolor[HTML]{4FA076}\textbf{97.9} & \cellcolor[HTML]{BAE293}85.4 & \cellcolor[HTML]{9CD490}88.7 & \cellcolor[HTML]{86CA8C}90.9 & \cellcolor[HTML]{5FB782}94.4 & \cellcolor[HTML]{53A87A}96.8 & \cellcolor[HTML]{79C489}\textbf{92.0} & \cellcolor[HTML]{B1DE92}86.5 & \cellcolor[HTML]{5BB581}94.8 & \cellcolor[HTML]{FEC58E}63.2 \\
\midrule
\multirow{2}{*}{Typology} & NC & 21 & 10,014 & \cellcolor[HTML]{9DD590}88.6 & \cellcolor[HTML]{5FB782}94.3 & \cellcolor[HTML]{91CF8F}90.0 & \cellcolor[HTML]{A8DA91}87.3 & \cellcolor[HTML]{6ABD85}93.4 & \cellcolor[HTML]{55AB7C}96.3 & \cellcolor[HTML]{4FA076}\textbf{98.0} & \cellcolor[HTML]{BAE293}85.5 & \cellcolor[HTML]{9CD490}88.8 & \cellcolor[HTML]{8DCD8E}90.3 & \cellcolor[HTML]{57AF7E}95.7 & \cellcolor[HTML]{51A478}97.3 & \cellcolor[HTML]{79C489}\textbf{92.1} & \cellcolor[HTML]{A7D991}87.5 & \cellcolor[HTML]{57AF7E}95.8 & \cellcolor[HTML]{FECE95}64.6 \\
 & NPI & 12 & 2,852 & \cellcolor[HTML]{A1D790}88.3 & \cellcolor[HTML]{5AB481}94.9 & \cellcolor[HTML]{AADB91}87.2 & \cellcolor[HTML]{9CD490}88.8 & \cellcolor[HTML]{68BC84}93.7 & \cellcolor[HTML]{53A87A}96.8 & \cellcolor[HTML]{4FA177}\textbf{97.7} & \cellcolor[HTML]{BEE394}85.1 & \cellcolor[HTML]{9CD490}88.7 & \cellcolor[HTML]{7BC589}92.0 & \cellcolor[HTML]{79C489}92.2 & \cellcolor[HTML]{56AD7D}96.0 & \cellcolor[HTML]{7EC68A}\textbf{91.8} & \cellcolor[HTML]{C1E595}84.7 & \cellcolor[HTML]{73C187}92.7 & \cellcolor[HTML]{FBB184}60.7 \\
\addlinespace[2pt]
\multirow{5}{*}{Family} & Slavic & 10 & 7,330 & \cellcolor[HTML]{B3DF92}86.3 & \cellcolor[HTML]{6CBE86}93.3 & \cellcolor[HTML]{B0DD92}86.5 & \cellcolor[HTML]{A8DA91}87.5 & \cellcolor[HTML]{5FB782}94.3 & \cellcolor[HTML]{53A87A}96.8 & \cellcolor[HTML]{4C9A73}\textbf{98.7} & \cellcolor[HTML]{BEE394}85.1 & \cellcolor[HTML]{9FD690}88.3 & \cellcolor[HTML]{84C98B}91.2 & \cellcolor[HTML]{56AD7D}96.0 & \cellcolor[HTML]{4FA076}98.0 & \cellcolor[HTML]{7BC589}\textbf{91.8} & \cellcolor[HTML]{AEDC92}86.8 & \cellcolor[HTML]{58B07F}95.6 & \cellcolor[HTML]{FEE2A6}67.9 \\
 & Germanic & 8 & 2,645 & \cellcolor[HTML]{AADB91}87.2 & \cellcolor[HTML]{6FBF86}93.1 & \cellcolor[HTML]{CBE99C}83.3 & \cellcolor[HTML]{ACDB92}86.9 & \cellcolor[HTML]{6FBF86}93.2 & \cellcolor[HTML]{55AB7C}96.5 & \cellcolor[HTML]{50A377}\textbf{97.5} & \cellcolor[HTML]{DAEFA5}81.3 & \cellcolor[HTML]{BAE293}85.5 & \cellcolor[HTML]{94D18F}89.5 & \cellcolor[HTML]{75C288}92.5 & \cellcolor[HTML]{55AC7C}96.2 & \cellcolor[HTML]{8FCE8E}\textbf{90.2} & \cellcolor[HTML]{C8E89A}83.7 & \cellcolor[HTML]{79C489}92.1 & \cellcolor[HTML]{FDC48D}62.9 \\
 & Other (10) & 10 & 1,986 & \cellcolor[HTML]{8DCD8E}90.3 & \cellcolor[HTML]{58B07F}95.7 & \cellcolor[HTML]{71C087}92.9 & \cellcolor[HTML]{A1D790}88.2 & \cellcolor[HTML]{73C187}92.7 & \cellcolor[HTML]{58B07F}95.5 & \cellcolor[HTML]{50A377}\textbf{97.6} & \cellcolor[HTML]{AEDC92}86.9 & \cellcolor[HTML]{94D18F}89.6 & \cellcolor[HTML]{8DCD8E}90.4 & \cellcolor[HTML]{66BB84}93.8 & \cellcolor[HTML]{54A97B}96.5 & \cellcolor[HTML]{75C288}\textbf{92.5} & \cellcolor[HTML]{AADB91}87.2 & \cellcolor[HTML]{58B17F}95.4 & \cellcolor[HTML]{FECF96}64.7 \\
 & Romance & 3 & 738 & \cellcolor[HTML]{86CA8C}90.9 & \cellcolor[HTML]{55AC7C}96.2 & \cellcolor[HTML]{5DB682}94.6 & \cellcolor[HTML]{A1D790}88.3 & \cellcolor[HTML]{5FB782}94.5 & \cellcolor[HTML]{53A87A}96.8 & \cellcolor[HTML]{51A478}\textbf{97.4} & \cellcolor[HTML]{A1D790}88.1 & \cellcolor[HTML]{71C087}92.9 & \cellcolor[HTML]{73C187}92.7 & \cellcolor[HTML]{54A97B}96.5 & \cellcolor[HTML]{58B17F}95.4 & \cellcolor[HTML]{68BC84}\textbf{93.7} & \cellcolor[HTML]{86CA8C}90.9 & \cellcolor[HTML]{52A77A}97.0 & \cellcolor[HTML]{E16963}51.0 \\
 & Indo-Aryan & 2 & 167 & \cellcolor[HTML]{77C388}92.2 & \cellcolor[HTML]{4C9A73}98.7 & \cellcolor[HTML]{57AF7E}95.7 & \cellcolor[HTML]{82C88B}91.3 & \cellcolor[HTML]{64BA83}94.0 & \cellcolor[HTML]{499570}\textbf{99.5} & \cellcolor[HTML]{51A478}97.4 & \cellcolor[HTML]{82C88B}91.4 & \cellcolor[HTML]{71C087}92.9 & \cellcolor[HTML]{59B380}95.2 & \cellcolor[HTML]{62B883}94.2 & \cellcolor[HTML]{51A478}97.3 & \cellcolor[HTML]{5AB481}\textbf{95.0} & \cellcolor[HTML]{AEDC92}86.9 & \cellcolor[HTML]{6CBE86}93.3 & \cellcolor[HTML]{E16963}51.0 \\
\addlinespace[2pt]
\multirow{2}{*}{5-gram} & solvable & 33 & 8,456 & \cellcolor[HTML]{6FBF86}93.1 & \cellcolor[HTML]{52A77A}96.9 & \cellcolor[HTML]{86CA8C}91.0 & \cellcolor[HTML]{75C288}92.5 & \cellcolor[HTML]{55AC7C}96.1 & \cellcolor[HTML]{4FA076}97.9 & \cellcolor[HTML]{4B9973}\textbf{98.9} & \cellcolor[HTML]{9FD690}88.4 & \cellcolor[HTML]{7EC68A}91.8 & \cellcolor[HTML]{6ABD85}93.5 & \cellcolor[HTML]{55AC7C}96.3 & \cellcolor[HTML]{4FA076}97.9 & \cellcolor[HTML]{5FB782}\textbf{94.5} & \cellcolor[HTML]{6ABD85}93.5 & \cellcolor[HTML]{4D9C74}98.5 & \cellcolor[HTML]{47926F}100.0 \\
 & unsolvable & 33 & 4,356 & \cellcolor[HTML]{D8EFA4}81.5 & \cellcolor[HTML]{91CF8F}90.0 & \cellcolor[HTML]{B7E093}85.9 & \cellcolor[HTML]{E7F5B0}79.4 & \cellcolor[HTML]{B0DD92}86.6 & \cellcolor[HTML]{62B883}94.2 & \cellcolor[HTML]{55AC7C}\textbf{96.1} & \cellcolor[HTML]{DFF2A9}80.6 & \cellcolor[HTML]{C0E494}84.9 & \cellcolor[HTML]{ACDB92}87.0 & \cellcolor[HTML]{7EC68A}91.6 & \cellcolor[HTML]{5BB581}94.8 & \cellcolor[HTML]{A5D891}\textbf{87.7} & \cellcolor[HTML]{FFFAC9}73.6 & \cellcolor[HTML]{A5D891}87.8 & \cellcolor[HTML]{BE4763}0.0 \\
\bottomrule
\end{tabular}
}
\caption{Accuracy (\%) by model over subsets of \blimppi. Every cell is a mean over the languages in the subset, each counted once; \emph{Mean} averages the multilingual models to its left. Typology follows Table~\ref{tab:languages}. Rows are ordered by subset size. Best model per row in bold.}
\label{tab:acc-subsets}
\end{table*}

\paragraph{Typology} NCIs are largely restricted to negation, whereas NPIs are distributed more widely (Section~\ref{sec:background}). We therefore expect NC languages to be easier, and this holds modestly for the baselines, which score higher on NC languages by 2.8 (Goldfish 100MB), 3.1 (Goldfish 1000MB) and 3.9 points ($n$-gram). Among multilingual models the difference is only 0.3 points and inconsistent in direction; only Apertus scores lower on NPI languages at both sizes ($-3.5$ for 8B, $-1.3$ for 70B).

\paragraph{Language Family.}

We aggregate language-level scores by language family or branch. Germanic languages are the most challenging group; 7 out of 12 multilingual LMs and Goldfish models score lowest in this family.
Romance and Indo-Aryan groups, which have fewer members compared to others, acquire the overall highest scores. 

\paragraph{N-gram Solvability.} 
A minimal pair is n-gram solvable if the n-gram model correctly assigns higher probability to the grammatical sentence. We compare how models perform on the n-gram solvable portion of \blimppi, in comparison to the unsolvable portion. We find that n-gram solvability is the best oracle for predicting where LMs are likelier to err, and this effect appears to be model family independent. While the mean aggregated accuracy of multilingual models on n-gram solvable pairs is 94.5, the mean for n-gram unsolvable pairs is 87.7. Within every model family, the effect decreases strictly monotonically with size, from 13.1 accuracy points in the Gemma-1B to 3.1 for Apertus-70B and 2.8 for Gemma-27B, the highest-scoring models in our sample.

\paragraph{Distance.} We define distance as the number of tokens separating the NPI from its licenser in the grammatical sentence. We report the correlation of distance with model accuracy, at the minimal pair level. We find that accuracy decreases as the distance increases, as shown in \autoref{fig:corr_strips}. The negative correlation is significant for every evaluated model, but its magnitude decreases with model size. N-gram models also have -0.62 correlation with the distance, significant but not absolute. Appendix~\ref{app:distance} shows the language level breakdown.

\paragraph{Common Crawl Frequency.} We correlate language-level accuracy with Common Crawl frequencies (CC, \citealp{NEURIPS2024_1e605762}). The correlation separates models that report balanced multilingual training data (Gemma, Apertus) from those whose scores track CC significantly (Aya, Llama, Qwen), which also underperform on \blimppi. All models reach ceiling on high-resource languages such as English, Russian, German and Spanish; only monolingual and balanced multilingual models also do so on low-resource ones.


\paragraph{Licensing Entropy}

As described in section \ref{sec:background}, the \textit{strength hierarchy} of \citet{zwartsThreeTypesPolarity1998} sorts NPIs into categories by variability of their licensing environments. 
\citet{hoeksemaNaturalHistoryNegative} shows that individual NPIs pattern more idiosyncratically than the strength hierarchy (Section~\ref{sec:background}) predicts, and \citet{decarloNPIsArentExactly2023} find that LMs learn items occurring in fewer environments more easily. We generalize this with a per-item variability score, expected to correlate negatively with accuracy.

We treat each item as dictating its own licensing environments and quantify the item-level variability as licensing entropy over its distribution. For a lemma $\ell$ with count $n_e$ in environment $e$ and total occurrences
$T_\ell$, occurrences outside any detected environment form a remainder bucket
$r_\ell = T_\ell - \sum_e n_e$, added as a single \textit{other} bucket. With
$p_e$ the resulting distribution,
\begin{equation*}
H(\ell) = -\sum_{e} p_e \log_2 p_e .
\end{equation*}
This metric can only approximate the true distribution depending on the corpus data. To increase reliability, only lemmas that occur with a licenser at least 40 times are considered, which means that out of 173 lemmas, only 89 (51\%) are included in the analysis. This metric represents weak-to-strong taxonomy as a spectrum, where the entropy of an item is inversely associated with the traditional sense of ``strength''.

As shown in \autoref{fig:corr_strips}, negative correlation is significant for the larger Goldfish model. Among multilingual models, it is significant only for Gemma-3 27B. The tail of the high-entropy items is dominated by Germanic languages, especially Dutch and English. As such, licensing entropy interacts with the training distribution. Models whose scores correlate significantly with the CC frequencies exhibit the reverse relationship with the licensing entropy. We hypothesize that biases inherited from the training data distribution override this finer-grained linguistic pattern.


\paragraph{When NPIs are ``learned''}

We evaluate ParalleLM \citep{yangAutoregressiveTransformerModels2026}, a set of monolingual models trained over 20M English aligned-words, in their successive checkpoints, for typologically different languages. Among those, we evaluate 7 models and their checkpoints that overlap with \blimppi. \autoref{fig:corr_strips} shows the trajectory of learning curves. For all languages, except Turkish, learning stabilizes by the 5th checkpoint. All models reach well beyond chance-level accuracy.
Since the models see the same 20M words (for 10 epochs) repeatedly rather than a growing corpus, this indicates that canonical NPI-licenser dependencies are learnable from a small corpus. In terms of outliers, Turkish is learned last: accuracy rises above chance only at the fourth checkpoint and it is also the hardest language for the 100MB Goldfish model (62.7). Rather than a typological effect, this is likely due to tokenization of the Turkish negation suffix (Appendix~\ref{app:turkish}).

\paragraph{Comparison with MultiBLiMP.} We compare \blimppi\ with MultiBLiMP \citep{jumeletMultiBLiMP10Massively2026} on the 28 languages both cover on 6 models reported in MultiBLiMP (see Appendix~\ref{app:multiblimp} for per-language scores). On average, models score 4.7 points lower on \blimppi\ (89.7 vs.\ 94.4). The gap depends on how close a model is to ceiling: it is negligible for the strongest models (Gemma 27B $+0.2$, Goldfish 1GB $-1.3$, Llama 70B $-1.6$), but reaches $-9.2$ for EuroLLM 9B and $-11.3$ for OLMo-2 32B. Although the two benchmarks test different phenomena, the gaps are largest for the models that also score low on agreement. This suggests that models need more exposure to a language to perform well on NPI licensing than on agreement, as NPIs are far rarer.

\section{Conclusion}


\blimppi{} offers a multilingual linguistic evaluation suite of minimal pairs, complemented with licensing entropy, distance and n-gram solvability metrics to support not only benchmarking but also understanding of the linguistic skills in LMs. We draw three takeaways from the evaluation and analysis we conduct with 14 models. The first is that, regardless of the target language, model size or training regime, $n$-gram solvability proves the most predictive of all the factors we have checked. $n$-gram baselines are therefore crucial for assessing the difficulty of a benchmark, especially a linguistic one, and should be standard for this line of work. Secondly, our results confirm the finding of \citet{jumeletMultiBLiMP10Massively2026} that models prioritizing multilingual balance in pretraining perform best. Third, we show that when conditions are balanced, linguistic hypothesis can help explain model behavior as well: $n$-grams work as an error oracle, but offer no interpretability, whereas factors such as distance and licensing entropy do.




\section*{Limitations}

\paragraph{Limited coverage and quality checks.}  



\blimppi{} is limited to languages that are well-covered in UD, and is not fully quality checked by native speakers. 
As with any benchmarking efforts, especially multilingual ones, quality and coverage are at odds here.
Therefore, we specifically focus on automatically extracted evaluation data to maintain high coverage; at the same time, to balance quality, we manually create the per-language configurations based on descriptive studies and in collaboration with linguists specializing in the language and native speakers. As a result, the pipeline has two parts: a shared engine of language-agnostic syntactic and modification functions, and language-specific configuration files that supply its parameters: discovery thresholds, the licensing environments that apply, and, where relevant, lists of NPIs and lexical licensers.



\paragraph{Resource imbalance.} Germanic languages have more resources than the others, and \blimppi\ is affected by this. For English, we add examples from the NPI bank \citep{npibank}, and for Dutch from \citet{hoeksemaNegatiefpolaireUitdrukkingenVan2024}. No comparable resource was available for most of the remaining languages, so their coverage rests on what automatic discovery and the descriptive literature could supply.

\paragraph{Licenser coverage.} We could not cover conditional and comparison licensers, as such sentences do not lend themselves to minimal pairs. This is unfortunate, as \citet{jumeletLanguageModelsUse2021} show that different licensing paradigms reinforce other's acquisition in language models in English, and it would have been possible to see how this appears for other languages.

\paragraph{Training data.} We only have information for the data distribution for the Apertus and Goldfish model families, and only for the distribution of non-English languages. Without knowing the distribution for the other model families, it is difficult to analyze at what data volume a multilingual model is able to capture the polarity phenomena of a language.

\section*{Ethics Statement}

\paragraph{Broader impact.} This study aims to provide a more accurate assessment of the linguistic functionality of LLMs across 33 languages. Our results offer further evidence that the language models still struggle most with lower-resource languages, which means unequal access to language technology for the communities speaking these languages. Furthermore, our results highlight the necessity of n-gram baselines for assessing the difficulty of benchmarks. Since this is unfortunately not a common practice, the assessment of language model performance may be over-inflated.

\paragraph{Data sourcing.} As described in section \ref{sec:data_sources}, we rely on existing linguistic resources: UD v.2.18, NPI bank, and Dutch NPI Lexicon. Hence, this work creates no new risks related to data sharing. To the best of our knowledge, these sources do not contain personally identifiable information or toxic content.


\section*{Acknowledgments}
 We thank Jack Hoeksema for access to the Dutch NPI Lexicon, and Manfred Sailer and Frank Richter for access to NPI Bank. We thank László Károly, Riitta-Liisa Valijärvi and her sisters, and Johan Sjons for help configuring their respective languages. We also thank Christian Hardmeier, Stefan Heinrich, Ratish Puduppully, Lucía Gómez-Zaragozá, Terne Sasha Thorn Jakobsen, Nikolas Vitsakis, Amelie Wührl, Harold Rubio, Mattes Ruckdeschel, Johannes Gabriel Sindlinger, Ioana-Daria Vasile, Shiling Deng, Zhuojun Gu, Carmin Vidé, Mikas Jankeliūnas, Virginija Juozapaityte, Argyrios Dritsoulas Mousiadis, Elmiro Vetere, Leo Vitasovic, Satenik Kocharyan, Tomasz Chwałek and Małgorzata Chwałek for native-speaker judgments.

The authors acknowledge the financial support by the Federal Ministry of Research, Technology and Space of Germany and by Sächsische Staatsministerium für Wissenschaft, Kultur und Tourismus in the programme Center of Excellence for AI-research ``Center for Scalable Data Analytics and Artificial Intelligence Dresden/Leipzig'', project identification number: ScaDS.AI.

Arzu Burcu Güven is supported by the AI Pioneer Center. Rob van der Goot is supported by Grant cf25-1861 (LMLM: Linguistically Motivated Language Models) from the Carlsberg Foundation.


\bibliography{latex/NPI,latex/handmade}

\appendix





\section{Excluded Treebanks}
\label{sec:Excl}
Table~\ref{tab:exclusions} lists the UD v2.18 treebanks that meet the size criterion but were excluded, with the criterion from Section~\ref{sec:data_sources} that each fails.

\newcommand{\reasonrule}{\arrayrulecolor{black!25}\cmidrule[0.3pt](l){3-3}\arrayrulecolor{black}}

\begin{table}[t]
\centering\footnotesize
\setlength{\tabcolsep}{2pt}
\renewcommand{\arraystretch}{1}
\begin{tabular}{@{}ll>{\raggedright\arraybackslash}p{3.2cm}@{}}
\toprule
\textbf{Language} & \textbf{Category} & \textbf{Reason} \\
\midrule
Ancient Greek        & Extinct & \multirow{11}{=}{No living speaker community; no native judgements available} \\
Ancient Hebrew       & Extinct & \\
Classical Armenian   & Extinct & \\
Classical Chinese    & Extinct & \\
Coptic               & Extinct & \\
Latin                & Extinct & \\
Middle French        & Extinct & \\
Old Church Slavonic  & Extinct & \\
Old East Slavic      & Extinct & \\
Old French           & Extinct & \\
Sanskrit             & Extinct & \\
\midrule
Basque    & Negation & \multirow{4}{=}{Asymmetric negation: negation alters clausal syntax beyond the negator itself} \\
Estonian  & Negation & \\
Irish     & Negation & \\
Latvian   & Negation & \\
\reasonrule
Arabic    & Negation & Negator varies with tense, mood and predicate type; not tractable by rule \\
\midrule
French     & In flux & \multirow{3}{=}{Negation in flux; judgements vary by register and dialect} \\
Italian    & In flux & \\
Portuguese & In flux & \\
\midrule
Korean   & Annotation & Negation not marked with \texttt{Polarity=Neg} \\
\reasonrule
Japanese & Annotation & Source corpus not redistributable \\
\reasonrule
Naija    & Annotation & Spoken, code-switched corpus \\
\reasonrule
Galician & Annotation & \multirow{2}{=}{Too little manually annotated data} \\
Marathi  & Annotation & \\
\midrule
Thai & Yield & Too few NPI candidates extracted \\
\bottomrule
\end{tabular}
\caption{Universal Dependencies treebanks above \num{100000} tokens that were not included in \blimppi.}
\label{tab:exclusions}
\end{table}

\section{Consulted Work}
\label{sec:Consulted}

Table~\ref{tab:resources} lists the linguistic resources used to build each language's configuration file.

\begin{table*}[t]
\centering
\footnotesize
\setlength{\tabcolsep}{3.5pt}
\renewcommand{\arraystretch}{0.95}
\begin{tabular}{@{}>{\raggedright\arraybackslash}p{1.9cm}
                  >{\raggedright\arraybackslash}p{5.55cm}
                  >{\raggedright\arraybackslash}p{1.9cm}
                  >{\raggedright\arraybackslash}p{5.55cm}@{}}
\toprule
Language & Source & Language & Source \\
\midrule
Armenian & \citet{khanjianNegativeConcordWestern2010} & Norwegian (Nynorsk) & \citet{maehlum-etal-2021-negation} \\
Belarusian & --- & Norwegian (Bokm\aa{}l) & \citet{maehlum-etal-2021-negation} \\
Bulgarian & \citet{haspelmathIndefinitePronouns2017} & Persian & \citet{haspelmathIndefinitePronouns2017} \\
Catalan & \citet{haspelmathIndefinitePronouns2017, espinalNegativePolarityItems2022, giannakidouLongDistanceLicensingNegative1997, vallduviPolarityItemsNwords1994} & Polish & \citet{haspelmathIndefinitePronouns2017} \\
Chinese & \citet{xueNegativePolarityItems2023, linMANDARINSHENMESUPERWEAK} & Romanian & \citet{haspelmathIndefinitePronouns2017, falausThereAnyNegative2008} \\
Croatian & \citet{haspelmathIndefinitePronouns2017} & Russian & \citet{haspelmathIndefinitePronouns2017} \\
Czech & \citet{docekalNwordsNPIsSyntax2020} & Serbian & \citet{haspelmathIndefinitePronouns2017} \\
Danish & \emph{Den Danske Ordbog} via \citet{viknerTreesFieldsNegative2017} & Slovak & --- \\
Dutch & \citet{hoeksemaNegatiefpolaireUitdrukkingenVan2024, haspelmathIndefinitePronouns2017}& Slovenian & --- \\
English & \citet{danescu-niculescu-mizilDontHaveClue, yusukekubotaNPILicensingLogic2021} & Spanish & \citet{aranovichNegativePolarityScalar2007, vallduviPolarityItemsNwords1994} \\
German & \citet{richterBrauchtNiemandenScheren2006, haspelmathIndefinitePronouns2017} \& \citet{npibank} & Swedish & \citet{haspelmathIndefinitePronouns2017, brandtlerEvaluabilityHypothesisSyntax2010} \\
Modern Greek & \citet{giannakidouLongDistanceLicensingNegative1997, haspelmathIndefinitePronouns2017} & Turkish &  \citet{gokselTurkishComprehensiveGrammar2005} \\
Hebrew & \citet{tonciulescu2007indefinites, haspelmathIndefinitePronouns2017} & Ukrainian & --- \\
Hindi & \citet{57d2e1b3-c195-3cc6-84a9-5c2bb14160ce, haspelmathIndefinitePronouns2017} & Urdu & Follows \textit{Hindi} \\
Hungarian & \citet{haspelmathIndefinitePronouns2017, NEGCRITERIONHUNGARIAN, toth1999negative} & Finnish & \citet{haspelmathIndefinitePronouns2017} \\
Icelandic & \citet{haspelmathIndefinitePronouns2017, hoskuldurthrainssonSyntaxIcelandic2007} & Lithuanian & \citet{haspelmathIndefinitePronouns2017} \\
Indonesian & \citet{kroegerExternalNegationMalay2014} &  &  \\
\bottomrule
\end{tabular}
\caption{Descriptive sources consulted when assembling the language configurations.}
\label{tab:resources}
\end{table*}

\section{Licensing Scope}
\label{app:scope}

In order to identify cases where licensing occurs over longer dependencies, we
employ various head-walking strategies. Because different licensers have
different scopal strengths, both in terms of depth and of which structures
they can govern over, we encode these in a licenser- and language-specific
manner. For example, matrix negation --- one of the more common forms of
long-distance licensing, as in (\ref{ex:matrix}) --- is tracked only if the
NPI-containing clause is connected to the negated phrase through a
\texttt{ccomp}, \texttt{csubj}, \texttt{csubj:cop} or \texttt{xcomp} relation.

{\small
\begin{exe}
\ex \label{ex:matrix} I do \mplic{not} believe that she \mpnpi{ever} drank
    \mpnpi{any} milk.
\end{exe}
}

\noindent We follow the linguistic resources when encoding the scopal strength
of lexical licensers. This has been studied extensively for German, Dutch and
English; other languages have less coverage in that respect. Our language
configurations specify numerically how much head-walking is allowed before the
licenser is discarded.

\paragraph{Islands.} Relative and adverbial clauses and parentheticals
(\texttt{acl}, \texttt{acl:relcl}, \texttt{advcl}, \texttt{parataxis}) are not
scope-permeable, and no head-walking crosses them.

\paragraph{Coordination.} Coordination allows licensing depending on which
it attaches. Clausal coordination does not leak the licenser (\ref{ex:s-coord}). Verb-phrase coordination is scope permeable (\ref{ex:vp-coord}); which also holds for questions (\ref{ex:q-coord}).

{\small
\begin{exe}
\ex[??]{\label{ex:s-coord} She is \mplic{not} \mpnpi{ever} calling and he is
    \mpnpi{ever} answering.}
\ex[]{\label{ex:vp-coord} She is \mplic{not} \mpnpi{ever} calling or
    \mpnpi{ever} answering.}
\ex[]{\label{ex:q-coord} \mplic{Does} she \mpnpi{ever} call or \mpnpi{ever}
    answer?}
\end{exe}
}

\paragraph{Complement clauses.} \texttt{ccomp} and \texttt{xcomp} are permeable
to several licenser types; clausal negation (\ref{ex:ccomp-neg}), negative
predicates (\ref{ex:ccomp-pred}), licensing adverbs (\ref{ex:xcomp}).

{\small
\begin{exe}
\ex[]{\label{ex:ccomp-neg} I do \mplic{not} think that he \mpnpi{ever} reads a
    book.}
\ex[]{\label{ex:ccomp-pred} I \mplic{doubt} he \mpnpi{ever} reads a book.}
\ex[]{\label{ex:xcomp} I \mplic{hardly} see him \mpnpi{ever} reading a book.}
\end{exe}
}

\noindent We do not include imperatives, as they are underrepresented in
written corpora.

\subsection{Special Cases}
\label{appendix:spe-cases}

\paragraph{Asymmetrical Negation}

Languages that have an asymmetrical negation paradigm posed a challenge. Where the asymmetry between the negated and positive forms is severe (e.g., Finnish, Estonian, Basque), creating minimal pairs by restoring the licensing sentence to its unlicensing form is not possible with the current UD and UniMorph \citep{kirov-etal-2018-unimorph} resources. Pairs can instead be built by swapping the NPI for a positive polarity item, as we did for Finnish and Lithuanian, but this requires close collaboration with native speakers or experts. As this was not possible for us for the remaining asymmetrical-negation languages, they are excluded from our benchmark.

A less severe form of asymmetrical negation, the genitive of negation, is prevalent in Slavic languages, affecting Polish, Ukrainian, Russian, Belarusian and Slovenian in our benchmark. For these languages, we extracted from UD the case markings conditioned on the positive and negative forms of the verbs. If a verb normally takes an accusative object but undergoes the genitive shift when negated, we do not create minimal pairs with it. As a result, these subsets are limited to intransitive verbs and to verbs that take a genitive object in both positive and negative forms.

\paragraph{Order}

\citet{giannakidouPolaritySensitivityNonVeridical1998} classifies Spanish and Catalan as non-strict NC languages, as the preverbal NPIs in these languages can occur without a licenser (\ref{ex:spa-pre}):

{\small
\begin{exe}
\ex[]{\label{ex:spa}}
\begin{xlist}
\ex \label{ex:spa-pre} \gll \mpnpi{Nadie} bebe leche.\\
     nobody drink.\textsc{prs.3sg} milk\\
\glt `Nobody drinks milk.'
\end{xlist}
\end{exe}
}

We therefore track the position of the NCI relative to the verb when creating pairs for these two languages.

\section{Swap vs. Drop}

\label{appendix:swapdrop}

Dropping the licenser leaves the ungrammatical member one token shorter, which could raise its score independently of grammar. Swapping it for a non-licensing word of the same length does not. \autoref{tab:swap-drop} compares the two over the 7 languages that contain both. If the benchmark were driven by length, drop items would be systematically easier; $\Delta$ is positive or near zero for 5 of the evaluated 14 transformer based models. Complemented with the fact that swapping could also lead to oversimple examples by introducing a semantically incongruent positive item to a negative context, we conclude that the token difference does not distort the linguistic evaluation for transformer-based models. The effect is more significant for n-grams, but they are especially sensitive to swapping, as we discuss in \autoref{appendix:ngrams}.

\begin{table*}[t]
\centering
\small
\begin{tabular}{l rr rr r rr}
\toprule
& \multicolumn{2}{c}{Pairs} & \multicolumn{2}{c}{Accuracy (\%)} & & \multicolumn{2}{c}{Mean $\delta$} \\
\cmidrule(lr){2-3} \cmidrule(lr){4-5} \cmidrule(lr){7-8}
Model & Swap & Drop & Swap & Drop & $\Delta$ & Swap & Drop \\
\midrule
Goldfish 100MB & 939 & 1,835 & 73.3 & 76.3 & +3.1 & 2.69 & 2.38 \\
Goldfish 1GB & 939 & 1,835 & 91.8 & 88.4 & -3.4 & 5.23 & 4.16 \\
Gemma-3 1B & 939 & 1,835 & 88.6 & 88.6 & +0.0 & 6.92 & 5.48 \\
Gemma-3 4B & 939 & 1,835 & 94.0 & 91.7 & -2.3 & 8.49 & 6.19 \\
Apertus 8B & 939 & 1,835 & 92.7 & 92.2 & -0.5 & 8.54 & 6.40 \\
Llama-3.1 8B & 939 & 1,835 & 92.5 & 89.9 & -2.7 & 7.66 & 5.60 \\
Qwen3 8B & 939 & 1,835 & 84.2 & 81.0 & -3.3 & 6.79 & 6.05 \\
Gemma-3 12B & 939 & 1,835 & 97.0 & 94.8 & -2.2 & 9.52 & 6.97 \\
Qwen3 14B & 939 & 1,835 & 90.2 & 85.2 & -5.0 & 8.35 & 6.62 \\
Gemma-3 27B & 939 & 1,835 & 97.6 & 96.6 & -0.9 & 10.39 & 7.89 \\
Qwen3 32B & 939 & 1,835 & 92.8 & 86.8 & -6.0 & 9.66 & 6.58 \\
Aya 32B & 939 & 1,835 & 94.4 & 92.2 & -2.1 & 8.90 & 7.04 \\
Apertus 70B & 939 & 1,835 & 96.0 & 95.4 & -0.6 & 9.29 & 7.16 \\
Llama-3.1 70B & 939 & 1,835 & 97.1 & 95.4 & -1.7 & 9.77 & 7.16 \\
\midrule
\textbf{Multilingual (12)} & \textbf{11,268} & \textbf{22,020} & \textbf{93.1} & \textbf{90.8} & \textbf{-2.3} & \textbf{8.69} & \textbf{6.59} \\
\textbf{Monolingual (2)} & \textbf{1,878} & \textbf{3,670} & \textbf{82.5} & \textbf{82.4} & \textbf{-0.2} & \textbf{3.96} & \textbf{3.27} \\
\textbf{$n$-gram} & \textbf{906} & \textbf{1,492} & \textbf{71.6} & \textbf{60.0} & \textbf{-11.6} & \textbf{1.21} & \textbf{0.57} \\
\bottomrule
\end{tabular}
\caption{Accuracy on \emph{swap} items, where the licenser is replaced by a
non-licensing word, against \emph{drop} items, where it is
deleted and the ungrammatical member is one token shorter. Mean $\delta$ is the average log-probability margin within a pair.
The last three rows average each kind of model separately and are not
comparable with each other. Restricted to the 7 languages holding both item
types.}
\label{tab:swap-drop}
\end{table*}

\section{Example Pairs}
\label{app:npi-list}

Figure~\ref{fig:npi-list} gives one example minimal pair per language.

\begin{figure*}[p]
  \centering
  \includegraphics[page=1,width=0.80\textwidth]{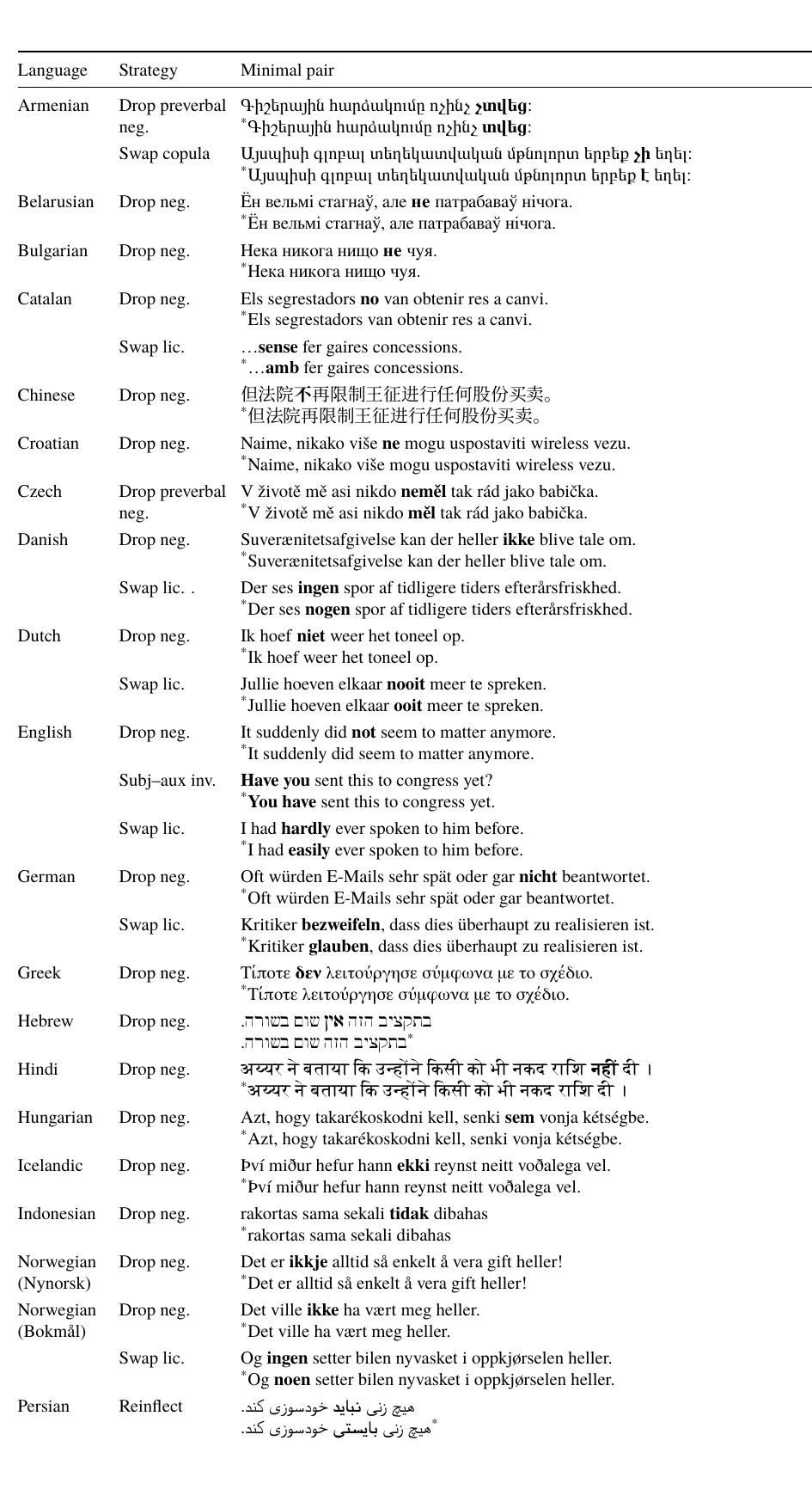}
  \caption{One example minimal pair per language in \blimppi.
    Continued on the following page.}
  \label{fig:npi-list}
\end{figure*}

\begin{figure*}[p]
  \ContinuedFloat
  \centering
  \includegraphics[page=2,width=0.80\textwidth]{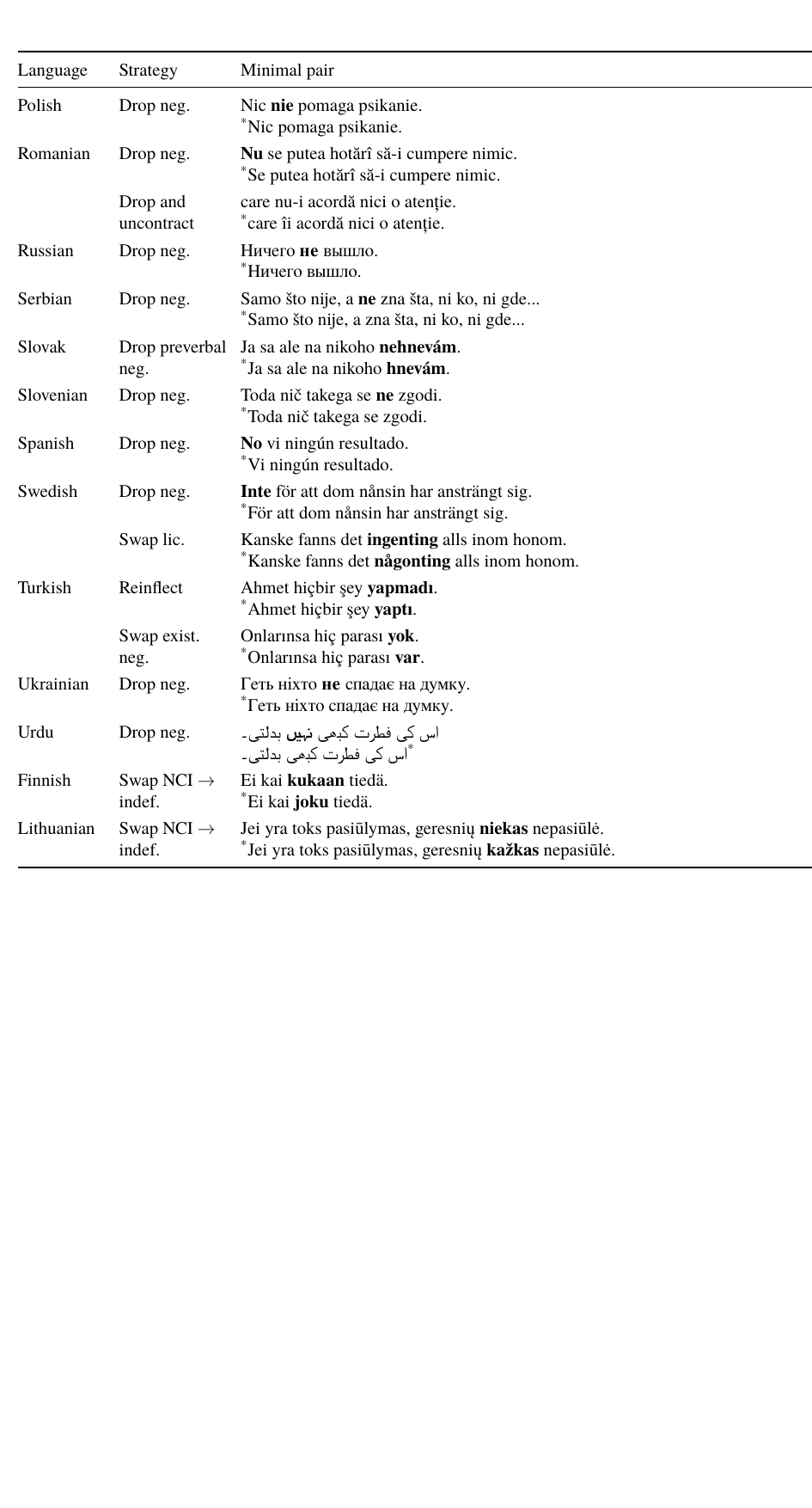}
  \caption{(Continued) One example minimal pair per language in
    \blimppi.}
\end{figure*}

\section{Multiple NPIs}
\label{appendix:mulNPIs}

\autoref{tab:multi-npi} shows the numbers of pairs that contain multiple NPIs per language along with their respective aggregated accuracies over multilingual models.

\begin{table}[t]
\centering
\footnotesize
\setlength{\tabcolsep}{3.5pt}
\begin{tabular}{lrrrrrr}
\toprule
& & \multicolumn{2}{c}{Multi-NPI} & \multicolumn{3}{c}{Accuracy (\%)} \\
\cmidrule(lr){3-4} \cmidrule(lr){5-7}
Language & Pairs & $n$ & \% & Single & Multi & $\Delta$ \\
\midrule
Armenian & 42 & 8 & 19.1 & 86.2 & 88.5 & $+2.3$ \\
Belarusian & 171 & 23 & 13.4 & 90.6 & 92.6 & $+2.0$ \\
Bulgarian & 131 & 14 & 10.7 & 90.9 & 95.6 & $+4.7$ \\
Catalan & 220 & 17 & 7.7 & 93.9 & 95.9 & $+2.0$ \\
Chinese & 34 & 1 & 2.9 & 92.8 & 84.6 & $-8.2$ \\
Croatian & 71 & 9 & 12.7 & 93.1 & 96.6 & $+3.5$ \\
Czech & 1213 & 175 & 14.4 & 90.4 & 91.9 & $+1.5$ \\
Danish & 55 & 0 & 0.0 & 93.4 & -- & -- \\
Dutch & 1771 & 14 & 0.8 & 89.2 & 83.5 & $-5.7$ \\
English & 212 & 5 & 2.4 & 98.4 & 100.0 & $+1.6$ \\
Finnish & 400 & 18 & 4.5 & 91.0 & 94.0 & $+3.0$ \\
German & 428 & 5 & 1.2 & 98.3 & 100.0 & $+1.7$ \\
Greek & 87 & 9 & 10.3 & 88.2 & 98.3 & $+10.1$ \\
Hebrew & 96 & 9 & 9.4 & 85.1 & 96.6 & $+11.5$ \\
Hindi & 107 & 5 & 4.7 & 95.6 & 95.4 & $-0.2$ \\
Hungarian & 13 & 0 & 0.0 & 93.5 & -- & -- \\
Icelandic & 47 & 1 & 2.1 & 72.6 & 53.8 & $-18.8$ \\
Indonesian & 6 & 0 & 0.0 & 94.9 & -- & -- \\
Lithuanian & 46 & 4 & 8.7 & 91.6 & 94.2 & $+2.6$ \\
Bokmål & 43 & 0 & 0.0 & 87.3 & -- & -- \\
Nynorsk & 30 & 0 & 0.0 & 81.5 & -- & -- \\
Persian & 449 & 18 & 4.0 & 90.5 & 94.0 & $+3.5$ \\
Polish & 764 & 62 & 8.1 & 96.2 & 97.4 & $+1.2$ \\
Romanian & 246 & 41 & 16.7 & 91.7 & 89.3 & $-2.4$ \\
Russian & 4199 & 351 & 8.4 & 98.7 & 99.0 & $+0.3$ \\
Serbian & 24 & 4 & 16.7 & 90.4 & 80.8 & $-9.6$ \\
Slovak & 294 & 34 & 11.6 & 85.1 & 81.7 & $-3.4$ \\
Slovenian & 168 & 20 & 11.9 & 88.5 & 86.5 & $-2.0$ \\
Spanish & 272 & 25 & 9.2 & 95.8 & 98.5 & $+2.7$ \\
Swedish & 59 & 2 & 3.4 & 94.9 & 100.0 & $+5.1$ \\
Turkish & 813 & 32 & 3.9 & 94.7 & 94.2 & $-0.5$ \\
Ukrainian & 295 & 55 & 18.6 & 96.1 & 95.4 & $-0.7$ \\
Urdu & 60 & 5 & 8.3 & 90.5 & 96.9 & $+6.4$ \\
\midrule
\textbf{All} & 12866 & 966 & 7.5 & 94.1 & 95.0 & $+0.9$ \\
\bottomrule
\end{tabular}
\caption{Share of minimal pairs whose sentence contains more than one NPI, and accuracy on single- vs.\ multiple-NPI pairs, averaged over the 12 multilingual models. $\Delta$ = Multi $-$ Single (percentage points).}
\label{tab:multi-npi}
\end{table}

\section{Human Audit}
\label{app:audit}

\paragraph{English.} The audit surfaced two problems in English, one
item-level and one systematic.

First, pairs built on \textit{anyone} were flagged because the ungrammatical
member is rescued by a free-choice reading: \textit{She talks to anyone} is
fine without a licenser. We removed \textit{anyone} from the English
inventory rather than attempting to separate the two readings
automatically.

Second, our rule swapped \textit{without} for \textit{with} to remove the
licenser. This works where \textit{without} takes a nominal phrase
\ref{ex:without-np}, but not where it governs a gerund
\ref{ex:without-ger}, in which case the swap produces a string that is
ungrammatical for the wrong reason: the preposition, not the polarity
item. To remedy this, we changed the mechanism to only swap without to with when it governs a nominal phrase

{\small
\begin{exe}
\ex \label{ex:without-np}
\begin{xlist}
\ex[]{\dots would say the most embarrassing things in company,
      \mplic{without} \mpnpi{any} sense of shame at all.}
\ex[*]{\dots would say the most embarrassing things in company,
      \mpbad{with} \mpnpi{any} sense of shame at all.}
\end{xlist}
\ex \label{ex:without-ger}
\begin{xlist}
\ex[]{They sat there for a short time \mplic{without} saying
      \mpnpi{anything}.}
\ex[*]{They sat there for a short time \mpbad{with} saying
      \mpnpi{anything}.}
\end{xlist}
\end{exe}
}

\begin{table}[t]
\centering
\small
\setlength{\tabcolsep}{4pt}
\begin{tabular}{@{}lrrrr@{}}
\toprule
\textbf{Language} & \textbf{Aud.} & \textbf{Pairs} & \textbf{Flagged} & \textbf{\%} \\
\midrule
English    & 1 & 30 &  9 & 30.0 \\
Hindi      & 1 & 15 &  4 & 26.7 \\
German     & 4 & 30 & 24 & 20.0 \\
Chinese    & 2 & 30 &  6 & 10.0 \\
Danish     & 1 & 30 &  3 & 10.0 \\
Romanian   & 1 & 30 &  2 &  6.7 \\
Spanish    & 1 & 15 &  1 &  6.7 \\
Greek      & 3 & 30 &  5 &  5.6 \\
Polish     & 2 & 30 &  3 &  5.0 \\
Croatian   & 1 & 23 &  1 &  4.3 \\
Czech      & 1 & 30 &  1 &  3.3 \\
Russian    & 1 & 30 &  1 &  3.3 \\
Ukrainian  & 1 & 30 &  1 &  3.3 \\
Armenian   & 1 & 15 &  0 &  0.0 \\
Serbian    & 1 & 15 &  0 &  0.0 \\
Turkish    & 1 & 30 &  0 &  0.0 \\
Urdu       & 1 & 15 &  0 &  0.0 \\
\midrule
\textbf{All} & & \textbf{428} & \textbf{61} & \textbf{9.6} \\
\bottomrule
\end{tabular}
\caption{Human audit. \emph{Aud.}\ is the number of auditors, \emph{Pairs} the
sample size per auditor, \emph{Flagged} the number of judgments marking a pair
as not carrying the intended contrast, out of $\mathit{Aud.}\times
\mathit{Pairs}$ judgments. Dutch, Finnish, Hungarian, Lithuanian, Norwegian
and Swedish are absent because their configurations were written with a
native-speaker linguist and vetted during construction.}
\label{tab:audit}
\end{table}


\section{Transformer-based LMs}
\label{app:transformers}

We evaluate Llama-3.1~\cite{grattafiori2024llama3herdmodels}, Aya~\cite{dangAyaExpanseCombining2024}, Gemma-3~\cite{gemmateamGemma3Technical2025}, Qwen3~\cite{yangQwen3TechnicalReport2025}, and Apertus~\cite{apertusApertusDemocratizingOpen2025} series. This set is complemented with Goldfish models~\cite{changGoldfishMonolingualLanguage2026}: monolingual LMs trained on equal-sized data, on GPT-2 architecture. 

\section{n-grams}
\label{appendix:ngrams}
\begin{table}[t]
\centering
\scriptsize
\setlength{\tabcolsep}{3.5pt}
\begin{tabular}{lrrrrr}
\toprule
ISO & Acc. & Ties & OOV/sent & OOV rate & Asym. \\
\midrule
\texttt{hye} & \cellcolor[HTML]{D5D6E9}0.643 & \cellcolor[HTML]{F2F7FD}0.000 & \cellcolor[HTML]{DFECF7}0.14 & \cellcolor[HTML]{E1EDF8}0.007 & \cellcolor[HTML]{61A7D2}0.071 \\
\texttt{bel} & \cellcolor[HTML]{B4AED3}0.754 & \cellcolor[HTML]{F2F7FD}0.000 & \cellcolor[HTML]{D2E3F3}0.25 & \cellcolor[HTML]{CDDFF1}0.016 & \cellcolor[HTML]{E9F2FA}0.006 \\
\texttt{bul} & \cellcolor[HTML]{BAB5D7}0.733 & \cellcolor[HTML]{F2F7FD}0.000 & \cellcolor[HTML]{E3EEF9}0.11 & \cellcolor[HTML]{DCEAF6}0.009 & \cellcolor[HTML]{F2F7FD}0.000 \\
\texttt{cat} & \cellcolor[HTML]{E9EAF2}0.559 & \cellcolor[HTML]{F2F7FD}0.000 & \cellcolor[HTML]{D2E3F3}0.25 & \cellcolor[HTML]{DAE8F6}0.010 & \cellcolor[HTML]{E5EFF9}0.009 \\
\texttt{zho} & \cellcolor[HTML]{BAB5D7}0.735 & \cellcolor[HTML]{F2F7FD}0.000 & \cellcolor[HTML]{F2F7FD}0.00 & \cellcolor[HTML]{F2F7FD}0.000 & \cellcolor[HTML]{F2F7FD}0.000 \\
\texttt{hrv} & \cellcolor[HTML]{DADCEC}0.620 & \cellcolor[HTML]{F2F7FD}0.000 & \cellcolor[HTML]{CBDEF1}0.31 & \cellcolor[HTML]{D3E4F3}0.013 & \cellcolor[HTML]{F2F7FD}0.000 \\
\texttt{ces} & \cellcolor[HTML]{F0F1F4}0.528 & \cellcolor[HTML]{8FC2DE}0.026 & \cellcolor[HTML]{D8E7F5}0.20 & \cellcolor[HTML]{CFE1F2}0.015 & \cellcolor[HTML]{B9D6EA}0.037 \\
\texttt{dan} & \cellcolor[HTML]{A39BC7}0.800 & \cellcolor[HTML]{F2F7FD}0.000 & \cellcolor[HTML]{B2D2E8}0.45 & \cellcolor[HTML]{B7D4EA}0.023 & \cellcolor[HTML]{F2F7FD}0.000 \\
\texttt{nld} & \cellcolor[HTML]{DFE1EE}0.602 & \cellcolor[HTML]{F2F7FD}0.000 & \cellcolor[HTML]{D8E7F5}0.20 & \cellcolor[HTML]{D3E4F3}0.013 & \cellcolor[HTML]{F2F7FD}0.000 \\
\texttt{eng} & \cellcolor[HTML]{DADCEC}0.623 & \cellcolor[HTML]{F2F7FD}0.000 & \cellcolor[HTML]{E7F1FA}0.08 & \cellcolor[HTML]{E6F0F9}0.005 & \cellcolor[HTML]{EAF3FB}0.005 \\
\texttt{fin} & \cellcolor[HTML]{8073AC}0.895 & \cellcolor[HTML]{F2F7FD}0.000 & \cellcolor[HTML]{C7DCEF}0.34 & \cellcolor[HTML]{A0CBE2}0.029 & \cellcolor[HTML]{F2F7FD}0.000 \\
\texttt{deu} & \cellcolor[HTML]{988DBE}0.834 & \cellcolor[HTML]{F2F7FD}0.000 & \cellcolor[HTML]{A9CFE5}0.49 & \cellcolor[HTML]{AFD1E7}0.025 & \cellcolor[HTML]{EEF5FC}0.002 \\
\texttt{ell} & \cellcolor[HTML]{E5E7F0}0.575 & \cellcolor[HTML]{F2F7FD}0.000 & \cellcolor[HTML]{DFECF7}0.14 & \cellcolor[HTML]{E1EDF8}0.007 & \cellcolor[HTML]{F2F7FD}0.000 \\
\texttt{heb} & \cellcolor[HTML]{E8E9F1}0.562 & \cellcolor[HTML]{F2F7FD}0.000 & \cellcolor[HTML]{D1E2F3}0.26 & \cellcolor[HTML]{D1E2F3}0.014 & \cellcolor[HTML]{F2F7FD}0.000 \\
\texttt{hin} & \cellcolor[HTML]{F8F5F1}0.486 & \cellcolor[HTML]{F2F7FD}0.000 & \cellcolor[HTML]{E5EFF9}0.10 & \cellcolor[HTML]{E8F1FA}0.004 & \cellcolor[HTML]{F2F7FD}0.000 \\
\texttt{hun} & \cellcolor[HTML]{7764A5}0.923 & \cellcolor[HTML]{F2F7FD}0.000 & \cellcolor[HTML]{65AAD4}0.77 & \cellcolor[HTML]{61A7D2}0.043 & \cellcolor[HTML]{F2F7FD}0.000 \\
\texttt{isl} & \cellcolor[HTML]{EFF0F4}0.532 & \cellcolor[HTML]{F2F7FD}0.000 & \cellcolor[HTML]{D7E6F5}0.21 & \cellcolor[HTML]{DCEAF6}0.009 & \cellcolor[HTML]{F2F7FD}0.000 \\
\texttt{ind} & \cellcolor[HTML]{F6F6F7}0.500 & \cellcolor[HTML]{F2F7FD}0.000 & \cellcolor[HTML]{DCE9F6}0.17 & \cellcolor[HTML]{DFEBF7}0.008 & \cellcolor[HTML]{F2F7FD}0.000 \\
\texttt{lit} & \cellcolor[HTML]{8275AD}0.891 & \cellcolor[HTML]{F2F7FD}0.000 & \cellcolor[HTML]{C6DBEF}0.35 & \cellcolor[HTML]{BAD6EB}0.022 & \cellcolor[HTML]{F2F7FD}0.000 \\
\texttt{nob} & \cellcolor[HTML]{FED39C}0.326 & \cellcolor[HTML]{F2F7FD}0.000 & \cellcolor[HTML]{E8F1FA}0.07 & \cellcolor[HTML]{E8F1FA}0.004 & \cellcolor[HTML]{F2F7FD}0.000 \\
\texttt{nno} & \cellcolor[HTML]{DFE1EE}0.600 & \cellcolor[HTML]{F2F7FD}0.000 & \cellcolor[HTML]{BCD7EB}0.40 & \cellcolor[HTML]{C2D9EE}0.020 & \cellcolor[HTML]{F2F7FD}0.000 \\
\texttt{pes} & \cellcolor[HTML]{FED8A6}0.341 & \cellcolor[HTML]{EBF3FB}0.002 & \cellcolor[HTML]{DFECF7}0.14 & \cellcolor[HTML]{DFEBF7}0.008 & \cellcolor[HTML]{E2EDF8}0.011 \\
\texttt{pol} & \cellcolor[HTML]{B9B3D6}0.741 & \cellcolor[HTML]{F2F7FD}0.000 & \cellcolor[HTML]{D7E6F5}0.21 & \cellcolor[HTML]{CADEF0}0.017 & \cellcolor[HTML]{ECF4FB}0.004 \\
\texttt{ron} & \cellcolor[HTML]{FAEEDF}0.451 & \cellcolor[HTML]{F2F7FD}0.000 & \cellcolor[HTML]{B8D5EA}0.42 & \cellcolor[HTML]{C2D9EE}0.020 & \cellcolor[HTML]{ECF4FB}0.004 \\
\texttt{rus} & \cellcolor[HTML]{ABA3CD}0.782 & \cellcolor[HTML]{F2F7FD}0.000 & \cellcolor[HTML]{C7DCEF}0.34 & \cellcolor[HTML]{BED8EC}0.021 & \cellcolor[HTML]{F2F7FD}0.000 \\
\texttt{srp} & \cellcolor[HTML]{B6B0D4}0.750 & \cellcolor[HTML]{F2F7FD}0.000 & \cellcolor[HTML]{61A7D2}0.79 & \cellcolor[HTML]{8FC2DE}0.033 & \cellcolor[HTML]{F2F7FD}0.000 \\
\texttt{slk} & \cellcolor[HTML]{F6F6F7}0.500 & \cellcolor[HTML]{61A7D2}0.034 & \cellcolor[HTML]{D1E2F3}0.26 & \cellcolor[HTML]{A4CCE3}0.028 & \cellcolor[HTML]{84BCDB}0.058 \\
\texttt{slv} & \cellcolor[HTML]{C2BEDC}0.708 & \cellcolor[HTML]{F2F7FD}0.000 & \cellcolor[HTML]{BCD7EB}0.40 & \cellcolor[HTML]{C2D9EE}0.020 & \cellcolor[HTML]{F2F7FD}0.000 \\
\texttt{spa} & \cellcolor[HTML]{F3F3F5}0.518 & \cellcolor[HTML]{F2F7FD}0.000 & \cellcolor[HTML]{C8DCF0}0.33 & \cellcolor[HTML]{D6E5F4}0.012 & \cellcolor[HTML]{F2F7FD}0.000 \\
\texttt{swe} & \cellcolor[HTML]{C0BDDB}0.712 & \cellcolor[HTML]{F2F7FD}0.000 & \cellcolor[HTML]{D8E7F5}0.20 & \cellcolor[HTML]{D6E5F4}0.012 & \cellcolor[HTML]{F2F7FD}0.000 \\
\texttt{tur} & \cellcolor[HTML]{FCE6C8}0.407 & \cellcolor[HTML]{C9DDF0}0.014 & \cellcolor[HTML]{CEE0F2}0.28 & \cellcolor[HTML]{A4CCE3}0.028 & \cellcolor[HTML]{9DCAE1}0.049 \\
\texttt{ukr} & \cellcolor[HTML]{CBC9E2}0.678 & \cellcolor[HTML]{F2F7FD}0.000 & \cellcolor[HTML]{BCD7EB}0.40 & \cellcolor[HTML]{BAD6EB}0.022 & \cellcolor[HTML]{E7F1FA}0.007 \\
\texttt{urd} & \cellcolor[HTML]{EFF0F4}0.533 & \cellcolor[HTML]{F2F7FD}0.000 & \cellcolor[HTML]{ABD0E6}0.48 & \cellcolor[HTML]{CADEF0}0.017 & \cellcolor[HTML]{F2F7FD}0.000 \\
\midrule
\textbf{Mean} & \textbf{0.632} & \textbf{0.002} & \textbf{0.29} & \textbf{0.016} & \textbf{0.008} \\
\bottomrule
\end{tabular}
\caption{KenLM 5-gram baseline by language (ISO 639-3). \textit{Ties} is the proportion of pairs scored identically; \textit{OOV/sent} the mean number of out-of-vocabulary tokens in the acceptable member; \textit{OOV rate} the OOV proportion over all tokens; \textit{Asym.} the proportion of pairs whose two members differ in OOV count. Asymmetry near zero indicates that OOV mass is shared across pair members and therefore cannot account for the score difference. Colour is per column: accuracy is orange below chance and purple above; the diagnostics darken as they rise.}
\label{tab:ngram-oov}
\end{table}

We train 5-gram models; word-level, based on whitespace and punctuation splits. For Chinese we trained a character based n-gram model, as \citet{xiangCLiMPBenchmarkChinese2021b} demonstrates that character level n-grams performs better than word based n-gram for their minimal pair benchmark. We use KenLM with default hyperparameters \citep{heafield-2011-kenlm} and Goldfish monolingual corpora for training \citep{changGoldfishMonolingualLanguage2026}. To make the amount of information provided to the model fairer, we use byte premiums by \citet{arnett-etal-2024-bit}, and train models on 1 GB of English data and its byte-premium counterparts. Only for Nynorsk, we trained on smaller data (541 MB) as the required size was not available. The overall average $n$-gram accuracy is 63.2, indicating that the task is challenging for $n$-gram-based models (random baseline: 50.0\%). The n-gram per-language scores are given in \autoref{tab:acc-full}, most languages are in the 50-65 range, and most outliers have plausible data-based explanations.

Hungarian (92.3), Lithuanian (89.1), and Finnish (89.5) constitute the highest scores, whereas Turkish (40.7), Persian (34.1) and Norwegian-Bokmål (32.6) constitute the lowest. The highest scores follow from two properties of the data. First, Lithuanian and Finnish use a different minimal-pairing strategy, in which the ungrammatical
member is formed by substituting a PPI for the NPI in a negative context; such substitutions appear to be entirely unattested in local co-occurrence statistics. Hungarian combines a small dataset size with low licenser-to-NPI distance (\autoref{app:distance}), both of which could boost $n$-gram performance.

In Turkish, negation is expressed by a bound morph, which is challenging for word-level $n$-gram models to capture. Similarly, Persian negative morph is bound, which is challenging for $n$-grams, but since the morph is a prefix, it is more identifiable for subword tokenisation-based neural models \citep{lernerLikelyUnlikelyBPEbased2025}. As such, while $n$-gram accuracy is below chance, the neural models struggle less with Persian. Czech and Slovak also mark negation with a bound morph and are likewise difficult for $n$-grams (52.8 and 50.0 respectively, against an average of 63.2).
Norwegian-Bokmål is the outlier we can account for least well. Both Norwegian varieties do, however, show larger distance effects (\autoref{app:distance}) and rest on fewer pairs (\autoref{tab:languages}).

OOV rates are highest for Hungarian and Serbian, which are our second and
third smallest sets ($n = 13$ and $24$); with so few items, a handful of
mismatches --- proper nouns, in both cases --- is enough to push the rate up.
The high rates for Danish and German (0.45 and 0.49) come instead from
compounds, which Goldfish's vocabulary does not cover as whole words. Genre
mismatch is a further likely source, as UD draws heavily on newspaper text
while the Goldfish corpora do not. Urdu is a separate case: UD tokenisation
joins compound expressions with underscores, and this is carried over into the
benchmark.

OOV mass is symmetric. The minimal edit targets negation, so the acceptable
and unacceptable members of a pair contain the same OOV tokens; OOV inflates
both scores equally and cannot account for the reported accuracies.

\begin{table*}[t]
\centering
\scriptsize
\setlength{\tabcolsep}{3.2pt}
\resizebox{\textwidth}{!}{%
\begin{tabular}{l l | cccccccccccc | c | cc}
\toprule
 &  & \multicolumn{2}{c}{Llama-3.1} & \multicolumn{1}{c}{Aya} & \multicolumn{4}{c}{Gemma-3} & \multicolumn{3}{c}{Qwen3} & \multicolumn{2}{c}{Apertus} & \multicolumn{1}{c}{\textit{Mean}} & \multicolumn{2}{c}{Goldfish} \\
\cmidrule(lr){3-4} \cmidrule(lr){5-5} \cmidrule(lr){6-9} \cmidrule(lr){10-12} \cmidrule(lr){13-14} \cmidrule(lr){16-17}
\texttt{ISO} & Language & \texttt{8B} & \texttt{70B} & \texttt{32B} & \texttt{1B} & \texttt{4B} & \texttt{12B} & \texttt{27B} & \texttt{8B} & \texttt{14B} & \texttt{32B} & \texttt{8B} & \texttt{70B} &  & \texttt{100MB} & \texttt{1GB} \\
\midrule
\texttt{hye} & Armenian & \cellcolor[HTML]{55AB7C}96.4 & \cellcolor[HTML]{55AB7C}96.4 & \cellcolor[HTML]{55AB7C}96.4 & \cellcolor[HTML]{B9E193}85.7 & \cellcolor[HTML]{47926F}\textbf{100.0} & \cellcolor[HTML]{47926F}\textbf{100.0} & \cellcolor[HTML]{47926F}\textbf{100.0} & \cellcolor[HTML]{B9E193}85.7 & \cellcolor[HTML]{71C087}92.9 & \cellcolor[HTML]{B9E193}85.7 & \cellcolor[HTML]{96D28F}89.3 & \cellcolor[HTML]{47926F}\textbf{100.0} & \cellcolor[HTML]{64BA83}\textbf{94.0} & \cellcolor[HTML]{47926F}100.0 & \cellcolor[HTML]{47926F}100.0 \\
\texttt{bel} & Belarusian & \cellcolor[HTML]{B9E193}85.6 & \cellcolor[HTML]{59B380}95.2 & \cellcolor[HTML]{79C489}92.0 & \cellcolor[HTML]{59B380}95.2 & \cellcolor[HTML]{71C087}92.8 & \cellcolor[HTML]{53A87A}96.8 & \cellcolor[HTML]{4A9872}\textbf{99.2} & \cellcolor[HTML]{B1DE92}86.4 & \cellcolor[HTML]{94D18F}89.6 & \cellcolor[HTML]{56AD7D}96.0 & \cellcolor[HTML]{68BC84}93.6 & \cellcolor[HTML]{50A377}97.6 & \cellcolor[HTML]{6CBE86}\textbf{93.3} & \cellcolor[HTML]{5FB782}94.4 & \cellcolor[HTML]{4D9D75}98.4 \\
\texttt{bul} & Bulgarian & \cellcolor[HTML]{B0DD92}86.6 & \cellcolor[HTML]{52A77A}96.9 & \cellcolor[HTML]{F3FAC0}77.3 & \cellcolor[HTML]{88CB8D}90.7 & \cellcolor[HTML]{4FA076}97.9 & \cellcolor[HTML]{4B9973}99.0 & \cellcolor[HTML]{4B9973}99.0 & \cellcolor[HTML]{9DD590}88.7 & \cellcolor[HTML]{57AF7E}95.9 & \cellcolor[HTML]{71C087}92.8 & \cellcolor[HTML]{47926F}\textbf{100.0} & \cellcolor[HTML]{47926F}\textbf{100.0} & \cellcolor[HTML]{68BC84}\textbf{93.7} & \cellcolor[HTML]{4B9973}99.0 & \cellcolor[HTML]{4B9973}99.0 \\
\texttt{cat} & Catalan & \cellcolor[HTML]{6ABD85}93.5 & \cellcolor[HTML]{50A377}97.6 & \cellcolor[HTML]{96D28F}89.4 & \cellcolor[HTML]{A5D891}87.8 & \cellcolor[HTML]{7BC589}91.9 & \cellcolor[HTML]{53A87A}96.7 & \cellcolor[HTML]{4D9D75}\textbf{98.4} & \cellcolor[HTML]{ACDB92}87.0 & \cellcolor[HTML]{62B883}94.3 & \cellcolor[HTML]{59B380}95.1 & \cellcolor[HTML]{59B380}95.1 & \cellcolor[HTML]{50A377}97.6 & \cellcolor[HTML]{68BC84}\textbf{93.7} & \cellcolor[HTML]{4A9872}99.2 & \cellcolor[HTML]{47926F}100.0 \\
\texttt{zho} & Chinese & \cellcolor[HTML]{79C489}92.0 & \cellcolor[HTML]{47926F}\textbf{100.0} & \cellcolor[HTML]{79C489}92.0 & \cellcolor[HTML]{56AD7D}96.0 & \cellcolor[HTML]{56AD7D}96.0 & \cellcolor[HTML]{56AD7D}96.0 & \cellcolor[HTML]{47926F}\textbf{100.0} & \cellcolor[HTML]{79C489}92.0 & \cellcolor[HTML]{56AD7D}96.0 & \cellcolor[HTML]{56AD7D}96.0 & \cellcolor[HTML]{56AD7D}96.0 & \cellcolor[HTML]{79C489}92.0 & \cellcolor[HTML]{58B17F}\textbf{95.3} & \cellcolor[HTML]{C5E698}84.0 & \cellcolor[HTML]{56AD7D}96.0 \\
\texttt{hrv} & Croatian & \cellcolor[HTML]{58B17F}95.5 & \cellcolor[HTML]{4FA177}97.7 & \cellcolor[HTML]{B1DE92}86.4 & \cellcolor[HTML]{4FA177}97.7 & \cellcolor[HTML]{4FA177}97.7 & \cellcolor[HTML]{47926F}\textbf{100.0} & \cellcolor[HTML]{47926F}\textbf{100.0} & \cellcolor[HTML]{6CBE86}93.2 & \cellcolor[HTML]{86CA8C}90.9 & \cellcolor[HTML]{6CBE86}93.2 & \cellcolor[HTML]{4FA177}97.7 & \cellcolor[HTML]{47926F}\textbf{100.0} & \cellcolor[HTML]{57AF7E}\textbf{95.8} & \cellcolor[HTML]{9DD590}88.6 & \cellcolor[HTML]{4FA177}97.7 \\
\texttt{ces} & Czech & \cellcolor[HTML]{7EC68A}91.7 & \cellcolor[HTML]{52A77A}97.0 & \cellcolor[HTML]{55AC7C}96.2 & \cellcolor[HTML]{A8DA91}87.5 & \cellcolor[HTML]{53A87A}96.7 & \cellcolor[HTML]{51A478}97.3 & \cellcolor[HTML]{4A9872}99.2 & \cellcolor[HTML]{B9E193}85.6 & \cellcolor[HTML]{8FCE8E}90.1 & \cellcolor[HTML]{5FB782}94.4 & \cellcolor[HTML]{4D9D75}98.3 & \cellcolor[HTML]{499570}\textbf{99.5} & \cellcolor[HTML]{5FB782}\textbf{94.5} & \cellcolor[HTML]{93D08F}89.8 & \cellcolor[HTML]{4FA177}97.8 \\
\texttt{dan} & Danish & \cellcolor[HTML]{6CBE86}93.2 & \cellcolor[HTML]{58B17F}95.5 & \cellcolor[HTML]{9DD590}88.6 & \cellcolor[HTML]{4FA177}97.7 & \cellcolor[HTML]{58B17F}95.5 & \cellcolor[HTML]{47926F}\textbf{100.0} & \cellcolor[HTML]{47926F}\textbf{100.0} & \cellcolor[HTML]{9DD590}88.6 & \cellcolor[HTML]{6CBE86}93.2 & \cellcolor[HTML]{86CA8C}90.9 & \cellcolor[HTML]{4FA177}97.7 & \cellcolor[HTML]{4FA177}97.7 & \cellcolor[HTML]{5BB581}\textbf{94.9} & \cellcolor[HTML]{58B17F}95.5 & \cellcolor[HTML]{47926F}100.0 \\
\texttt{nld} & Dutch & \cellcolor[HTML]{77C388}92.3 & \cellcolor[HTML]{51A478}97.3 & \cellcolor[HTML]{58B17F}95.4 & \cellcolor[HTML]{6FBF86}93.2 & \cellcolor[HTML]{57AF7E}95.8 & \cellcolor[HTML]{52A579}97.1 & \cellcolor[HTML]{4D9C74}\textbf{98.5} & \cellcolor[HTML]{DEF1A8}80.7 & \cellcolor[HTML]{A8DA91}87.4 & \cellcolor[HTML]{9DD590}88.6 & \cellcolor[HTML]{62B883}94.2 & \cellcolor[HTML]{53A87A}96.7 & \cellcolor[HTML]{6FBF86}\textbf{93.1} & \cellcolor[HTML]{D4EDA1}82.2 & \cellcolor[HTML]{55AB7C}96.3 \\
\texttt{eng} & English & \cellcolor[HTML]{47926F}\textbf{100.0} & \cellcolor[HTML]{47926F}\textbf{100.0} & \cellcolor[HTML]{47926F}\textbf{100.0} & \cellcolor[HTML]{47926F}\textbf{100.0} & \cellcolor[HTML]{47926F}\textbf{100.0} & \cellcolor[HTML]{4A9671}99.2 & \cellcolor[HTML]{47926F}\textbf{100.0} & \cellcolor[HTML]{4A9671}99.2 & \cellcolor[HTML]{4A9671}99.2 & \cellcolor[HTML]{4A9671}99.2 & \cellcolor[HTML]{4A9671}99.2 & \cellcolor[HTML]{4A9671}99.2 & \cellcolor[HTML]{489470}\textbf{99.6} & \cellcolor[HTML]{58B17F}95.5 & \cellcolor[HTML]{4A9671}99.2 \\
\texttt{fin} & Finnish & \cellcolor[HTML]{6CBE86}93.3 & \cellcolor[HTML]{55AC7C}96.1 & \cellcolor[HTML]{C1E595}84.7 & \cellcolor[HTML]{73C187}92.8 & \cellcolor[HTML]{5AB481}95.0 & \cellcolor[HTML]{50A377}97.5 & \cellcolor[HTML]{54A97B}96.7 & \cellcolor[HTML]{D8EFA4}81.6 & \cellcolor[HTML]{B5DF93}86.1 & \cellcolor[HTML]{96D28F}89.4 & \cellcolor[HTML]{52A579}97.2 & \cellcolor[HTML]{4FA177}\textbf{97.8} & \cellcolor[HTML]{77C388}\textbf{92.3} & \cellcolor[HTML]{50A377}97.5 & \cellcolor[HTML]{4B9973}98.9 \\
\texttt{deu} & German & \cellcolor[HTML]{4B9973}98.9 & \cellcolor[HTML]{4A9872}99.2 & \cellcolor[HTML]{489470}\textbf{99.7} & \cellcolor[HTML]{52A77A}96.9 & \cellcolor[HTML]{4B9973}98.9 & \cellcolor[HTML]{499570}99.4 & \cellcolor[HTML]{499570}99.4 & \cellcolor[HTML]{4D9D75}98.3 & \cellcolor[HTML]{499570}99.4 & \cellcolor[HTML]{4D9D75}98.3 & \cellcolor[HTML]{499570}99.4 & \cellcolor[HTML]{489470}\textbf{99.7} & \cellcolor[HTML]{4B9973}\textbf{99.0} & \cellcolor[HTML]{5FB782}94.4 & \cellcolor[HTML]{55AB7C}96.3 \\
\texttt{ell} & Greek & \cellcolor[HTML]{7BC589}91.8 & \cellcolor[HTML]{66BB84}93.9 & \cellcolor[HTML]{56AD7D}95.9 & \cellcolor[HTML]{56AD7D}95.9 & \cellcolor[HTML]{A5D891}87.8 & \cellcolor[HTML]{56AD7D}95.9 & \cellcolor[HTML]{4FA076}\textbf{98.0} & \cellcolor[HTML]{C8E89A}83.7 & \cellcolor[HTML]{7BC589}91.8 & \cellcolor[HTML]{7BC589}91.8 & \cellcolor[HTML]{56AD7D}95.9 & \cellcolor[HTML]{4FA076}\textbf{98.0} & \cellcolor[HTML]{6ABD85}\textbf{93.4} & \cellcolor[HTML]{56AD7D}95.9 & \cellcolor[HTML]{4FA076}98.0 \\
\texttt{heb} & Hebrew & \cellcolor[HTML]{5FB782}94.3 & \cellcolor[HTML]{55AC7C}96.2 & \cellcolor[HTML]{4E9E75}\textbf{98.1} & \cellcolor[HTML]{AEDC92}86.8 & \cellcolor[HTML]{5FB782}94.3 & \cellcolor[HTML]{5FB782}94.3 & \cellcolor[HTML]{4E9E75}\textbf{98.1} & \cellcolor[HTML]{75C288}92.5 & \cellcolor[HTML]{5FB782}94.3 & \cellcolor[HTML]{C0E494}84.9 & \cellcolor[HTML]{4E9E75}\textbf{98.1} & \cellcolor[HTML]{75C288}92.5 & \cellcolor[HTML]{68BC84}\textbf{93.7} & \cellcolor[HTML]{5FB782}94.3 & \cellcolor[HTML]{4E9E75}98.1 \\
\texttt{hin} & Hindi & \cellcolor[HTML]{4E9E75}98.1 & \cellcolor[HTML]{47926F}\textbf{100.0} & \cellcolor[HTML]{47926F}\textbf{100.0} & \cellcolor[HTML]{62B883}94.2 & \cellcolor[HTML]{47926F}\textbf{100.0} & \cellcolor[HTML]{47926F}\textbf{100.0} & \cellcolor[HTML]{4E9E75}98.1 & \cellcolor[HTML]{4E9E75}98.1 & \cellcolor[HTML]{62B883}94.2 & \cellcolor[HTML]{4E9E75}98.1 & \cellcolor[HTML]{4E9E75}98.1 & \cellcolor[HTML]{4E9E75}98.1 & \cellcolor[HTML]{4E9E75}\textbf{98.1} & \cellcolor[HTML]{8DCD8E}90.4 & \cellcolor[HTML]{4E9E75}98.1 \\
\texttt{hun} & Hungarian & \cellcolor[HTML]{7EC68A}91.7 & \cellcolor[HTML]{47926F}\textbf{100.0} & \cellcolor[HTML]{CBE99C}83.3 & \cellcolor[HTML]{CBE99C}83.3 & \cellcolor[HTML]{7EC68A}91.7 & \cellcolor[HTML]{47926F}\textbf{100.0} & \cellcolor[HTML]{47926F}\textbf{100.0} & \cellcolor[HTML]{7EC68A}91.7 & \cellcolor[HTML]{7EC68A}91.7 & \cellcolor[HTML]{CBE99C}83.3 & \cellcolor[HTML]{47926F}\textbf{100.0} & \cellcolor[HTML]{47926F}\textbf{100.0} & \cellcolor[HTML]{6FBF86}\textbf{93.1} & \cellcolor[HTML]{47926F}100.0 & \cellcolor[HTML]{47926F}100.0 \\
\texttt{isl} & Icelandic & \cellcolor[HTML]{F9FDC9}76.0 & \cellcolor[HTML]{A3D791}88.0 & \cellcolor[HTML]{BE4763}44.0 & \cellcolor[HTML]{F9FDC9}76.0 & \cellcolor[HTML]{79C489}92.0 & \cellcolor[HTML]{79C489}92.0 & \cellcolor[HTML]{56AD7D}\textbf{96.0} & \cellcolor[HTML]{FEE3A7}68.0 & \cellcolor[HTML]{FECB93}64.0 & \cellcolor[HTML]{E4F3AB}80.0 & \cellcolor[HTML]{A3D791}88.0 & \cellcolor[HTML]{56AD7D}\textbf{96.0} & \cellcolor[HTML]{E4F3AB}\textbf{80.0} & \cellcolor[HTML]{A3D791}88.0 & \cellcolor[HTML]{56AD7D}96.0 \\
\texttt{ind} & Indonesian & \cellcolor[HTML]{47926F}\textbf{100.0} & \cellcolor[HTML]{47926F}\textbf{100.0} & \cellcolor[HTML]{47926F}\textbf{100.0} & \cellcolor[HTML]{47926F}\textbf{100.0} & \cellcolor[HTML]{47926F}\textbf{100.0} & \cellcolor[HTML]{47926F}\textbf{100.0} & \cellcolor[HTML]{47926F}\textbf{100.0} & \cellcolor[HTML]{47926F}\textbf{100.0} & \cellcolor[HTML]{47926F}\textbf{100.0} & \cellcolor[HTML]{47926F}\textbf{100.0} & \cellcolor[HTML]{FEDBA0}66.7 & \cellcolor[HTML]{47926F}\textbf{100.0} & \cellcolor[HTML]{52A579}\textbf{97.2} & \cellcolor[HTML]{47926F}100.0 & \cellcolor[HTML]{47926F}100.0 \\
\texttt{lit} & Lithuanian & \cellcolor[HTML]{8DCD8E}90.2 & \cellcolor[HTML]{50A377}97.6 & \cellcolor[HTML]{73C187}92.7 & \cellcolor[HTML]{CEEA9E}82.9 & \cellcolor[HTML]{73C187}92.7 & \cellcolor[HTML]{8DCD8E}90.2 & \cellcolor[HTML]{50A377}97.6 & \cellcolor[HTML]{8DCD8E}90.2 & \cellcolor[HTML]{BAE293}85.4 & \cellcolor[HTML]{47926F}\textbf{100.0} & \cellcolor[HTML]{73C187}92.7 & \cellcolor[HTML]{50A377}97.6 & \cellcolor[HTML]{75C288}\textbf{92.5} & \cellcolor[HTML]{50A377}97.6 & \cellcolor[HTML]{50A377}97.6 \\
\texttt{nob} & Norwegian-Bokmaal & \cellcolor[HTML]{47926F}\textbf{100.0} & \cellcolor[HTML]{47926F}\textbf{100.0} & \cellcolor[HTML]{71C087}92.9 & \cellcolor[HTML]{71C087}92.9 & \cellcolor[HTML]{71C087}92.9 & \cellcolor[HTML]{47926F}\textbf{100.0} & \cellcolor[HTML]{47926F}\textbf{100.0} & \cellcolor[HTML]{EBF7B6}78.6 & \cellcolor[HTML]{71C087}92.9 & \cellcolor[HTML]{47926F}\textbf{100.0} & \cellcolor[HTML]{47926F}\textbf{100.0} & \cellcolor[HTML]{71C087}92.9 & \cellcolor[HTML]{59B380}\textbf{95.2} & \cellcolor[HTML]{71C087}92.9 & \cellcolor[HTML]{47926F}100.0 \\
\texttt{nno} & Norwegian-Nynorsk & \cellcolor[HTML]{9AD390}88.9 & \cellcolor[HTML]{5FB782}94.4 & \cellcolor[HTML]{EFF8BC}77.8 & \cellcolor[HTML]{EFF8BC}77.8 & \cellcolor[HTML]{5FB782}94.4 & \cellcolor[HTML]{5FB782}94.4 & \cellcolor[HTML]{5FB782}94.4 & \cellcolor[HTML]{FFF5C0}72.2 & \cellcolor[HTML]{EFF8BC}77.8 & \cellcolor[HTML]{5FB782}94.4 & \cellcolor[HTML]{47926F}\textbf{100.0} & \cellcolor[HTML]{5FB782}94.4 & \cellcolor[HTML]{9FD690}\textbf{88.4} & \cellcolor[HTML]{9AD390}88.9 & --- \\
\texttt{pes} & Persian & \cellcolor[HTML]{4C9A73}98.7 & \cellcolor[HTML]{4A9671}99.3 & \cellcolor[HTML]{4C9A73}98.7 & \cellcolor[HTML]{4FA076}98.0 & \cellcolor[HTML]{4A9671}99.3 & \cellcolor[HTML]{4A9671}99.3 & \cellcolor[HTML]{47926F}\textbf{100.0} & \cellcolor[HTML]{A1D790}88.2 & \cellcolor[HTML]{8FCE8E}90.2 & \cellcolor[HTML]{5BB581}94.8 & \cellcolor[HTML]{4A9671}99.3 & \cellcolor[HTML]{4A9671}99.3 & \cellcolor[HTML]{52A579}\textbf{97.1} & \cellcolor[HTML]{58B17F}95.4 & \cellcolor[HTML]{47926F}100.0 \\
\texttt{pol} & Polish & \cellcolor[HTML]{5DB682}94.7 & \cellcolor[HTML]{59B380}95.2 & \cellcolor[HTML]{56AD7D}96.0 & \cellcolor[HTML]{54A97B}96.7 & \cellcolor[HTML]{4A9872}99.1 & \cellcolor[HTML]{4A9671}99.3 & \cellcolor[HTML]{47926F}\textbf{100.0} & \cellcolor[HTML]{79C489}92.1 & \cellcolor[HTML]{58B17F}95.4 & \cellcolor[HTML]{58B07F}95.6 & \cellcolor[HTML]{4FA076}97.9 & \cellcolor[HTML]{4D9D75}98.4 & \cellcolor[HTML]{53A87A}\textbf{96.7} & \cellcolor[HTML]{52A77A}97.0 & \cellcolor[HTML]{4A9671}99.3 \\
\texttt{ron} & Romanian & \cellcolor[HTML]{7EC68A}91.7 & \cellcolor[HTML]{55AB7C}96.3 & \cellcolor[HTML]{52A579}97.2 & \cellcolor[HTML]{9AD390}89.0 & \cellcolor[HTML]{52A579}97.2 & \cellcolor[HTML]{52A579}97.2 & \cellcolor[HTML]{52A579}97.2 & \cellcolor[HTML]{CAE89B}83.5 & \cellcolor[HTML]{58B17F}95.4 & \cellcolor[HTML]{91CF8F}89.9 & \cellcolor[HTML]{4A9872}\textbf{99.1} & \cellcolor[HTML]{52A579}97.2 & \cellcolor[HTML]{62B883}\textbf{94.3} & \cellcolor[HTML]{68BC84}93.6 & \cellcolor[HTML]{47926F}100.0 \\
\texttt{rus} & Russian & \cellcolor[HTML]{4A9671}99.3 & \cellcolor[HTML]{47926F}\textbf{99.9} & \cellcolor[HTML]{499570}99.5 & \cellcolor[HTML]{4A9671}99.3 & \cellcolor[HTML]{499570}99.6 & \cellcolor[HTML]{489470}99.8 & \cellcolor[HTML]{47926F}\textbf{99.9} & \cellcolor[HTML]{4C9A73}98.7 & \cellcolor[HTML]{4C9A73}98.7 & \cellcolor[HTML]{4B9973}98.8 & \cellcolor[HTML]{4A9671}99.3 & \cellcolor[HTML]{47926F}99.8 & \cellcolor[HTML]{4A9671}\textbf{99.4} & \cellcolor[HTML]{50A377}97.6 & \cellcolor[HTML]{4A9872}99.1 \\
\texttt{srp} & Serbian & \cellcolor[HTML]{9AD390}88.9 & \cellcolor[HTML]{9AD390}88.9 & \cellcolor[HTML]{FEDBA0}66.7 & \cellcolor[HTML]{5FB782}94.4 & \cellcolor[HTML]{5FB782}94.4 & \cellcolor[HTML]{47926F}\textbf{100.0} & \cellcolor[HTML]{47926F}\textbf{100.0} & \cellcolor[HTML]{FFF5C0}72.2 & \cellcolor[HTML]{CBE99C}83.3 & \cellcolor[HTML]{47926F}\textbf{100.0} & \cellcolor[HTML]{47926F}\textbf{100.0} & \cellcolor[HTML]{47926F}\textbf{100.0} & \cellcolor[HTML]{88CB8D}\textbf{90.7} & \cellcolor[HTML]{9AD390}88.9 & \cellcolor[HTML]{5FB782}94.4 \\
\texttt{slk} & Slovak & \cellcolor[HTML]{E6F4AE}79.6 & \cellcolor[HTML]{75C288}92.5 & \cellcolor[HTML]{8ACC8D}90.5 & \cellcolor[HTML]{C4E697}84.4 & \cellcolor[HTML]{5DB682}94.6 & \cellcolor[HTML]{59B380}95.2 & \cellcolor[HTML]{4C9A73}\textbf{98.6} & \cellcolor[HTML]{ACDB92}87.1 & \cellcolor[HTML]{A5D891}87.8 & \cellcolor[HTML]{7BC589}91.8 & \cellcolor[HTML]{4FA076}98.0 & \cellcolor[HTML]{4FA076}98.0 & \cellcolor[HTML]{80C78A}\textbf{91.5} & \cellcolor[HTML]{A5D891}87.8 & \cellcolor[HTML]{4A9671}99.3 \\
\texttt{slv} & Slovenian & \cellcolor[HTML]{CDEA9D}83.2 & \cellcolor[HTML]{88CB8D}90.8 & \cellcolor[HTML]{D8EFA4}81.5 & \cellcolor[HTML]{B0DD92}86.6 & \cellcolor[HTML]{75C288}92.4 & \cellcolor[HTML]{50A377}97.5 & \cellcolor[HTML]{47926F}\textbf{100.0} & \cellcolor[HTML]{C5E698}84.0 & \cellcolor[HTML]{91CF8F}89.9 & \cellcolor[HTML]{91CF8F}89.9 & \cellcolor[HTML]{5AB481}95.0 & \cellcolor[HTML]{54A97B}96.6 & \cellcolor[HTML]{8ACC8D}\textbf{90.6} & \cellcolor[HTML]{75C288}92.4 & \cellcolor[HTML]{4A9872}99.2 \\
\texttt{spa} & Spanish & \cellcolor[HTML]{59B380}95.2 & \cellcolor[HTML]{4D9C74}98.6 & \cellcolor[HTML]{4D9C74}98.6 & \cellcolor[HTML]{4FA076}97.9 & \cellcolor[HTML]{4A9671}99.3 & \cellcolor[HTML]{4A9671}99.3 & \cellcolor[HTML]{4A9671}99.3 & \cellcolor[HTML]{66BB84}93.8 & \cellcolor[HTML]{75C288}92.4 & \cellcolor[HTML]{6FBF86}93.1 & \cellcolor[HTML]{47926F}\textbf{100.0} & \cellcolor[HTML]{4FA076}97.9 & \cellcolor[HTML]{52A579}\textbf{97.1} & \cellcolor[HTML]{7EC68A}91.7 & \cellcolor[HTML]{4D9C74}98.6 \\
\texttt{swe} & Swedish & \cellcolor[HTML]{50A377}97.6 & \cellcolor[HTML]{47926F}\textbf{100.0} & \cellcolor[HTML]{71C087}92.9 & \cellcolor[HTML]{50A377}97.6 & \cellcolor[HTML]{50A377}97.6 & \cellcolor[HTML]{50A377}97.6 & \cellcolor[HTML]{50A377}97.6 & \cellcolor[HTML]{8ACC8D}90.5 & \cellcolor[HTML]{59B380}95.2 & \cellcolor[HTML]{8ACC8D}90.5 & \cellcolor[HTML]{59B380}95.2 & \cellcolor[HTML]{50A377}97.6 & \cellcolor[HTML]{57AF7E}\textbf{95.8} & \cellcolor[HTML]{47926F}100.0 & \cellcolor[HTML]{59B380}95.2 \\
\texttt{tur} & Turkish & \cellcolor[HTML]{52A77A}97.0 & \cellcolor[HTML]{4A9671}99.4 & \cellcolor[HTML]{4A9671}99.4 & \cellcolor[HTML]{4D9C74}98.5 & \cellcolor[HTML]{4D9C74}98.5 & \cellcolor[HTML]{489470}\textbf{99.7} & \cellcolor[HTML]{4A9671}99.4 & \cellcolor[HTML]{59B380}95.1 & \cellcolor[HTML]{4FA076}97.9 & \cellcolor[HTML]{4FA076}97.9 & \cellcolor[HTML]{4D9C74}98.5 & \cellcolor[HTML]{4A9671}99.4 & \cellcolor[HTML]{4D9D75}\textbf{98.4} & \cellcolor[HTML]{DAEFA5}81.4 & \cellcolor[HTML]{4A9872}99.1 \\
\texttt{ukr} & Ukrainian & \cellcolor[HTML]{54A97B}96.5 & \cellcolor[HTML]{4B9973}99.0 & \cellcolor[HTML]{4FA076}98.0 & \cellcolor[HTML]{4D9C74}98.5 & \cellcolor[HTML]{499570}\textbf{99.5} & \cellcolor[HTML]{4B9973}99.0 & \cellcolor[HTML]{499570}\textbf{99.5} & \cellcolor[HTML]{79C489}92.1 & \cellcolor[HTML]{5AB481}95.0 & \cellcolor[HTML]{58B07F}95.5 & \cellcolor[HTML]{50A377}97.5 & \cellcolor[HTML]{4D9C74}98.5 & \cellcolor[HTML]{51A478}\textbf{97.4} & \cellcolor[HTML]{56AD7D}96.0 & \cellcolor[HTML]{4B9973}99.0 \\
\texttt{urd} & Urdu & \cellcolor[HTML]{47926F}\textbf{100.0} & \cellcolor[HTML]{47926F}\textbf{100.0} & \cellcolor[HTML]{47926F}\textbf{100.0} & \cellcolor[HTML]{47926F}\textbf{100.0} & \cellcolor[HTML]{52A77A}97.0 & \cellcolor[HTML]{47926F}\textbf{100.0} & \cellcolor[HTML]{47926F}\textbf{100.0} & \cellcolor[HTML]{52A77A}97.0 & \cellcolor[HTML]{47926F}\textbf{100.0} & \cellcolor[HTML]{52A77A}97.0 & \cellcolor[HTML]{47926F}\textbf{100.0} & \cellcolor[HTML]{47926F}\textbf{100.0} & \cellcolor[HTML]{4A9671}\textbf{99.2} & \cellcolor[HTML]{52A77A}97.0 & \cellcolor[HTML]{47926F}100.0 \\
\midrule
 & \textbf{Mean} & \cellcolor[HTML]{6FBF86}93.1 & \cellcolor[HTML]{52A77A}96.9 & \cellcolor[HTML]{86CA8C}91.0 & \cellcolor[HTML]{75C288}92.5 & \cellcolor[HTML]{55AC7C}96.1 & \cellcolor[HTML]{4FA076}97.9 & \cellcolor[HTML]{4B9973}\textbf{98.9} & \cellcolor[HTML]{9FD690}88.4 & \cellcolor[HTML]{7EC68A}91.8 & \cellcolor[HTML]{6ABD85}93.5 & \cellcolor[HTML]{55AC7C}96.3 & \cellcolor[HTML]{4FA076}97.9 & \cellcolor[HTML]{5FB782}\textbf{94.5} & \cellcolor[HTML]{6ABD85}93.5 & \cellcolor[HTML]{4D9C74}98.5 \\
\bottomrule
\end{tabular}
}
\caption{Accuracy (\%) on items the 5-gram baseline also gets right, by language and model (8,456 pairs in 33 languages). \emph{Mean} averages the multilingual models to its left; Goldfish sits to its right as a monolingual reference point and is excluded from it, as is the bolding. Best multilingual model per language in bold. The 5-gram is absent by construction: the split is defined by its own verdict.}\label{tab:acc-ngram-solvable}
\end{table*}
\begin{table*}[t]
\centering
\scriptsize
\setlength{\tabcolsep}{3.2pt}
\resizebox{\textwidth}{!}{%
\begin{tabular}{l l | cccccccccccc | c | cc}
\toprule
 &  & \multicolumn{2}{c}{Llama-3.1} & \multicolumn{1}{c}{Aya} & \multicolumn{4}{c}{Gemma-3} & \multicolumn{3}{c}{Qwen3} & \multicolumn{2}{c}{Apertus} & \multicolumn{1}{c}{\textit{Mean}} & \multicolumn{2}{c}{Goldfish} \\
\cmidrule(lr){3-4} \cmidrule(lr){5-5} \cmidrule(lr){6-9} \cmidrule(lr){10-12} \cmidrule(lr){13-14} \cmidrule(lr){16-17}
\texttt{ISO} & Language & \texttt{8B} & \texttt{70B} & \texttt{32B} & \texttt{1B} & \texttt{4B} & \texttt{12B} & \texttt{27B} & \texttt{8B} & \texttt{14B} & \texttt{32B} & \texttt{8B} & \texttt{70B} &  & \texttt{100MB} & \texttt{1GB} \\
\midrule
\texttt{hye} & Armenian & \cellcolor[HTML]{71C087}92.9 & \cellcolor[HTML]{B9E193}85.7 & \cellcolor[HTML]{71C087}92.9 & \cellcolor[HTML]{B9E193}85.7 & \cellcolor[HTML]{71C087}92.9 & \cellcolor[HTML]{B9E193}85.7 & \cellcolor[HTML]{71C087}92.9 & \cellcolor[HTML]{47926F}\textbf{100.0} & \cellcolor[HTML]{B9E193}85.7 & \cellcolor[HTML]{B9E193}85.7 & \cellcolor[HTML]{B9E193}85.7 & \cellcolor[HTML]{71C087}92.9 & \cellcolor[HTML]{91CF8F}\textbf{89.9} & \cellcolor[HTML]{FECC94}64.3 & \cellcolor[HTML]{B9E193}85.7 \\
\texttt{bel} & Belarusian & \cellcolor[HTML]{E1F2AA}80.4 & \cellcolor[HTML]{82C88B}91.3 & \cellcolor[HTML]{C0E494}84.8 & \cellcolor[HTML]{F9FDC9}76.1 & \cellcolor[HTML]{98D38F}89.1 & \cellcolor[HTML]{6ABD85}93.5 & \cellcolor[HTML]{58B07F}95.7 & \cellcolor[HTML]{FFFBCA}73.9 & \cellcolor[HTML]{98D38F}89.1 & \cellcolor[HTML]{E1F2AA}80.4 & \cellcolor[HTML]{47926F}\textbf{100.0} & \cellcolor[HTML]{6ABD85}93.5 & \cellcolor[HTML]{A8DA91}\textbf{87.3} & \cellcolor[HTML]{F9FDC9}76.1 & \cellcolor[HTML]{6ABD85}93.5 \\
\texttt{bul} & Bulgarian & \cellcolor[HTML]{FFF9C7}73.5 & \cellcolor[HTML]{BCE293}85.3 & \cellcolor[HTML]{C95263}47.1 & \cellcolor[HTML]{E7F5B0}79.4 & \cellcolor[HTML]{64BA83}94.1 & \cellcolor[HTML]{A1D790}88.2 & \cellcolor[HTML]{47926F}\textbf{100.0} & \cellcolor[HTML]{BCE293}85.3 & \cellcolor[HTML]{F7FCC6}76.5 & \cellcolor[HTML]{F7FCC6}76.5 & \cellcolor[HTML]{64BA83}94.1 & \cellcolor[HTML]{52A77A}97.1 & \cellcolor[HTML]{CDEA9D}\textbf{83.1} & \cellcolor[HTML]{FFF9C7}73.5 & \cellcolor[HTML]{84C98B}91.2 \\
\texttt{cat} & Catalan & \cellcolor[HTML]{71C087}92.8 & \cellcolor[HTML]{52A77A}96.9 & \cellcolor[HTML]{66BB84}93.8 & \cellcolor[HTML]{88CB8D}90.7 & \cellcolor[HTML]{66BB84}93.8 & \cellcolor[HTML]{57AF7E}95.9 & \cellcolor[HTML]{4B9973}\textbf{99.0} & \cellcolor[HTML]{7EC68A}91.8 & \cellcolor[HTML]{7EC68A}91.8 & \cellcolor[HTML]{71C087}92.8 & \cellcolor[HTML]{52A77A}96.9 & \cellcolor[HTML]{66BB84}93.8 & \cellcolor[HTML]{62B883}\textbf{94.2} & \cellcolor[HTML]{5BB581}94.8 & \cellcolor[HTML]{5BB581}94.8 \\
\texttt{zho} & Chinese & \cellcolor[HTML]{47926F}\textbf{100.0} & \cellcolor[HTML]{9AD390}88.9 & \cellcolor[HTML]{EFF8BC}77.8 & \cellcolor[HTML]{FEDBA0}66.7 & \cellcolor[HTML]{EFF8BC}77.8 & \cellcolor[HTML]{EFF8BC}77.8 & \cellcolor[HTML]{9AD390}88.9 & \cellcolor[HTML]{EFF8BC}77.8 & \cellcolor[HTML]{9AD390}88.9 & \cellcolor[HTML]{47926F}\textbf{100.0} & \cellcolor[HTML]{9AD390}88.9 & \cellcolor[HTML]{EFF8BC}77.8 & \cellcolor[HTML]{C4E697}\textbf{84.3} & \cellcolor[HTML]{BE4763}44.4 & \cellcolor[HTML]{FEDBA0}66.7 \\
\texttt{hrv} & Croatian & \cellcolor[HTML]{73C187}92.6 & \cellcolor[HTML]{73C187}92.6 & \cellcolor[HTML]{D8EFA4}81.5 & \cellcolor[HTML]{BCE293}85.2 & \cellcolor[HTML]{73C187}92.6 & \cellcolor[HTML]{55AB7C}96.3 & \cellcolor[HTML]{47926F}\textbf{100.0} & \cellcolor[HTML]{EFF8BC}77.8 & \cellcolor[HTML]{EFF8BC}77.8 & \cellcolor[HTML]{D8EFA4}81.5 & \cellcolor[HTML]{55AB7C}96.3 & \cellcolor[HTML]{47926F}\textbf{100.0} & \cellcolor[HTML]{94D18F}\textbf{89.5} & \cellcolor[HTML]{9AD390}88.9 & \cellcolor[HTML]{73C187}92.6 \\
\texttt{ces} & Czech & \cellcolor[HTML]{F1F9BD}77.7 & \cellcolor[HTML]{84C98B}91.1 & \cellcolor[HTML]{73C187}92.6 & \cellcolor[HTML]{FEE7AA}68.8 & \cellcolor[HTML]{94D18F}89.5 & \cellcolor[HTML]{77C388}92.3 & \cellcolor[HTML]{5AB481}95.0 & \cellcolor[HTML]{FFF1BA}71.4 & \cellcolor[HTML]{F2F9BF}77.5 & \cellcolor[HTML]{C3E596}84.5 & \cellcolor[HTML]{77C388}92.3 & \cellcolor[HTML]{55AC7C}\textbf{96.1} & \cellcolor[HTML]{B9E193}\textbf{85.7} & \cellcolor[HTML]{F79878}57.4 & \cellcolor[HTML]{B5DF93}86.0 \\
\texttt{dan} & Danish & \cellcolor[HTML]{D7EEA3}81.8 & \cellcolor[HTML]{D7EEA3}81.8 & \cellcolor[HTML]{D7EEA3}81.8 & \cellcolor[HTML]{FFF7C3}72.7 & \cellcolor[HTML]{D7EEA3}81.8 & \cellcolor[HTML]{47926F}\textbf{100.0} & \cellcolor[HTML]{47926F}\textbf{100.0} & \cellcolor[HTML]{D7EEA3}81.8 & \cellcolor[HTML]{47926F}\textbf{100.0} & \cellcolor[HTML]{86CA8C}90.9 & \cellcolor[HTML]{D7EEA3}81.8 & \cellcolor[HTML]{86CA8C}90.9 & \cellcolor[HTML]{AADB91}\textbf{87.1} & \cellcolor[HTML]{D7EEA3}81.8 & \cellcolor[HTML]{FEC890}63.6 \\
\texttt{nld} & Dutch & \cellcolor[HTML]{DCF0A7}81.0 & \cellcolor[HTML]{82C88B}91.3 & \cellcolor[HTML]{B7E093}85.8 & \cellcolor[HTML]{FFFBCA}74.0 & \cellcolor[HTML]{D2ECA0}82.4 & \cellcolor[HTML]{94D18F}89.6 & \cellcolor[HTML]{7EC68A}\textbf{91.8} & \cellcolor[HTML]{FEE3A7}68.1 & \cellcolor[HTML]{FFF6C2}72.6 & \cellcolor[HTML]{EAF6B4}78.7 & \cellcolor[HTML]{CEEA9E}83.0 & \cellcolor[HTML]{8FCE8E}90.2 & \cellcolor[HTML]{D2ECA0}\textbf{82.4} & \cellcolor[HTML]{BE4763}43.1 & \cellcolor[HTML]{FFF3BD}71.8 \\
\texttt{eng} & English & \cellcolor[HTML]{50A377}\textbf{97.5} & \cellcolor[HTML]{50A377}\textbf{97.5} & \cellcolor[HTML]{55AC7C}96.2 & \cellcolor[HTML]{5AB481}95.0 & \cellcolor[HTML]{55AC7C}96.2 & \cellcolor[HTML]{55AC7C}96.2 & \cellcolor[HTML]{50A377}\textbf{97.5} & \cellcolor[HTML]{55AC7C}96.2 & \cellcolor[HTML]{50A377}\textbf{97.5} & \cellcolor[HTML]{55AC7C}96.2 & \cellcolor[HTML]{50A377}\textbf{97.5} & \cellcolor[HTML]{55AC7C}96.2 & \cellcolor[HTML]{54A97B}\textbf{96.7} & \cellcolor[HTML]{FFF1BA}71.2 & \cellcolor[HTML]{D1EB9F}82.5 \\
\texttt{fin} & Finnish & \cellcolor[HTML]{CEEA9E}82.9 & \cellcolor[HTML]{BAE293}85.4 & \cellcolor[HTML]{FEC78F}63.4 & \cellcolor[HTML]{EEF8BA}78.0 & \cellcolor[HTML]{FBFDCC}75.6 & \cellcolor[HTML]{8DCD8E}\textbf{90.2} & \cellcolor[HTML]{DFF2A9}80.5 & \cellcolor[HTML]{FBB586}61.0 & \cellcolor[HTML]{FEE4A8}68.3 & \cellcolor[HTML]{EEF8BA}78.0 & \cellcolor[HTML]{BAE293}85.4 & \cellcolor[HTML]{8DCD8E}\textbf{90.2} & \cellcolor[HTML]{EDF8B9}\textbf{78.3} & \cellcolor[HTML]{FBFDCC}75.6 & \cellcolor[HTML]{50A377}97.6 \\
\texttt{deu} & German & \cellcolor[HTML]{6FBF86}93.1 & \cellcolor[HTML]{52A579}97.2 & \cellcolor[HTML]{6FBF86}93.1 & \cellcolor[HTML]{9AD390}88.9 & \cellcolor[HTML]{7EC68A}91.7 & \cellcolor[HTML]{5FB782}94.4 & \cellcolor[HTML]{52A579}97.2 & \cellcolor[HTML]{6FBF86}93.1 & \cellcolor[HTML]{52A579}97.2 & \cellcolor[HTML]{5FB782}94.4 & \cellcolor[HTML]{5FB782}94.4 & \cellcolor[HTML]{4D9C74}\textbf{98.6} & \cellcolor[HTML]{5FB782}\textbf{94.4} & \cellcolor[HTML]{FAAA81}59.7 & \cellcolor[HTML]{A7D991}87.5 \\
\texttt{ell} & Greek & \cellcolor[HTML]{FEC58E}63.2 & \cellcolor[HTML]{94D18F}89.5 & \cellcolor[HTML]{51A478}\textbf{97.4} & \cellcolor[HTML]{E9F6B3}78.9 & \cellcolor[HTML]{C4E697}84.2 & \cellcolor[HTML]{94D18F}89.5 & \cellcolor[HTML]{5BB581}94.7 & \cellcolor[HTML]{FEC58E}63.2 & \cellcolor[HTML]{FEE4A8}68.4 & \cellcolor[HTML]{E9F6B3}78.9 & \cellcolor[HTML]{94D18F}89.5 & \cellcolor[HTML]{5BB581}94.7 & \cellcolor[HTML]{CFEB9E}\textbf{82.7} & \cellcolor[HTML]{FFFAC9}73.7 & \cellcolor[HTML]{AEDC92}86.8 \\
\texttt{heb} & Hebrew & \cellcolor[HTML]{B5DF93}86.0 & \cellcolor[HTML]{B5DF93}86.0 & \cellcolor[HTML]{88CB8D}90.7 & \cellcolor[HTML]{F6FBC4}76.7 & \cellcolor[HTML]{B5DF93}86.0 & \cellcolor[HTML]{9FD690}88.4 & \cellcolor[HTML]{88CB8D}90.7 & \cellcolor[HTML]{E9F6B3}79.1 & \cellcolor[HTML]{FFFDCD}74.4 & \cellcolor[HTML]{FFF4BF}72.1 & \cellcolor[HTML]{6FBF86}\textbf{93.0} & \cellcolor[HTML]{88CB8D}90.7 & \cellcolor[HTML]{C3E596}\textbf{84.5} & \cellcolor[HTML]{FEEBB0}69.8 & \cellcolor[HTML]{B5DF93}86.0 \\
\texttt{hin} & Hindi & \cellcolor[HTML]{86CA8C}90.9 & \cellcolor[HTML]{4E9E75}\textbf{98.2} & \cellcolor[HTML]{55AB7C}96.4 & \cellcolor[HTML]{86CA8C}90.9 & \cellcolor[HTML]{73C187}92.7 & \cellcolor[HTML]{4E9E75}\textbf{98.2} & \cellcolor[HTML]{4E9E75}\textbf{98.2} & \cellcolor[HTML]{86CA8C}90.9 & \cellcolor[HTML]{86CA8C}90.9 & \cellcolor[HTML]{73C187}92.7 & \cellcolor[HTML]{98D38F}89.1 & \cellcolor[HTML]{5DB682}94.5 & \cellcolor[HTML]{68BC84}\textbf{93.6} & \cellcolor[HTML]{FFF0B7}70.9 & \cellcolor[HTML]{BAE293}85.5 \\
\texttt{hun} & Hungarian & \cellcolor[HTML]{47926F}\textbf{100.0} & \cellcolor[HTML]{47926F}\textbf{100.0} & \cellcolor[HTML]{47926F}\textbf{100.0} & \cellcolor[HTML]{47926F}\textbf{100.0} & \cellcolor[HTML]{BE4763}0.0 & \cellcolor[HTML]{47926F}\textbf{100.0} & \cellcolor[HTML]{47926F}\textbf{100.0} & \cellcolor[HTML]{47926F}\textbf{100.0} & \cellcolor[HTML]{47926F}\textbf{100.0} & \cellcolor[HTML]{47926F}\textbf{100.0} & \cellcolor[HTML]{47926F}\textbf{100.0} & \cellcolor[HTML]{47926F}\textbf{100.0} & \cellcolor[HTML]{7EC68A}\textbf{91.7} & \cellcolor[HTML]{47926F}100.0 & \cellcolor[HTML]{47926F}100.0 \\
\texttt{isl} & Icelandic & \cellcolor[HTML]{F9A67F}59.1 & \cellcolor[HTML]{D7EEA3}81.8 & \cellcolor[HTML]{FEC890}63.6 & \cellcolor[HTML]{F9A67F}59.1 & \cellcolor[HTML]{86CA8C}\textbf{90.9} & \cellcolor[HTML]{86CA8C}\textbf{90.9} & \cellcolor[HTML]{86CA8C}\textbf{90.9} & \cellcolor[HTML]{FEC890}63.6 & \cellcolor[HTML]{DC6463}50.0 & \cellcolor[HTML]{FFF7C3}72.7 & \cellcolor[HTML]{B1DE92}86.4 & \cellcolor[HTML]{86CA8C}\textbf{90.9} & \cellcolor[HTML]{FEFFD0}\textbf{75.0} & \cellcolor[HTML]{FEC890}63.6 & \cellcolor[HTML]{B1DE92}86.4 \\
\texttt{ind} & Indonesian & \cellcolor[HTML]{FEDBA0}66.7 & \cellcolor[HTML]{47926F}\textbf{100.0} & \cellcolor[HTML]{47926F}\textbf{100.0} & \cellcolor[HTML]{47926F}\textbf{100.0} & \cellcolor[HTML]{47926F}\textbf{100.0} & \cellcolor[HTML]{47926F}\textbf{100.0} & \cellcolor[HTML]{47926F}\textbf{100.0} & \cellcolor[HTML]{47926F}\textbf{100.0} & \cellcolor[HTML]{47926F}\textbf{100.0} & \cellcolor[HTML]{47926F}\textbf{100.0} & \cellcolor[HTML]{47926F}\textbf{100.0} & \cellcolor[HTML]{47926F}\textbf{100.0} & \cellcolor[HTML]{52A579}\textbf{97.2} & \cellcolor[HTML]{47926F}100.0 & \cellcolor[HTML]{47926F}100.0 \\
\texttt{lit} & Lithuanian & \cellcolor[HTML]{FAAC82}60.0 & \cellcolor[HTML]{FAAC82}60.0 & \cellcolor[HTML]{47926F}\textbf{100.0} & \cellcolor[HTML]{BE4763}40.0 & \cellcolor[HTML]{E4F3AB}80.0 & \cellcolor[HTML]{47926F}\textbf{100.0} & \cellcolor[HTML]{47926F}\textbf{100.0} & \cellcolor[HTML]{FAAC82}60.0 & \cellcolor[HTML]{47926F}\textbf{100.0} & \cellcolor[HTML]{E4F3AB}80.0 & \cellcolor[HTML]{47926F}\textbf{100.0} & \cellcolor[HTML]{47926F}\textbf{100.0} & \cellcolor[HTML]{D7EEA3}\textbf{81.7} & \cellcolor[HTML]{FAAC82}60.0 & \cellcolor[HTML]{E4F3AB}80.0 \\
\texttt{nob} & Norwegian-Bokmaal & \cellcolor[HTML]{FAFDCA}75.9 & \cellcolor[HTML]{B3DF92}86.2 & \cellcolor[HTML]{CFEB9E}82.8 & \cellcolor[HTML]{E7F5B0}79.3 & \cellcolor[HTML]{CFEB9E}82.8 & \cellcolor[HTML]{54A97B}96.6 & \cellcolor[HTML]{54A97B}96.6 & \cellcolor[HTML]{FED49A}65.5 & \cellcolor[HTML]{FAFDCA}75.9 & \cellcolor[HTML]{CFEB9E}82.8 & \cellcolor[HTML]{FFF5C0}72.4 & \cellcolor[HTML]{47926F}\textbf{100.0} & \cellcolor[HTML]{CDEA9D}\textbf{83.0} & \cellcolor[HTML]{E7F5B0}79.3 & \cellcolor[HTML]{93D08F}89.7 \\
\texttt{nno} & Norwegian-Nynorsk & \cellcolor[HTML]{DC6463}50.0 & \cellcolor[HTML]{FEDBA0}66.7 & \cellcolor[HTML]{BE4763}41.7 & \cellcolor[HTML]{FEFFD0}75.0 & \cellcolor[HTML]{CBE99C}83.3 & \cellcolor[HTML]{7EC68A}91.7 & \cellcolor[HTML]{47926F}\textbf{100.0} & \cellcolor[HTML]{FEDBA0}66.7 & \cellcolor[HTML]{FEFFD0}75.0 & \cellcolor[HTML]{FEFFD0}75.0 & \cellcolor[HTML]{CBE99C}83.3 & \cellcolor[HTML]{7EC68A}91.7 & \cellcolor[HTML]{FEFFD0}\textbf{75.0} & \cellcolor[HTML]{FEDBA0}66.7 & --- \\
\texttt{pes} & Persian & \cellcolor[HTML]{58B07F}95.6 & \cellcolor[HTML]{4D9D75}98.3 & \cellcolor[HTML]{54A97B}96.6 & \cellcolor[HTML]{AEDC92}86.8 & \cellcolor[HTML]{59B380}95.3 & \cellcolor[HTML]{50A377}97.6 & \cellcolor[HTML]{4C9A73}\textbf{98.6} & \cellcolor[HTML]{E6F4AE}79.7 & \cellcolor[HTML]{A8DA91}87.5 & \cellcolor[HTML]{84C98B}91.2 & \cellcolor[HTML]{51A478}97.3 & \cellcolor[HTML]{4C9A73}\textbf{98.6} & \cellcolor[HTML]{68BC84}\textbf{93.6} & \cellcolor[HTML]{B1DE92}86.4 & \cellcolor[HTML]{55AC7C}96.3 \\
\texttt{pol} & Polish & \cellcolor[HTML]{5BB581}94.9 & \cellcolor[HTML]{66BB84}93.9 & \cellcolor[HTML]{56AD7D}95.9 & \cellcolor[HTML]{96D28F}89.3 & \cellcolor[HTML]{51A478}97.4 & \cellcolor[HTML]{52A77A}96.9 & \cellcolor[HTML]{4D9C74}\textbf{98.5} & \cellcolor[HTML]{71C087}92.9 & \cellcolor[HTML]{66BB84}93.9 & \cellcolor[HTML]{5BB581}94.9 & \cellcolor[HTML]{66BB84}93.9 & \cellcolor[HTML]{55AB7C}96.4 & \cellcolor[HTML]{5BB581}\textbf{94.9} & \cellcolor[HTML]{C8E89A}83.7 & \cellcolor[HTML]{5FB782}94.4 \\
\texttt{ron} & Romanian & \cellcolor[HTML]{D1EB9F}82.5 & \cellcolor[HTML]{82C88B}91.2 & \cellcolor[HTML]{7BC589}92.0 & \cellcolor[HTML]{FFF8C4}73.0 & \cellcolor[HTML]{7BC589}92.0 & \cellcolor[HTML]{5BB581}\textbf{94.9} & \cellcolor[HTML]{62B883}94.2 & \cellcolor[HTML]{F2F9BF}77.4 & \cellcolor[HTML]{AEDC92}86.9 & \cellcolor[HTML]{8ACC8D}90.5 & \cellcolor[HTML]{6ABD85}93.4 & \cellcolor[HTML]{6ABD85}93.4 & \cellcolor[HTML]{9FD690}\textbf{88.4} & \cellcolor[HTML]{C7E799}83.9 & \cellcolor[HTML]{58B07F}95.6 \\
\texttt{rus} & Russian & \cellcolor[HTML]{59B380}95.2 & \cellcolor[HTML]{52A579}97.1 & \cellcolor[HTML]{56AD7D}96.1 & \cellcolor[HTML]{5DB682}94.6 & \cellcolor[HTML]{54A97B}96.6 & \cellcolor[HTML]{4FA177}97.8 & \cellcolor[HTML]{4D9C74}98.5 & \cellcolor[HTML]{68BC84}93.6 & \cellcolor[HTML]{6ABD85}93.4 & \cellcolor[HTML]{5BB581}94.8 & \cellcolor[HTML]{4D9D75}98.2 & \cellcolor[HTML]{4C9A73}\textbf{98.7} & \cellcolor[HTML]{55AC7C}\textbf{96.2} & \cellcolor[HTML]{F2F9BF}77.5 & \cellcolor[HTML]{86CA8C}90.9 \\
\texttt{srp} & Serbian & \cellcolor[HTML]{DC6463}50.0 & \cellcolor[HTML]{47926F}\textbf{100.0} & \cellcolor[HTML]{47926F}\textbf{100.0} & \cellcolor[HTML]{DC6463}50.0 & \cellcolor[HTML]{CBE99C}83.3 & \cellcolor[HTML]{47926F}\textbf{100.0} & \cellcolor[HTML]{47926F}\textbf{100.0} & \cellcolor[HTML]{CBE99C}83.3 & \cellcolor[HTML]{CBE99C}83.3 & \cellcolor[HTML]{CBE99C}83.3 & \cellcolor[HTML]{CBE99C}83.3 & \cellcolor[HTML]{47926F}\textbf{100.0} & \cellcolor[HTML]{C1E595}\textbf{84.7} & \cellcolor[HTML]{FEDBA0}66.7 & \cellcolor[HTML]{FEDBA0}66.7 \\
\texttt{slk} & Slovak & \cellcolor[HTML]{F89B7A}57.7 & \cellcolor[HTML]{FEFFD0}75.2 & \cellcolor[HTML]{FEFFD0}75.2 & \cellcolor[HTML]{FED59B}65.7 & \cellcolor[HTML]{E6F4AE}79.6 & \cellcolor[HTML]{A7D991}87.6 & \cellcolor[HTML]{6ABD85}\textbf{93.4} & \cellcolor[HTML]{FED99E}66.4 & \cellcolor[HTML]{FEECB1}70.1 & \cellcolor[HTML]{FFFDCD}74.5 & \cellcolor[HTML]{93D08F}89.8 & \cellcolor[HTML]{7BC589}92.0 & \cellcolor[HTML]{F3FAC0}\textbf{77.3} & \cellcolor[HTML]{FEDDA2}67.2 & \cellcolor[HTML]{73C187}92.7 \\
\texttt{slv} & Slovenian & \cellcolor[HTML]{FFF1BA}71.4 & \cellcolor[HTML]{F1F9BD}77.6 & \cellcolor[HTML]{FEEAAE}69.4 & \cellcolor[HTML]{FEDFA3}67.3 & \cellcolor[HTML]{7BC589}91.8 & \cellcolor[HTML]{93D08F}89.8 & \cellcolor[HTML]{66BB84}93.9 & \cellcolor[HTML]{FFF1BA}71.4 & \cellcolor[HTML]{FCFECD}75.5 & \cellcolor[HTML]{D8EFA4}81.6 & \cellcolor[HTML]{C8E89A}83.7 & \cellcolor[HTML]{56AD7D}\textbf{95.9} & \cellcolor[HTML]{DEF1A8}\textbf{80.8} & \cellcolor[HTML]{FFF9C7}73.5 & \cellcolor[HTML]{66BB84}93.9 \\
\texttt{spa} & Spanish & \cellcolor[HTML]{8ACC8D}90.6 & \cellcolor[HTML]{53A87A}96.9 & \cellcolor[HTML]{50A377}\textbf{97.6} & \cellcolor[HTML]{71C087}92.9 & \cellcolor[HTML]{71C087}92.9 & \cellcolor[HTML]{53A87A}96.9 & \cellcolor[HTML]{53A87A}96.9 & \cellcolor[HTML]{53A87A}96.9 & \cellcolor[HTML]{50A377}\textbf{97.6} & \cellcolor[HTML]{5FB782}94.5 & \cellcolor[HTML]{59B380}95.3 & \cellcolor[HTML]{79C489}92.1 & \cellcolor[HTML]{5AB481}\textbf{95.1} & \cellcolor[HTML]{D5EDA2}81.9 & \cellcolor[HTML]{79C489}92.1 \\
\texttt{swe} & Swedish & \cellcolor[HTML]{64BA83}94.1 & \cellcolor[HTML]{47926F}\textbf{100.0} & \cellcolor[HTML]{F7FCC6}76.5 & \cellcolor[HTML]{A1D790}88.2 & \cellcolor[HTML]{47926F}\textbf{100.0} & \cellcolor[HTML]{47926F}\textbf{100.0} & \cellcolor[HTML]{47926F}\textbf{100.0} & \cellcolor[HTML]{F7FCC6}76.5 & \cellcolor[HTML]{64BA83}94.1 & \cellcolor[HTML]{47926F}\textbf{100.0} & \cellcolor[HTML]{47926F}\textbf{100.0} & \cellcolor[HTML]{64BA83}94.1 & \cellcolor[HTML]{68BC84}\textbf{93.6} & \cellcolor[HTML]{D2ECA0}82.4 & \cellcolor[HTML]{A1D790}88.2 \\
\texttt{tur} & Turkish & \cellcolor[HTML]{9CD490}88.8 & \cellcolor[HTML]{64BA83}94.1 & \cellcolor[HTML]{5FB782}94.5 & \cellcolor[HTML]{93D08F}89.7 & \cellcolor[HTML]{6ABD85}93.5 & \cellcolor[HTML]{55AC7C}96.2 & \cellcolor[HTML]{52A77A}\textbf{97.0} & \cellcolor[HTML]{C7E799}84.0 & \cellcolor[HTML]{96D28F}89.5 & \cellcolor[HTML]{96D28F}89.5 & \cellcolor[HTML]{5AB481}94.9 & \cellcolor[HTML]{58B07F}95.6 & \cellcolor[HTML]{77C388}\textbf{92.3} & \cellcolor[HTML]{E06863}50.8 & \cellcolor[HTML]{98D38F}89.2 \\
\texttt{ukr} & Ukrainian & \cellcolor[HTML]{5DB682}94.6 & \cellcolor[HTML]{58B07F}95.7 & \cellcolor[HTML]{6ABD85}93.5 & \cellcolor[HTML]{CFEB9E}82.8 & \cellcolor[HTML]{82C88B}91.4 & \cellcolor[HTML]{58B07F}95.7 & \cellcolor[HTML]{4B9973}\textbf{98.9} & \cellcolor[HTML]{82C88B}91.4 & \cellcolor[HTML]{ACDB92}87.1 & \cellcolor[HTML]{A1D790}88.2 & \cellcolor[HTML]{5DB682}94.6 & \cellcolor[HTML]{58B07F}95.7 & \cellcolor[HTML]{75C288}\textbf{92.5} & \cellcolor[HTML]{FFF0B7}71.0 & \cellcolor[HTML]{82C88B}91.4 \\
\texttt{urd} & Urdu & \cellcolor[HTML]{EFF8BC}77.8 & \cellcolor[HTML]{55AB7C}96.3 & \cellcolor[HTML]{BCE293}85.2 & \cellcolor[HTML]{EFF8BC}77.8 & \cellcolor[HTML]{BCE293}85.2 & \cellcolor[HTML]{47926F}\textbf{100.0} & \cellcolor[HTML]{73C187}92.6 & \cellcolor[HTML]{EFF8BC}77.8 & \cellcolor[HTML]{BCE293}85.2 & \cellcolor[HTML]{73C187}92.6 & \cellcolor[HTML]{9AD390}88.9 & \cellcolor[HTML]{55AB7C}96.3 & \cellcolor[HTML]{A3D791}\textbf{88.0} & \cellcolor[HTML]{9AD390}88.9 & \cellcolor[HTML]{9AD390}88.9 \\
\midrule
 & \textbf{Mean} & \cellcolor[HTML]{D8EFA4}81.5 & \cellcolor[HTML]{91CF8F}90.0 & \cellcolor[HTML]{B7E093}85.9 & \cellcolor[HTML]{E7F5B0}79.4 & \cellcolor[HTML]{B0DD92}86.6 & \cellcolor[HTML]{62B883}94.2 & \cellcolor[HTML]{55AC7C}\textbf{96.1} & \cellcolor[HTML]{DFF2A9}80.6 & \cellcolor[HTML]{C0E494}84.9 & \cellcolor[HTML]{ACDB92}87.0 & \cellcolor[HTML]{7EC68A}91.6 & \cellcolor[HTML]{5BB581}94.8 & \cellcolor[HTML]{A5D891}\textbf{87.7} & \cellcolor[HTML]{FFFAC9}73.6 & \cellcolor[HTML]{A5D891}87.8 \\
\bottomrule
\end{tabular}
}
\caption{Accuracy (\%) on items the 5-gram baseline gets wrong, by language and model (4,356 pairs in 33 languages). \emph{Mean} averages the multilingual models to its left; Goldfish sits to its right as a monolingual reference point and is excluded from it, as is the bolding. Best multilingual model per language in bold. The 5-gram is absent by construction: the split is defined by its own verdict.}\label{tab:acc-ngram-unsolvable}
\end{table*}

\section{Distance}
\label{app:distance}

\begin{figure}[t]
  \centering
  \includegraphics[width=\columnwidth]{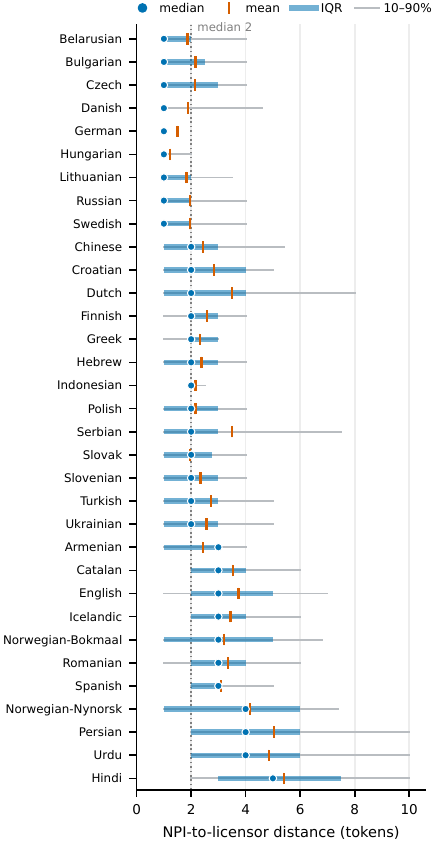}
  \caption{Distribution of NPI–licenser distance (in tokens) per language.}
  \label{fig:dist-by-lang}
\end{figure}

Figure~\ref{fig:dist-by-lang} shows the mean NPI--licenser distance per language.

\section{MultiBLiMP Comparison}
\label{app:multiblimp}

Table~\ref{tab:multiblimp-compare} compares our accuracies with the published MultiBLiMP scores for the shared models and languages.

\begin{table*}[t]
\centering
\small
\setlength{\tabcolsep}{3.2pt}
\resizebox{\textwidth}{!}{%
\begin{tabular}{l r r | cc | cc | cc | cc | cc | cc | c c c}
\toprule
& & & \multicolumn{2}{c}{\texttt{Goldfish\,1GB}} & \multicolumn{2}{c}{\texttt{Llama\,8B}} & \multicolumn{2}{c}{\texttt{EuroLLM\,9B}} & \multicolumn{2}{c}{\texttt{Gemma\,27B}} & \multicolumn{2}{c}{\texttt{OLMo\,32B}} & \multicolumn{2}{c}{\texttt{Llama\,70B}} & \multicolumn{3}{c}{Mean} \\
\cmidrule(lr){4-5}\cmidrule(lr){6-7}\cmidrule(lr){8-9}\cmidrule(lr){10-11}\cmidrule(lr){12-13}\cmidrule(lr){14-15}\cmidrule(lr){16-18}
Language & Pairs\_B & Pairs\_M & \textsc{b} & \textsc{m} & \textsc{b} & \textsc{m} & \textsc{b} & \textsc{m} & \textsc{b} & \textsc{m} & \textsc{b} & \textsc{m} & \textsc{b} & \textsc{m} & \textsc{b} & \textsc{m} & $\Delta$ \\
\midrule
Icelandic & 47 & 2,801 & \cellcolor[HTML]{80C78A}91.5 & \cellcolor[HTML]{4D9D75}98.4 & \cellcolor[HTML]{FEE3A7}68.1 & \cellcolor[HTML]{BCE293}85.3 & \cellcolor[HTML]{BE4763}6.4 & \cellcolor[HTML]{FEE2A6}67.8 & \cellcolor[HTML]{68BC84}93.6 & \cellcolor[HTML]{56AD7D}96.0 & \cellcolor[HTML]{F18A72}55.3 & \cellcolor[HTML]{E5F4AD}79.7 & \cellcolor[HTML]{BEE394}85.1 & \cellcolor[HTML]{86CA8C}91.0 & \cellcolor[HTML]{FEDBA0}66.7 & \cellcolor[HTML]{B1DE92}86.4 & \cellcolor[HTML]{69497F}$-$19.7 \\
Armenian & 42 & 1,415 & \cellcolor[HTML]{59B380}95.2 & \cellcolor[HTML]{4D9D75}98.4 & \cellcolor[HTML]{59B380}95.2 & \cellcolor[HTML]{54A97B}96.5 & \cellcolor[HTML]{BE4763}14.3 & \cellcolor[HTML]{FFF7C3}72.7 & \cellcolor[HTML]{50A377}97.6 & \cellcolor[HTML]{4FA076}97.9 & \cellcolor[HTML]{C04963}45.2 & \cellcolor[HTML]{AEDC92}86.9 & \cellcolor[HTML]{71C087}92.9 & \cellcolor[HTML]{4D9D75}98.4 & \cellcolor[HTML]{FFF9C7}73.4 & \cellcolor[HTML]{7BC589}91.8 & \cellcolor[HTML]{73528F}$-$18.4 \\
Greek & 87 & 1,096 & \cellcolor[HTML]{6FBF86}93.1 & \cellcolor[HTML]{4C9A73}98.8 & \cellcolor[HTML]{E7F5B0}79.3 & \cellcolor[HTML]{4FA076}97.9 & \cellcolor[HTML]{58B17F}95.4 & \cellcolor[HTML]{4A9671}99.3 & \cellcolor[HTML]{54A97B}96.6 & \cellcolor[HTML]{499570}99.5 & \cellcolor[HTML]{FEE9AB}69.0 & \cellcolor[HTML]{52A77A}96.9 & \cellcolor[HTML]{7BC589}92.0 & \cellcolor[HTML]{4A9671}99.3 & \cellcolor[HTML]{A7D991}87.5 & \cellcolor[HTML]{4D9C74}98.6 & \cellcolor[HTML]{ACA4CA}$-$11.1 \\
Slovak & 294 & 4,145 & \cellcolor[HTML]{55AC7C}96.3 & \cellcolor[HTML]{59B380}95.2 & \cellcolor[HTML]{FEE3A7}68.0 & \cellcolor[HTML]{BAE293}85.5 & \cellcolor[HTML]{93D08F}89.8 & \cellcolor[HTML]{55AB7C}96.3 & \cellcolor[HTML]{58B07F}95.6 & \cellcolor[HTML]{58B07F}95.7 & \cellcolor[HTML]{DD6563}50.3 & \cellcolor[HTML]{F3FAC0}77.2 & \cellcolor[HTML]{C4E697}84.4 & \cellcolor[HTML]{7BC589}91.8 & \cellcolor[HTML]{DEF1A8}80.7 & \cellcolor[HTML]{8DCD8E}90.3 & \cellcolor[HTML]{B9B2D3}$-$9.6 \\
Urdu & 60 & 550 & \cellcolor[HTML]{5AB481}95.0 & \cellcolor[HTML]{55AB7C}96.4 & \cellcolor[HTML]{91CF8F}90.0 & \cellcolor[HTML]{53A87A}96.7 & \cellcolor[HTML]{EA7A6B}53.3 & \cellcolor[HTML]{A7D991}87.6 & \cellcolor[HTML]{54A97B}96.7 & \cellcolor[HTML]{50A377}97.5 & \cellcolor[HTML]{CBE99C}83.3 & \cellcolor[HTML]{54A97B}96.5 & \cellcolor[HTML]{4D9D75}98.3 & \cellcolor[HTML]{50A377}97.5 & \cellcolor[HTML]{B5DF93}86.1 & \cellcolor[HTML]{58B17F}95.4 & \cellcolor[HTML]{BCB5D5}$-$9.3 \\
Dutch & 1,771 & 2,331 & \cellcolor[HTML]{B0DD92}86.6 & \cellcolor[HTML]{51A478}97.3 & \cellcolor[HTML]{A5D891}87.8 & \cellcolor[HTML]{55AC7C}96.1 & \cellcolor[HTML]{6ABD85}93.5 & \cellcolor[HTML]{4E9E75}98.2 & \cellcolor[HTML]{57AF7E}95.8 & \cellcolor[HTML]{50A377}97.6 & \cellcolor[HTML]{FEE6A9}68.7 & \cellcolor[HTML]{88CB8D}90.8 & \cellcolor[HTML]{5BB581}94.9 & \cellcolor[HTML]{4FA177}97.7 & \cellcolor[HTML]{A5D891}87.9 & \cellcolor[HTML]{55AC7C}96.3 & \cellcolor[HTML]{C4BFDC}$-$8.4 \\
Danish & 55 & 50 & \cellcolor[HTML]{73C187}92.7 & \cellcolor[HTML]{47926F}100.0 & \cellcolor[HTML]{86CA8C}90.9 & \cellcolor[HTML]{47926F}100.0 & \cellcolor[HTML]{5DB682}94.5 & \cellcolor[HTML]{47926F}100.0 & \cellcolor[HTML]{47926F}100.0 & \cellcolor[HTML]{47926F}100.0 & \cellcolor[HTML]{EDF8B9}78.2 & \cellcolor[HTML]{91CF8F}90.0 & \cellcolor[HTML]{73C187}92.7 & \cellcolor[HTML]{47926F}100.0 & \cellcolor[HTML]{80C78A}91.5 & \cellcolor[HTML]{4D9D75}98.3 & \cellcolor[HTML]{D0CCE4}$-$6.8 \\
Persian & 449 & 2,553 & \cellcolor[HTML]{50A377}97.6 & \cellcolor[HTML]{55AB7C}96.4 & \cellcolor[HTML]{54A97B}96.7 & \cellcolor[HTML]{5BB581}94.8 & \cellcolor[HTML]{BE4763}41.2 & \cellcolor[HTML]{F9FDC9}76.1 & \cellcolor[HTML]{4A9872}99.1 & \cellcolor[HTML]{50A377}97.5 & \cellcolor[HTML]{C8E89A}83.7 & \cellcolor[HTML]{8DCD8E}90.3 & \cellcolor[HTML]{4C9A73}98.7 & \cellcolor[HTML]{52A77A}97.0 & \cellcolor[HTML]{B3DF92}86.2 & \cellcolor[HTML]{79C489}92.0 & \cellcolor[HTML]{D6D4E8}$-$5.9 \\
Czech & 1,213 & 4,256 & \cellcolor[HTML]{75C288}92.4 & \cellcolor[HTML]{77C388}92.2 & \cellcolor[HTML]{C0E494}84.8 & \cellcolor[HTML]{79C489}92.1 & \cellcolor[HTML]{59B380}95.2 & \cellcolor[HTML]{52A579}97.2 & \cellcolor[HTML]{52A579}97.2 & \cellcolor[HTML]{55AC7C}96.2 & \cellcolor[HTML]{FBB385}60.8 & \cellcolor[HTML]{C8E89A}83.7 & \cellcolor[HTML]{64BA83}94.1 & \cellcolor[HTML]{58B07F}95.7 & \cellcolor[HTML]{A8DA91}87.4 & \cellcolor[HTML]{71C087}92.9 & \cellcolor[HTML]{D9D8EA}$-$5.4 \\
Swedish & 59 & 201 & \cellcolor[HTML]{6CBE86}93.2 & \cellcolor[HTML]{47926F}100.0 & \cellcolor[HTML]{54A97B}96.6 & \cellcolor[HTML]{47926F}100.0 & \cellcolor[HTML]{6CBE86}93.2 & \cellcolor[HTML]{499570}99.5 & \cellcolor[HTML]{4D9D75}98.3 & \cellcolor[HTML]{47926F}100.0 & \cellcolor[HTML]{B1DE92}86.4 & \cellcolor[HTML]{499570}99.5 & \cellcolor[HTML]{47926F}100.0 & \cellcolor[HTML]{47926F}100.0 & \cellcolor[HTML]{5DB682}94.6 & \cellcolor[HTML]{47926F}99.8 & \cellcolor[HTML]{DADAEB}$-$5.2 \\
Romanian & 246 & 2,056 & \cellcolor[HTML]{50A377}97.6 & \cellcolor[HTML]{54A97B}96.5 & \cellcolor[HTML]{B0DD92}86.6 & \cellcolor[HTML]{58B17F}95.4 & \cellcolor[HTML]{5DB682}94.7 & \cellcolor[HTML]{4FA177}97.7 & \cellcolor[HTML]{58B07F}95.5 & \cellcolor[HTML]{4FA177}97.7 & \cellcolor[HTML]{D4EDA1}82.1 & \cellcolor[HTML]{79C489}92.0 & \cellcolor[HTML]{6ABD85}93.5 & \cellcolor[HTML]{51A478}97.4 & \cellcolor[HTML]{7EC68A}91.7 & \cellcolor[HTML]{55AC7C}96.1 & \cellcolor[HTML]{E0E0EE}$-$4.4 \\
Hindi & 107 & 1,447 & \cellcolor[HTML]{80C78A}91.6 & \cellcolor[HTML]{4C9A73}98.7 & \cellcolor[HTML]{5FB782}94.4 & \cellcolor[HTML]{4A9671}99.3 & \cellcolor[HTML]{6ABD85}93.5 & \cellcolor[HTML]{4B9973}99.0 & \cellcolor[HTML]{4E9E75}98.1 & \cellcolor[HTML]{4A9872}99.2 & \cellcolor[HTML]{75C288}92.5 & \cellcolor[HTML]{4D9D75}98.3 & \cellcolor[HTML]{4A9872}99.1 & \cellcolor[HTML]{4A9671}99.3 & \cellcolor[HTML]{5BB581}94.9 & \cellcolor[HTML]{4B9973}99.0 & \cellcolor[HTML]{E2E3F0}$-$4.1 \\
Serbo-Croatian & 95 & 3,286 & --- & --- & \cellcolor[HTML]{8ACC8D}90.5 & \cellcolor[HTML]{64BA83}94.0 & \cellcolor[HTML]{80C78A}91.6 & \cellcolor[HTML]{5DB682}94.6 & \cellcolor[HTML]{47926F}100.0 & \cellcolor[HTML]{4D9D75}98.3 & \cellcolor[HTML]{FFFAC9}73.7 & \cellcolor[HTML]{A3D791}88.0 & \cellcolor[HTML]{5BB581}94.7 & \cellcolor[HTML]{56AD7D}96.0 & \cellcolor[HTML]{8FCE8E}90.1 & \cellcolor[HTML]{62B883}94.2 & \cellcolor[HTML]{E2E3F0}$-$4.1 \\
Bulgarian & 131 & 2,458 & \cellcolor[HTML]{52A77A}96.9 & \cellcolor[HTML]{52A579}97.2 & \cellcolor[HTML]{CBE99C}83.2 & \cellcolor[HTML]{68BC84}93.6 & \cellcolor[HTML]{52A77A}96.9 & \cellcolor[HTML]{50A377}97.6 & \cellcolor[HTML]{4A9671}99.2 & \cellcolor[HTML]{4A9872}99.2 & \cellcolor[HTML]{EBF7B6}78.6 & \cellcolor[HTML]{93D08F}89.7 & \cellcolor[HTML]{66BB84}93.9 & \cellcolor[HTML]{56AD7D}96.0 & \cellcolor[HTML]{80C78A}91.5 & \cellcolor[HTML]{58B07F}95.5 & \cellcolor[HTML]{E2E3F0}$-$4.1 \\
Belarusian & 171 & 2,570 & \cellcolor[HTML]{52A579}97.1 & \cellcolor[HTML]{51A478}97.3 & \cellcolor[HTML]{C4E697}84.2 & \cellcolor[HTML]{98D38F}89.2 & \cellcolor[HTML]{DEF1A8}80.7 & \cellcolor[HTML]{DFF2A9}80.5 & \cellcolor[HTML]{4D9D75}98.2 & \cellcolor[HTML]{52A77A}96.9 & \cellcolor[HTML]{F9A17D}58.5 & \cellcolor[HTML]{FFFECF}74.7 & \cellcolor[HTML]{62B883}94.2 & \cellcolor[HTML]{6FBF86}93.1 & \cellcolor[HTML]{BAE293}85.5 & \cellcolor[HTML]{9DD590}88.6 & \cellcolor[HTML]{E7E8F2}$-$3.1 \\
Hungarian & 13 & 845 & \cellcolor[HTML]{47926F}100.0 & \cellcolor[HTML]{50A377}97.5 & \cellcolor[HTML]{77C388}92.3 & \cellcolor[HTML]{51A478}97.3 & \cellcolor[HTML]{47926F}100.0 & \cellcolor[HTML]{4D9D75}98.3 & \cellcolor[HTML]{47926F}100.0 & \cellcolor[HTML]{4A9872}99.1 & \cellcolor[HTML]{FEEAAD}69.2 & \cellcolor[HTML]{A3D791}88.0 & \cellcolor[HTML]{47926F}100.0 & \cellcolor[HTML]{4A9872}99.2 & \cellcolor[HTML]{68BC84}93.6 & \cellcolor[HTML]{54A97B}96.6 & \cellcolor[HTML]{E8E9F3}$-$3.0 \\
Slovenian & 168 & 4,483 & \cellcolor[HTML]{50A377}97.6 & \cellcolor[HTML]{68BC84}93.6 & \cellcolor[HTML]{E5F4AD}79.8 & \cellcolor[HTML]{B1DE92}86.4 & \cellcolor[HTML]{5DB682}94.6 & \cellcolor[HTML]{66BB84}93.9 & \cellcolor[HTML]{4E9E75}98.2 & \cellcolor[HTML]{62B883}94.2 & \cellcolor[HTML]{FEE2A6}67.9 & \cellcolor[HTML]{D4EDA1}82.1 & \cellcolor[HTML]{AEDC92}86.9 & \cellcolor[HTML]{82C88B}91.3 & \cellcolor[HTML]{A7D991}87.5 & \cellcolor[HTML]{8DCD8E}90.2 & \cellcolor[HTML]{EAEBF3}$-$2.7 \\
Hebrew & 96 & 2,330 & \cellcolor[HTML]{73C187}92.7 & \cellcolor[HTML]{B7E093}85.9 & \cellcolor[HTML]{88CB8D}90.6 & \cellcolor[HTML]{A1D790}88.1 & \cellcolor[HTML]{C24B63}45.8 & \cellcolor[HTML]{FFF1B9}71.2 & \cellcolor[HTML]{5BB581}94.8 & \cellcolor[HTML]{88CB8D}90.7 & \cellcolor[HTML]{FFFBCA}74.0 & \cellcolor[HTML]{E2F3AB}80.1 & \cellcolor[HTML]{7EC68A}91.7 & \cellcolor[HTML]{91CF8F}90.0 & \cellcolor[HTML]{D8EFA4}81.6 & \cellcolor[HTML]{C4E697}84.3 & \cellcolor[HTML]{EAEBF3}$-$2.7 \\
Spanish & 272 & 2,541 & \cellcolor[HTML]{58B07F}95.6 & \cellcolor[HTML]{55AC7C}96.1 & \cellcolor[HTML]{6FBF86}93.0 & \cellcolor[HTML]{4FA177}97.8 & \cellcolor[HTML]{5BB581}94.9 & \cellcolor[HTML]{4D9D75}98.3 & \cellcolor[HTML]{4E9E75}98.2 & \cellcolor[HTML]{4FA177}97.7 & \cellcolor[HTML]{66BB84}93.8 & \cellcolor[HTML]{56AD7D}96.0 & \cellcolor[HTML]{4FA177}97.8 & \cellcolor[HTML]{4E9E75}98.2 & \cellcolor[HTML]{58B07F}95.5 & \cellcolor[HTML]{51A478}97.4 & \cellcolor[HTML]{EFF0F5}$-$1.8 \\
Lithuanian & 46 & 1,180 & \cellcolor[HTML]{58B07F}95.7 & \cellcolor[HTML]{4FA076}97.9 & \cellcolor[HTML]{ACDB92}87.0 & \cellcolor[HTML]{68BC84}93.6 & \cellcolor[HTML]{4FA177}97.8 & \cellcolor[HTML]{4D9C74}98.5 & \cellcolor[HTML]{4FA177}97.8 & \cellcolor[HTML]{4E9E75}98.1 & \cellcolor[HTML]{82C88B}91.3 & \cellcolor[HTML]{BEE394}85.1 & \cellcolor[HTML]{6ABD85}93.5 & \cellcolor[HTML]{54A97B}96.6 & \cellcolor[HTML]{66BB84}93.8 & \cellcolor[HTML]{5AB481}95.0 & \cellcolor[HTML]{F3F3F7}$-$1.1 \\
English & 212 & 770 & \cellcolor[HTML]{71C087}92.9 & \cellcolor[HTML]{55AB7C}96.4 & \cellcolor[HTML]{4A9872}99.1 & \cellcolor[HTML]{4A9671}99.4 & \cellcolor[HTML]{52A579}97.2 & \cellcolor[HTML]{4A9671}99.4 & \cellcolor[HTML]{4A9872}99.1 & \cellcolor[HTML]{4D9C74}98.6 & \cellcolor[HTML]{499570}99.5 & \cellcolor[HTML]{499570}99.5 & \cellcolor[HTML]{4A9872}99.1 & \cellcolor[HTML]{4B9973}99.0 & \cellcolor[HTML]{4FA177}97.8 & \cellcolor[HTML]{4C9A73}98.7 & \cellcolor[HTML]{F4F5F7}$-$0.9 \\
Catalan & 220 & 2,284 & \cellcolor[HTML]{4FA177}97.7 & \cellcolor[HTML]{4FA177}97.7 & \cellcolor[HTML]{6CBE86}93.2 & \cellcolor[HTML]{5BB581}94.8 & \cellcolor[HTML]{58B17F}95.5 & \cellcolor[HTML]{58B17F}95.5 & \cellcolor[HTML]{4C9A73}98.6 & \cellcolor[HTML]{55AB7C}96.4 & \cellcolor[HTML]{BAE293}85.5 & \cellcolor[HTML]{86CA8C}90.9 & \cellcolor[HTML]{51A478}97.3 & \cellcolor[HTML]{55AC7C}96.1 & \cellcolor[HTML]{5DB682}94.6 & \cellcolor[HTML]{59B380}95.2 & \cellcolor[HTML]{F6F6F8}$-$0.6 \\
German & 428 & 2,298 & \cellcolor[HTML]{5BB581}94.9 & \cellcolor[HTML]{4FA076}98.0 & \cellcolor[HTML]{4FA076}97.9 & \cellcolor[HTML]{4FA076}98.0 & \cellcolor[HTML]{4A9872}99.1 & \cellcolor[HTML]{4C9A73}98.8 & \cellcolor[HTML]{4A9872}99.1 & \cellcolor[HTML]{4C9A73}98.8 & \cellcolor[HTML]{52A579}97.2 & \cellcolor[HTML]{51A478}97.3 & \cellcolor[HTML]{4B9973}98.8 & \cellcolor[HTML]{4B9973}99.0 & \cellcolor[HTML]{4FA177}97.8 & \cellcolor[HTML]{4D9D75}98.3 & \cellcolor[HTML]{F6F6F8}$-$0.5 \\
Ukrainian & 295 & 2,744 & \cellcolor[HTML]{54A97B}96.6 & \cellcolor[HTML]{56AD7D}95.9 & \cellcolor[HTML]{56AD7D}95.9 & \cellcolor[HTML]{5FB782}94.5 & \cellcolor[HTML]{50A377}97.6 & \cellcolor[HTML]{4FA177}97.7 & \cellcolor[HTML]{4A9671}99.3 & \cellcolor[HTML]{4FA177}97.8 & \cellcolor[HTML]{CFEB9E}82.7 & \cellcolor[HTML]{ACDB92}87.1 & \cellcolor[HTML]{4FA076}98.0 & \cellcolor[HTML]{51A478}97.3 & \cellcolor[HTML]{5AB481}95.0 & \cellcolor[HTML]{5AB481}95.0 & \cellcolor[HTML]{F9F9F9}$-$0.0 \\
Turkish & 813 & 1,742 & \cellcolor[HTML]{6CBE86}93.2 & \cellcolor[HTML]{68BC84}93.6 & \cellcolor[HTML]{79C489}92.1 & \cellcolor[HTML]{96D28F}89.3 & \cellcolor[HTML]{59B380}95.2 & \cellcolor[HTML]{71C087}92.8 & \cellcolor[HTML]{4FA177}97.8 & \cellcolor[HTML]{52A579}97.2 & \cellcolor[HTML]{F8FCC7}76.3 & \cellcolor[HTML]{CDEA9D}83.2 & \cellcolor[HTML]{55AC7C}96.2 & \cellcolor[HTML]{64BA83}94.1 & \cellcolor[HTML]{7BC589}91.8 & \cellcolor[HTML]{7EC68A}91.7 & \cellcolor[HTML]{F9F9F8}+0.1 \\
Polish & 764 & 3,272 & \cellcolor[HTML]{4FA076}98.0 & \cellcolor[HTML]{55AB7C}96.3 & \cellcolor[HTML]{5BB581}94.8 & \cellcolor[HTML]{66BB84}93.9 & \cellcolor[HTML]{52A579}97.3 & \cellcolor[HTML]{51A478}97.3 & \cellcolor[HTML]{499570}99.6 & \cellcolor[HTML]{4C9A73}98.8 & \cellcolor[HTML]{BEE394}85.1 & \cellcolor[HTML]{B7E093}85.9 & \cellcolor[HTML]{5BB581}94.9 & \cellcolor[HTML]{53A87A}96.7 & \cellcolor[HTML]{5AB481}94.9 & \cellcolor[HTML]{5BB581}94.8 & \cellcolor[HTML]{F9F9F8}+0.1 \\
Finnish & 400 & 2,570 & \cellcolor[HTML]{4C9A73}98.8 & \cellcolor[HTML]{55AC7C}96.2 & \cellcolor[HTML]{77C388}92.2 & \cellcolor[HTML]{82C88B}91.4 & \cellcolor[HTML]{6ABD85}93.5 & \cellcolor[HTML]{59B380}95.2 & \cellcolor[HTML]{5AB481}95.0 & \cellcolor[HTML]{55AB7C}96.4 & \cellcolor[HTML]{6FBF86}93.0 & \cellcolor[HTML]{80C78A}91.5 & \cellcolor[HTML]{5AB481}95.0 & \cellcolor[HTML]{5FB782}94.5 & \cellcolor[HTML]{5DB682}94.6 & \cellcolor[HTML]{62B883}94.2 & \cellcolor[HTML]{FAF8F5}+0.4 \\
Russian & 4,199 & 3,832 & \cellcolor[HTML]{51A478}97.3 & \cellcolor[HTML]{5FB782}94.5 & \cellcolor[HTML]{4D9D75}98.4 & \cellcolor[HTML]{51A478}97.4 & \cellcolor[HTML]{4C9A73}98.7 & \cellcolor[HTML]{4E9E75}98.2 & \cellcolor[HTML]{499570}99.6 & \cellcolor[HTML]{4D9C74}98.5 & \cellcolor[HTML]{54A97B}96.5 & \cellcolor[HTML]{64BA83}94.1 & \cellcolor[HTML]{4A9671}99.3 & \cellcolor[HTML]{4D9C74}98.6 & \cellcolor[HTML]{4D9D75}98.3 & \cellcolor[HTML]{52A77A}96.9 & \cellcolor[HTML]{FBF3E8}+1.4 \\
\midrule
\textbf{Mean} & & & \cellcolor[HTML]{5AB481}95.1 & \cellcolor[HTML]{55AB7C}96.4 & \cellcolor[HTML]{96D28F}89.4 & \cellcolor[HTML]{62B883}94.2 & \cellcolor[HTML]{C8E89A}83.6 & \cellcolor[HTML]{71C087}92.8 & \cellcolor[HTML]{4FA177}97.8 & \cellcolor[HTML]{50A377}97.6 & \cellcolor[HTML]{EFF8BC}77.8 & \cellcolor[HTML]{98D38F}89.1 & \cellcolor[HTML]{5BB581}94.9 & \cellcolor[HTML]{55AB7C}96.5 & \cellcolor[HTML]{93D08F}89.7 & \cellcolor[HTML]{5FB782}94.4 & \cellcolor[HTML]{DFDFEE}$-$4.7 \\
\bottomrule
\end{tabular}
}
\caption{\blimppi\ against MultiBLiMP \citep{jumeletMultiBLiMP10Massively2026} on the 28 languages both cover over 6 models reported in MultiBLiMP. Two of them, EuroLLM 9B \citep{martinsEuroLLM9BTechnicalReport2025} and OLMo-2 32B \citep{olmo20252olmo2furious}, are not part of our main evaluation. Each model gets a pair of columns: \textsc{b} is \blimppi\ and \textsc{m} is MultiBLiMP. Accuracy (\%) per language; \emph{Pairs\_B} counts \blimppi\ minimal pairs and \emph{Pairs\_M} MultiBLiMP items. MultiBLiMP treats Serbo-Croatian as one language, so the \blimppi\ Serbian and Croatian rows are pooled over items here. Empty cells (---) are excluded from both means of that row. $\Delta$ is the difference of the two means, purple where \blimppi\ is lower and orange where MultiBLiMP is.}
\label{tab:multiblimp-compare}
\end{table*}

\section{Turkish Negation and Tokenization}
\label{app:turkish}

Turkish is learned last by ParalleLM, rising above chance only at the fourth checkpoint, and is the hardest language for the 100MB Goldfish model (62.7). This is unlikely to be typological as the Turkish NPI paradigm is highly regular: 76.6\% of 1,532 NPI occurrences we extracted occur under negation. Moreover, other languages with bound negation do not suffer comparably (Czech 74.5, Persian 89.3). Czech and Persian negate with a \emph{prefix}, which sits at the word onset where subword segmentation is most reliable. Turkish \textit{-mA} is medial and followed by TAME suffixes, compounding the segmentation problem \citep{lernerLikelyUnlikelyBPEbased2025}.

We hypothesize that the Turkish negation suffix is not distinguished from its homophonous nominalizer. Goldfish segments both identically: \textit{duy-ma-dı} (hear-\textsc{neg}-\textsc{past}) and \textit{duy-ma-yı} (hear-\textsc{nmlz}-\textsc{acc}) share the initial token \texttt{\_duyma}. ParalleLM instead fuses \textit{-mA} with the following suffix (\texttt{\_duy|madı}, \texttt{\_duy|mayı}), and scores 75.5 on Turkish with less data.

\end{document}